\pdfoutput=1
\documentclass[11pt]{asiresearch}

\usepackage[T1]{fontenc}
\usepackage{XCharter}
\makeatletter
\IfFileExists{T1Crlt-LF.fd}{%
  \input{T1Crlt-LF.fd}%
  \newcommand{\tablefont}{\fontencoding{T1}\fontfamily{Crlt-LF}\selectfont}%
}{%
  \input{T1Carlito-LF.fd}\relax%
  \newcommand{\tablefont}{\fontencoding{T1}\fontfamily{Carlito-LF}\selectfont}%
}
\makeatother
\usepackage{amsmath,amssymb,booktabs,array,tabularx,graphicx}
\usepackage{listings}
\newcommand{\overviewwidth}{0.78\linewidth}
\usepackage{subcaption}

\usepackage{tikz,listings,xcolor,hyperref,url}
\usepackage{float}
\usepackage{colortbl}%
\usetikzlibrary{arrows.meta,positioning,fit,calc}

\newcommand{\resultrows}{\renewcommand{\arraystretch}{1.04}}
\newcommand{\resultstack}{\renewcommand{\arraystretch}{0.9}}
\definecolor{blueC}{HTML}{3669A5}
\definecolor{tealC}{HTML}{298C79}
\definecolor{grayC}{HTML}{78818C}
\definecolor{orangeC}{HTML}{D88927}
\definecolor{revisionBlue}{HTML}{1556B3}

\newif\ifcompact
\compactfalse
\newcommand{\apx}{\ifcompact Appendix \fi}
\newcommand{\projectpageline}{\vspace{0.5\baselineskip}\projectpage{https://github.com/frontiereval/AIChipBench}\vspace{-0.3\baselineskip}}
\title{\LARGE\bfseries AIChipBench: Benchmarking LLM Agents for\\Automated AI Accelerator Design}

\author{%
  \textbf{Guangyuan Wang}\textsuperscript{1,\,2}\quad
  \textbf{Chengran Li}\quad
  \textbf{Minchen Luo}\textsuperscript{3}\quad
  \textbf{Jichen Feng}\textsuperscript{4,\,5}\\
  \textbf{Mengdi Wang}\textsuperscript{4\,$\dagger$}\quad
  \textbf{Yifan Zhang}\textsuperscript{4\,$\dagger$}
  \\[0.6em]
  {\normalfont\small
    \textsuperscript{1}Mila\quad
    \textsuperscript{2}McGill University\quad
    \textsuperscript{3}Erised Labs}
  \\
  {\normalfont\small
    \textsuperscript{4}Princeton University\quad
    \textsuperscript{5}University of Pennsylvania}
}

\date{}

\begin{document}
\maketitle

\begin{abstract}
A language-model decoder carries state between decoding steps, and in hardware a wrong state update can leave the current token correct and corrupt the output several tokens later, so a design can pass output tests and still be wrong. AIChipBench asks what LLM coding agents need to build such hardware correctly. Agents implement quantized Qwen3.5, Gemma 3, Gemma 4, and Llama 3.2 decoders in synthesizable register-transfer-level (RTL) code against an executable integer reference that specifies the outputs and the stored state at every decoding step. They debug against public tests under fixed budgets; their frozen designs are graded on held-out token sequences, resets, memory stalls, and synthesis, and a second track asks them to speed up a correct design at nearly fixed area. Across 9,136 attempts by four coding agents, the difficulty lies in stored state. Feedback that points to the first state mismatch raises acceptance over output-only feedback in all 12 agent--task cells of a matched study, by 25.0 to 46.9 percentage points, and halves the agent-hours per accepted design. Regraded on new sequences and conditions, 45 of 69 accepted designs fail within the 32 steps their tests required. More than a third of attempts to speed up a correct design end without one. An overall score ranks Sol 5.6 and Fable 5.1 above Astra 6 and Opus 5.5, and both halves of the instances give this order in all 200 random splits. Hardware agents therefore need debugging feedback that exposes stored state, and their benchmarks need final tests well beyond the development cases.
\end{abstract}
\projectpageline

\begin{figure}[H]
  \centering
  \includegraphics[width=\overviewwidth]{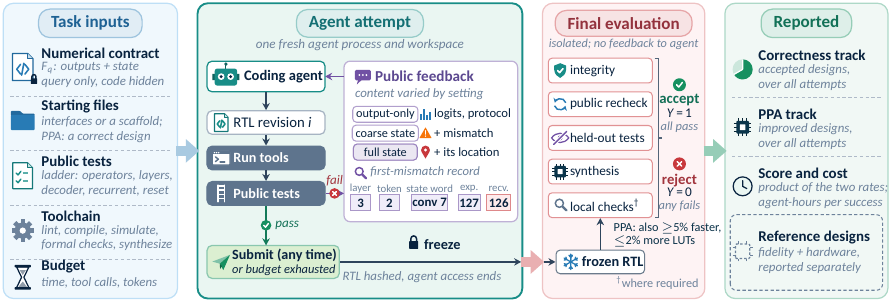}
  \caption{One agent attempt in AIChipBench. The agent starts from interfaces or a scaffold (correctness track) or from a correct design (PPA track) and debugs against public tests whose feedback depends on the setting. Its final RTL is then frozen and must pass every gate of an isolated evaluator; on the PPA track it must also be at least 5\% faster with at most 2\% more LUTs. Rates and cost are computed over all attempts.}
  \label{fig:overview}
\end{figure}

\begin{figure}[t]
\centering
\includegraphics[width=\linewidth]{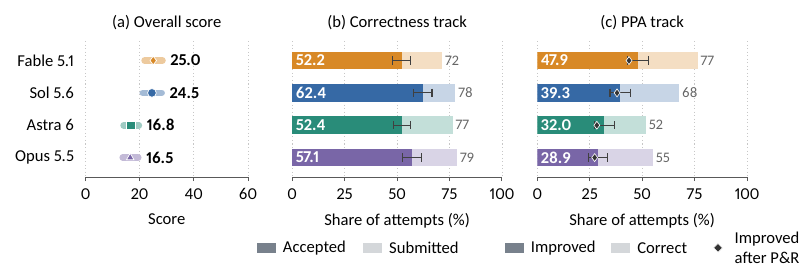}
\caption{\textbf{AIChipBench leaderboard: Sol 5.6 and Fable 5.1 score about 25, Astra 6 and Opus 5.5 below 17.} (a) Overall score, $100\,\bar c\,\bar q$, with paired bootstrap 95\% intervals (Section~\ref{sec:score}). (b) Correctness track: share of attempts that submit a design (light) and that are accepted (dark, $\bar c$), averaged over the twelve tasks. (c) PPA track: share of attempts whose design stays correct (light) and is improved (dark, $\bar q$), averaged over the four targets; diamonds give the share improved at the clock achieved after place and route (P\&R). Whiskers are 95\% intervals of $\bar c$ and $\bar q$. Appendix Table~\ref{tab:leaderboard} lists the values.}
\label{fig:leaderboard}
\end{figure}

\section{Introduction}
\label{sec:intro}

LLM coding agents now edit hardware code and run the compilers, simulators, and synthesis tools that check it. We study them on one such task, building an accelerator for a language-model decoder. At each decoding step, a decoder reads one token, computes scores over its vocabulary (logits), and updates values that later steps reuse, such as the attention key--value cache and the recurrent states of hybrid layers. This persistent state makes errors easy to miss. Suppose a hardware cache saves a result before the next module has taken it. If that module is busy for a few clock cycles, the cache saves the same result again and now holds an entry that should not exist. The token produced at that step is still correct, and the error reaches the output only when a later step reads the cache. In a small decoder we constructed with this defect, the stored state is first wrong at transaction 2, the logits at transaction 5, and the selected token at transaction 6 (Appendix~\ref{app:delayed}). A test that compares output tokens over the first few steps passes this design.

We ask what an agent needs in order to get such state right, and how an evaluator can tell whether it did. Two requirements follow. An agent working within a fixed budget needs feedback that points to the faulty update, because an output mismatch several steps later says little about where the error began. The final verdict must also come from tests the agent never saw, because an agent can fit its design to the tests it debugs against. In our runs, 148 agent-built designs pass every public test and fail final acceptance, and when frozen submissions are regraded, 45 of 69 accepted designs fail a new sequence or condition within the 32 steps their tests required.

AIChipBench (Figure~\ref{fig:overview}) is built around these requirements and has three components. The \emph{LLM coding agent} is a language model run in a tool-using harness: it inspects files, writes and revises hardware code, runs lint, simulation, and synthesis, reads test feedback, and submits a design within caps on wall time, tool calls, and output tokens. The \emph{target}, a quantized Qwen3.5, Gemma 3, Gemma 4, or Llama 3.2 decoder, defines what the hardware must compute. The \emph{integer oracle}, an executable reference that fixes every arithmetic decision of the target, gives the exact outputs and the exact stored state expected at each \emph{transaction}, the point at which the hardware accepts a transfer and commits its result. Agents write register-transfer-level (RTL) code, which describes what a circuit computes and stores on each clock cycle. Synthesis maps that code onto the resources of a chip, here a field-programmable gate array (FPGA), whose logic and wiring are configured after manufacture. A submitted design is frozen and accepted only if it matches the oracle on held-out sequences, resets, and memory stalls, synthesizes, and leaves protected files intact. A second track, on power, performance, and area (PPA), starts the agent from a correct design and asks it to raise throughput by at least 5\% with at most 2\% more area while every check still passes. In the chip-design flow, which runs from architecture through RTL, verification, synthesis, and placement and routing to fabrication, the agents work at the RTL stage; their designs are placed and routed afterwards for measurement (Section~\ref{sec:related}).

We evaluate coding agents, which edit, test, and revise over many tool calls, because building a decoder is a multi-step engineering task. A decoder takes thousands of lines of RTL (6,843 to 31,884 in the seven reference designs of the correctness track), and one clean validation-and-synthesis pass on the 0.8B reference takes 162 minutes, so an attempt's outcome depends on how the agent edits, tests, and spends its budget as well as on the underlying model. This regime differs from single-shot RTL generation, and its results describe a model and its harness together.

Earlier benchmarks cover parts of this setting. CVDP grades tool-using agents that work in a repository against hidden test harnesses, ChipBench includes hierarchical designs and debugging with simulation waveforms, and Build-A-Chip-Bench follows agents from a workload specification to a routed layout (Section~\ref{sec:related}). AIChipBench adds two capabilities aimed at stateful accelerators: an oracle that specifies the committed state of a decoder at every transaction, and a matched study that varies what public feedback reveals (output mismatches only, a flag that state differs, or the first state word that differs) while submissions are frozen and graded independently. Repeated attempts under fixed budgets measure each agent's reliability and cost, both when building a correct design and, on the PPA track, when optimizing one, with place-and-route and board measurements.

Four coding agents make 9,136 attempts, and three results stand out. First, what an agent is told about state largely determines whether it succeeds. Feedback that locates the first state mismatch raises acceptance over output-only feedback in all 12 agent--task cells of a matched study, by 25.0 to 46.9 points, and lowers the agent-hours per accepted design from 13.4 to 6.5. It also beats a flag that only reports a state mismatch in all 12 cells, by 9.4 to 37.5 points, so locating the mismatch helps beyond reporting that state differs (Section~\ref{sec:res_feedback}). Second, passing a test suite does not establish that a design is exact, and output checks in particular miss errors of state. When 144 frozen submissions are regraded, 17 of the 85 sequences that fail within the required 32 steps are wrong only in stored state, so an output check of that length passes them (Section~\ref{sec:evaluator_results}). Third, correctness is fragile under optimization: more than a third of the attempts to speed up a correct design end without one (Section~\ref{sec:findings}). An overall score that multiplies the two track scores ranks Sol 5.6 and Fable 5.1 above Astra 6 and Opus 5.5 (Figure~\ref{fig:leaderboard}), and in all 200 random splits of the instances into halves, both halves put the same two agents on top.

\section{AIChipBench}
\label{sec:design}
\label{sec:method}
\label{sec:background}

\subsection{Task shape}
\label{sec:design_overview}
\label{sec:bg_problem}
\label{sec:contract}
Each task asks an agent to implement a quantized language-model decoder, from embedding lookup to output projection, in synthesizable RTL (Figure~\ref{fig:overview}). The agent works in a sandboxed copy of a starting workspace, with tools for lint, compilation, simulation, bounded property checks, and synthesis, under caps on wall time, tool calls, and output tokens. An \emph{attempt} is one fresh agent process in a fresh workspace; it ends when the agent submits or reaches a cap.

What a correct design computes is fixed by the task's \emph{numerical contract}, the integer reference of Section~\ref{sec:intro}. For each target we fix the integer arithmetic (bit widths, accumulation order, rounding, saturation, and the lookup tables that approximate nonlinear functions), so that one decoding step is an integer state transition
\begin{equation}
(z_t,s_{t+1})=F_q(x_t,p_t,s_t;W),\qquad
s_t=(s_t^{\rm GDN},s_t^{\rm conv},K_t,V_t,m_t),
\label{eq:transition}
\end{equation}
from token $x_t$ at position $p_t$, persistent state $s_t$, and weights $W$ to logits $z_t$ and the next state. The state holds the recurrent and convolution state of Gated DeltaNet (GDN) layers, attention key and value caches $K_t$ and $V_t$, and valid lengths and control metadata $m_t$; the GDN parts are empty for attention-only decoders. Each integer in the state is a \emph{state word}. The contract runs as an integer oracle outside the agent's workspace; agents can query it but not read its code (Appendix~\ref{app:bindings}).

\label{sec:feedback}
Public tests in the workspace compare the agent's RTL with the oracle, from single operators and layers up to the full decoder and multi-step decoding. By default, a failed test reports the first mismatching state word: its layer, token, tensor, and index, and the expected and received integers (\emph{full-state} feedback). Two weaker settings, used in Section~\ref{sec:res_feedback}, report output mismatches only (\emph{output-only}) or report that state mismatched without locating it (\emph{coarse-state}).

\label{sec:acceptance}
When the agent submits, its RTL is hashed, frozen, and graded on an independent clean copy. A submission is \emph{accepted} only if it passes every gate:
\begin{equation}
Y=I_{\rm integrity}\cdot I_{\rm public}\cdot I_{\rm heldout}\cdot I_{\rm synth}\cdot I_{\rm local}.
\label{eq:acceptance}
\end{equation}
The gates are (i) an integrity check on protected files; (ii) the full public suite, rerun with state checks whatever feedback the agent saw; (iii) a held-out suite with new quantized weights and token sequences, nonzero cache positions, reset after a nonempty token sequence (\emph{reset after activity}), bounded memory stalls, and parameter reload (Appendix~\ref{app:tests}); (iv) synthesis; and (v) for all tasks except the two nano tasks, bounded safety checks on reduced instances of the controller (\emph{local checks}). Held-out results never return to the agent. On the correctness track, an attempt that ends at a cap is not graded and counts as a failure. Section~\ref{sec:evaluator_results} regrades accepted designs under new conditions.

\paragraph{An example task.}
One task asks the agent to adapt a supplied accelerator scaffold to the official 24-layer Qwen3.5-0.8B text configuration~\citep{qwen08} (\apx{}Figure~\ref{fig:task_card}). The scaffold is a working parameterized accelerator, so the module structure is given, and the agent writes the RTL that fits it to the target. Every task has a reference design that passes all gates, so the task can be completed. \ifcompact The 4B reference was derived from a gate-passing submission of a model outside the four evaluated agents, and the others were written by hand or from reviewed agent-drafted components (Appendix~\ref{app:bindings}).\else The nano and Gemma 4 E4B references were written by hand and reviewed independently; for the 0.8B and Llama 3.2 3B references, an agent drafted components that an engineer reviewed, edited, and integrated; the 4B reference was derived from a gate-passing submission of a model outside the four evaluated agents, with 70 lines changed in its reset sequencing and parameter-loading wrapper (Appendix~\ref{app:bindings}).\fi{} The 0.8B reference decodes 3.999 tokens per second at a 126\,MHz clock, and its contract agrees with the 32-bit floating-point (FP32) model on 99.14\% of top-1 next-token decisions.

\ifcompact\else\ifcompact\begin{figure}[htbp]\else\begin{figure}[tb]\fi
\centering
\small
\setlength{\fboxsep}{6pt}\setlength{\fboxrule}{.6pt}
\fcolorbox{black!35}{black!2}{%
\begin{minipage}{.95\linewidth}
{\normalsize\textbf{Qwen3.5-0.8B adaptation}}\hfill{\footnotesize correctness track; 32 attempts per agent}\par\vspace{3pt}
\renewcommand{\arraystretch}{1.12}\setlength{\tabcolsep}{3pt}
\begin{minipage}[t]{.53\linewidth}
\textbf{Task}\par\vspace{1pt}
\begin{tabularx}{\linewidth}{@{}>{\raggedright\arraybackslash}p{.27\linewidth}>{\raggedright\arraybackslash}X@{}}
Workload & official text configuration: 24 layers (18 GDN, 6 full attention), width 1,024 \\
Supplied & shared parameterized accelerator scaffold (module structure given) and supporting files; numerical contract queryable as an oracle, its code hidden \\
Agent writes & RTL adapting the scaffold to the 24-layer configuration \\
Feedback & full-state public feedback: outputs and persistent state, first mismatch located \\
Budget & 12\,h wall time, 420 tool calls, 260k output tokens; the first exhausted cap ends the attempt \\
Gates & after freeze: public suite (3,120 cases), held-out suite (232), synthesis, integrity, local checks \\
\end{tabularx}
\end{minipage}\hfill
\begin{minipage}[t]{.43\linewidth}
\textbf{Reference design} (baseline schedule)\par\vspace{1pt}
\begin{tabularx}{\linewidth}{@{}>{\raggedright\arraybackslash}X r@{}}
Target clock & 126\,MHz \\
Batch; prompt & 1; 8 tokens \\
Throughput & 3.999 tokens/s \\
Cycles per token & 31.5M \\
LUT / FF & 61.6k / 79.3k \\
BRAM / DSP & 151 / 224 \\
Public exact & 3,120/3,120 \\
Held-out exact & 232/232 \\
FP32 agreement & 7,931/8,000 (99.14\%) \\
\end{tabularx}
\par\vspace{2pt}{\footnotesize LUT, FF, BRAM, and DSP: post-route means over three implementation seeds. Agreement counts top-1 decisions.}
\end{minipage}
\end{minipage}}
\caption{Task card for the Qwen3.5-0.8B adaptation task. The left side gives what the agent receives, what it must write, and how the attempt is limited and graded (Section~\ref{sec:design}, Table~\ref{tab:sa-budgets}); the right side summarizes the reference design that qualifies the workload (Tables~\ref{tab:extended_reference} and~\ref{tab:physical}).}
\label{fig:task_card}
\end{figure}
\fi

\subsection{What AIChipBench stresses}
\label{sec:bg_decoding}
A decoder task combines \ifcompact four\else five\fi{} demands.

\textbf{Integer arithmetic to the bit.} The contract fixes every rounding and saturation decision. Requantization, for example, rounds to the nearest integer with ties to even and then clips (Appendix~\ref{app:rtl}), and a design with the wrong rounding mode fails the public tests (Table~\ref{tab:mutation_checks}).

\textbf{State committed only on accepted transfers.} A decoding step takes many clock cycles, and modules exchange data through valid/ready handshakes: a transfer happens only in a cycle in which the sender's data is valid and the receiver is ready, and otherwise the sender must hold its data (backpressure). A state word may change only when the transfer that produces it is accepted; a design that writes state whenever a result is valid can commit a value twice under backpressure (Appendix~\ref{app:decoder_case}). Because designs may pipeline or reorder work, correctness is checked at each \emph{transaction}, one accepted transfer of an input and its result, and the cycle-level timing is left to the design.

\textbf{State that stays exact across steps and conditions.} Later steps read what earlier steps commit, so a state defect can surface late, as in the constructed decoder of Section~\ref{sec:intro} (Appendix~\ref{app:decoder_case}). There, a scripted repair that guards only the cache payload makes every output exact while the state and commit checks still fail. A design must also stay exact when reset after activity, at nonzero cache positions, and when memory responses stall. A test that starts from reset at position zero with an always-ready memory exercises none of these.

\ifcompact\else
\textbf{Hardware quality without loss of exactness.} In the power, performance, and area (PPA) track (Section~\ref{sec:ppa}), a correct design is the starting point, and the agent must make it faster at nearly fixed area while every correctness check still passes.

\fi
\textbf{Slow tools under a fixed budget.} One clean pass of validation and synthesis on the 0.8B reference design takes 162 minutes\ifcompact{} (Appendix~\ref{app:traces})\else, 31 of them for synthesis and timing (Appendix~\ref{app:traces})\fi, so a 12-hour attempt fits fewer than five such passes.

\subsection{Correctness track: tasks and scoring}
\label{sec:tasks}
\label{sec:metrics}
\label{sec:bg_numerics}
The correctness track (Table~\ref{tab:benchmark}) has twelve tasks on seven targets from three model families: Qwen3.5~\citep{qwen08,qwen4}, Gemma 3 and 4~\citep{gemma3card,gemma4card}, and Llama 3.2~\citep{llama32card}. Tasks differ in the starting workspace. In \emph{construction from interfaces}, the workspace supplies module declarations, testbench interfaces, arithmetic components, and supporting files, but no decoder implementation; in \emph{scaffold adaptation}, it supplies a working accelerator to adapt. Tasks also differ in the decoding horizon, the number of decoding steps the tests require. Two tasks use Qwen3.5-nano, our name for a compact Qwen-derived configuration with four layers (three GDN, one full attention), width 256, and a 4,096-token vocabulary (Appendix~\ref{app:nano}): T2 decodes one step and T3 eight recurrent steps. The labels follow an earlier task series in which T1 is a single decoder layer (Appendix~\ref{sec:agents}). The other ten tasks require 32 steps. Across the twelve tasks, reference designs have 6,843 to 31,884 lines of RTL\ifcompact{} (Table~\ref{tab:rtl_scale}).\else. A scaffold supplies 16,308 to 27,492 of them, and accepted adaptations add a median of 1,382 to 2,311 lines; from interfaces, the agent starts with 284 to 746 lines and accepted designs add a median of 6,154 to 34,300 (Appendix Table~\ref{tab:rtl_scale}).\fi

\begin{table}[tb]
\centering
\caption{\textbf{The AIChipBench correctness track.} Twelve tasks on seven decoder targets; each entry in a caps column is one task. Steps: decoding steps the tests require. Cases: public / held-out test cases. Caps: wall time (h) / tool calls / output tokens per attempt, by starting workspace. Agreement: pooled top-1 agreement of the target's integer contract with FP32.}
\label{tab:benchmark}
{\footnotesize\tablefont\setlength{\tabcolsep}{3.5pt}\renewcommand{\arraystretch}{1.1}
\begin{tabular*}{\linewidth}{@{\extracolsep{\fill}}lccccc@{}}
\toprule
Target & Steps & Cases & Caps from interfaces & Caps from a scaffold & Agreement \\
\midrule
Qwen3.5-nano T2 (4 layers) & 1 & 1,320 / 64 & 5 / 260 / 140k & -- & 97.22\% \\
Qwen3.5-nano T3 (4 layers) & 8 & 1,384 / 96 & 6 / 320 / 180k & -- & 97.22\% \\
Qwen3.5-0.8B (24 layers) & 32 & 3,120 / 232 & -- & 12 / 420 / 260k & 99.14\% \\
Qwen3.5-4B (32 layers) & 32 & 3,680 / 288 & -- & 18 / 620 / 380k & 99.21\% \\
Gemma 4 E4B & 32 & 1,600 / 240 & 30 / 900 / 600k & 18 / 600 / 360k & 99.78\% \\
Llama 3.2 3B & 32 & 1,600 / 240 & 28 / 860 / 560k & 16 / 560 / 340k & 99.76\% \\
Gemma 3 4B (text) & 32 & 1,600 / 240 & 30 / 900 / 600k & 18 / 600 / 360k & 99.00\% \\
Llama 3.2-1B & 32 & 1,600 / 240 & 28 / 860 / 560k & 16 / 560 / 340k & 99.31\% \\
\bottomrule
\end{tabular*}}
\par\smallskip{\footnotesize\raggedright Attempts per agent and task: 32 on the nano tasks and Qwen3.5-0.8B, 64 on Gemma 3 4B and Llama 3.2-1B, 24 on the others. Every task except the nano tasks also requires local checks. Qwen3.5-nano falls below the 99\% fidelity threshold (Section~\ref{sec:tasks}).\par}
\end{table}

A target \emph{qualifies} when its reference RTL passes every gate and its contract agrees with the FP32 model on at least 99\% of pooled top-1 next-token decisions over common prefixes, a threshold set by this benchmark. The two conditions are distinct. \emph{Exactness} is whether RTL reproduces the contract, and it is what agents are graded on. \emph{Fidelity} is how closely the contract approximates the floating-point model; it is a property of the workload, and either can hold without the other (Section~\ref{sec:correctness}). Qwen3.5-nano falls below the fidelity threshold at 97.22\%; we keep it because its tasks measure exactness and its small size keeps repeated attempts affordable. The Gemma 3 4B (text) and Llama 3.2-1B references qualify: each passes all 1,600 public and 240 held-out cases of its 32-step suite, and their contracts agree with FP32 on 99.00\% and 99.31\% of decisions; both also meet timing (\apx{}Tables~\ref{tab:extended_reference} and~\ref{tab:T09-gates}).

On this track we report an agent's acceptance rate over all attempts, $\hat p=\sum_iY_i/N$, with a Wilson 95\% interval. We also report agent-hours per accepted design: total agent time, failed attempts included, divided by the number accepted (Appendix~\ref{app:cost}). Two diagnostic studies extend the track: a feedback study repeats three tasks under weaker feedback (Section~\ref{sec:res_feedback}), and a controlled matrix varies the horizon, an added state condition, and the budget from one nano scaffold (Section~\ref{sec:state_budget}).\ifcompact\else{} Appendix~\ref{sec:agents} covers the earlier task series, run by two cohorts at different times.\fi

\subsection{PPA track}
\label{sec:ppa}
The PPA track measures hardware quality on four targets: Qwen3.5-0.8B and 4B, Gemma 4 E4B, and Llama 3.2 3B. Every design targets an AMD Alveo U250 FPGA (xcu250-figd2104-2L-e) and is evaluated at a 126\,MHz target clock, batch size one, an eight-token prompt, and a common memory-service model. Throughput is $R=f/C$ for clock frequency $f$, where $C$ counts all elapsed cycles per decoded token, memory stalls included; area is the FPGA look-up-table (LUT) count $L$; power $P$ covers the FPGA and attached DRAM and excludes the host; and energy per token is $P/R$. An agent starts from the correct reference design and is asked to improve its throughput, resources, or energy efficiency. Attempt $i$ counts as \emph{improved} only if the design stays correct, gains at least 5\% throughput, and grows its LUT count by at most 2\%:
\begin{equation}
S_i=I^{\rm correct}_i\cdot I\!\left[R_i/R_0-1\ge0.05\right]\cdot I\!\left[L_i/L_0-1\le0.02\right],
\label{eq:ppa}
\end{equation}
where $R_0$ and $L_0$ are the evaluator's values for the starting design: 3.999 tokens/s and 61.8k LUTs for 0.8B, 0.800 tokens/s and 69.0k LUTs for 4B, 0.940 tokens/s and 74.4k LUTs for Gemma 4 E4B, and 1.120 tokens/s and 66.6k LUTs for Llama 3.2 3B. Energy efficiency is reported but does not enter the criterion. Each agent makes 96 attempts per target. The 0.8B and 4B reference designs are also carried through the backend: we place and route them with three implementation seeds, check timing closure at 126\,MHz, and measure power on an FPGA board (Section~\ref{sec:hardware}). Agent designs are scored on the evaluator's throughput, LUT, and energy values\ifcompact{} and routed afterwards as a check\else; as a check, every correct agent design is also placed and routed with three seeds\fi{} (Section~\ref{sec:hardware}).

\paragraph{Physical implementation.}
\label{sec:physical_flow}
The agent edits only the RTL; placement and routing run after it submits. Designs are implemented with Vivado 2024.1, AMD's placement-and-routing flow for its devices~\citep{vivado_impl}, which maps a synthesized design onto the U250's fixed look-up tables, block RAM, DSP slices, and programmable interconnect and produces a bitstream. OpenROAD~\citep{openroad} automates a different setting, ASIC physical design: its flow takes RTL for a chosen process design kit through floorplanning, placement, clock-tree synthesis, and routing to a layout of standard cells and memory macros. Vivado is the implementation flow for the U250, on which a routed design can be run and its power measured on the board; every routed and board result here is an FPGA result. An ASIC version would need a process design kit, cell and memory libraries, and physical verification, and its results would not be comparable with these FPGA numbers (Appendix~\ref{app:fpga_asic}).

\paragraph{Overall score.}
\label{sec:score}
One number summarizes both tracks: $\mathrm{Score}=100\,\bar c\,\bar q$, where $\bar c$ is the agent's acceptance rate averaged over the twelve correctness tasks and $\bar q$ its rate of improved designs averaged over the four PPA targets, both as fractions. The score is the product of the two track percentages, so it is high only when both are. All agents attempt the same task instances, so the score's 95\% interval is a paired bootstrap over instances (Appendix~\ref{app:add-score}).

\section{Results}
\label{sec:evaluation}
\label{sec:results}
\label{sec:setup}
\label{sec:sa-comparison}
We evaluate four coding agents, Sol 5.6, Astra 6, Opus 5.5, and Fable 5.1, all run through the MathCode harness~\citep{mathcode2026}. Each makes 472 attempts on the correctness track and 384 on the PPA track. The four agents receive the same task instances, starting files, instructions, tools, feedback, and caps; a \emph{task instance} fixes the initial RTL files, quantized parameter arrays, public inputs, and held-out test assignment. The diagnostic studies bring the total to 9,136 attempts (Appendix Table~\ref{tab:experiments}).

\paragraph{Overall score.}
Fable 5.1 and Sol 5.6 score 25.0 and 24.5, and Astra 6 and Opus 5.5 score 16.8 and 16.5 (Figure~\ref{fig:leaderboard}); the gap comes mostly from the PPA track. To test whether the order depends on which instances are drawn, we split the matched instances of every task into random halves: both halves rank Sol 5.6 and Fable 5.1 above Astra 6 and Opus 5.5 in all 200 splits, against 17\% of splits when the agents' outcomes are permuted within instances ($p=0.001$; Appendix Table~\ref{tab:T19-split-half}). Each track and the controlled matrix also yield an order that recurs across halves, though the orders differ (Appendix~\ref{app:add-agents}). \ifcompact With routed PPA values the same two agents lead: Sol, Fable, Astra, and Opus score 23.7, 22.8, 14.9, and 15.6 (Figure~\ref{fig:key_hardware}b).\else Scoring the PPA factor at the clock each design achieves after place and route gives 23.7 for Sol 5.6, 22.8 for Fable 5.1, 14.9 for Astra 6, and 15.6 for Opus 5.5, and all four differences between an agent of the first pair and one of the second stay below 0.05 after a Holm adjustment (Figure~\ref{fig:key_hardware}b, Appendix Table~\ref{tab:T06_scores}).\fi

\subsection{Correctness track}

Sol 5.6 has 61.2\% of its attempts accepted, Opus 5.5 55.9\%, Astra 6 49.8\%, and Fable 5.1 48.3\%, at 23.9 to 31.4 agent-hours per accepted design (Appendix Table~\ref{tab:leaderboard}). On matched instances, Sol 5.6 is ahead of Astra 6 and Fable 5.1, and Opus 5.5 is ahead of Fable 5.1 (Holm-adjusted $p\le0.024$; Appendix~\ref{app:add-agents}). \ifcompact About 26\% of attempts end at a cap without submitting, and 55 to 70\% pass each suite but only 48 to 61\% pass every gate.\else About 26\% of attempts end at a cap without submitting. Between 55 and 70\% of attempts pass each suite, but only 48 to 61\% pass every gate.\fi

\subsection{PPA track}
\label{sec:hardware}
\label{sec:matched_hardware}
\label{sec:energy}
\ifcompact\else\begin{figure}[tbp]
\centering
\includegraphics[width=\linewidth]{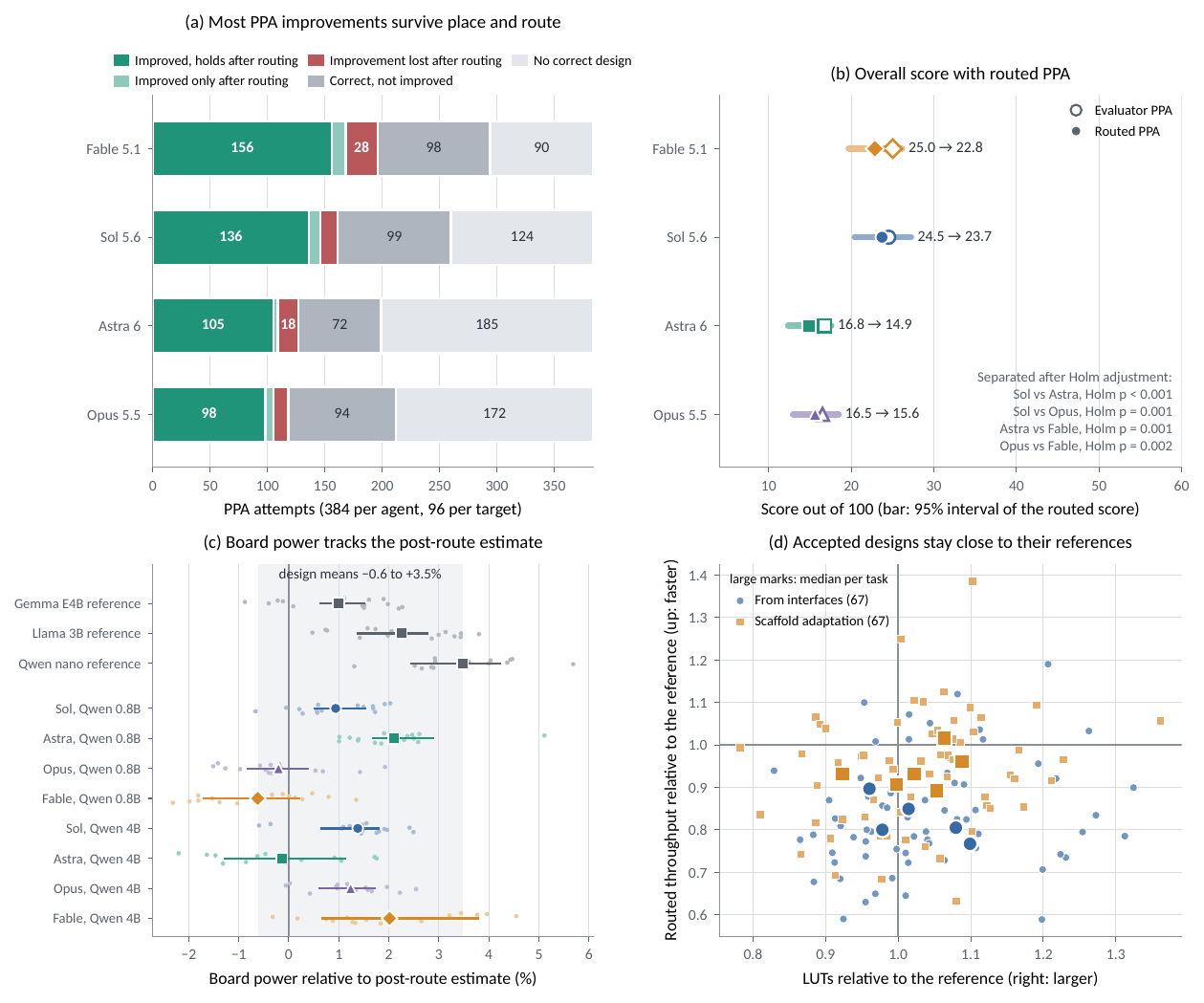}
\caption{\textbf{PPA designs after place and route, and agent designs on the board.} (a) Outcome of the 384 PPA attempts per agent, 96 on each target, when each correct design is placed and routed with three seeds and judged by Equation~\ref{eq:ppa} at its achieved clock. (b) Overall score with the evaluator's PPA values (hollow) and with routed PPA values (filled); bars are 95\% bootstrap intervals of the routed score. (c) Board power relative to the post-route activity estimate for three references and eight agent designs: mean of 15 measurement windows (marker), range of the three seed means (line), and single windows (dots); the shaded band spans the design means. (d) Routed throughput and LUTs of the 134 accepted correctness-track designs whose three routes complete, relative to the task's reference; large marks are per-task medians.}
\label{fig:key_hardware}
\end{figure}
\fi
\begin{table}[tb]
\centering
\caption{\textbf{PPA track: the best agent improves a correct design in 184 of 384 attempts.} Correct: the submission stays exact on the full evaluator. Improved: correct, with throughput at least 5\% higher and LUT count at most 2\% higher (Equation~\ref{eq:ppa}), with a Wilson 95\% interval; then the improved count on each target (96 attempts each). Throughput and efficiency: median gains among improved attempts. Hours: agent-hours per improved design, every attempt counted.}
\label{tab:ppa}
{\footnotesize\tablefont\setlength{\tabcolsep}{2.5pt}\resultrows
\begin{tabular*}{\linewidth}{@{\extracolsep{\fill}}lcccccccccc@{}}
\toprule
Agent & Correct & Improved & {\resultstack\begin{tabular}[b]{@{}c@{}}Qwen\\0.8B\end{tabular}} & {\resultstack\begin{tabular}[b]{@{}c@{}}Qwen\\4B\end{tabular}} & {\resultstack\begin{tabular}[b]{@{}c@{}}Gemma\\E4B\end{tabular}} & {\resultstack\begin{tabular}[b]{@{}c@{}}Llama\\3B\end{tabular}} & Throughput & Efficiency & Hours \\
\midrule
\textbf{Fable 5.1} & 294/384 & \cellcolor{blueC!26}\textbf{184/384}\,\textsubscript{\tiny(43.0--52.9)} & 40 & 55 & 42 & 47 & +14.5\% & +10.8\% & 24.9 \\
\textbf{Sol 5.6} & 260/384 & \cellcolor{blueC!22}\textbf{151/384}\,\textsubscript{\tiny(34.6--44.3)} & 42 & 44 & 32 & 33 & +17.0\% & +13.4\% & 31.6 \\
\textbf{Astra 6} & 199/384 & \cellcolor{blueC!19}\textbf{123/384}\,\textsubscript{\tiny(27.6--36.9)} & 32 & 27 & 31 & 33 & +12.9\% & +10.3\% & 39.8 \\
\textbf{Opus 5.5} & 212/384 & \cellcolor{blueC!18}\textbf{111/384}\,\textsubscript{\tiny(24.6--33.6)} & 18 & 42 & 27 & 24 & +12.2\% & +10.0\% & 43.0 \\
\bottomrule
\end{tabular*}
}
\par\smallskip{\footnotesize\raggedright Rows are ordered by improved designs. Efficiency gain is $(R/P)/(R_0/P_0)-1$. Per-target outcomes and hardware changes among all correct submissions: Tables~\ref{tab:T12-outcomes} and~\ref{tab:T12-changes}.\par}
\end{table}

Of 1,536 attempts, 965 keep the design correct and 569 also meet Equation~\ref{eq:ppa} (Table~\ref{tab:ppa}). Fable 5.1 improves the design in 184 of 384 attempts and Sol 5.6 in 151, against 123 for Astra 6 and 111 for Opus 5.5, a ranking that differs from the correctness track's (matched test, $p<0.001$). Among improved designs, the median throughput gain per agent is 12.2 to 17.0\%; counting every attempt, an improved design costs 24.9 to 43.0 agent-hours. Because the criterion caps LUT growth, these are gains at nearly fixed area; over all correct submissions, gains are smaller and LUT use rises slightly (Table~\ref{tab:T12-changes}).\ifcompact\else{} Throughput and LUT counts are also recorded for the accepted Qwen3.5-0.8B designs of an earlier Sol 5.6 and Astra 6 study on the correctness track, which run at 4.225 to 4.527 tokens/s with about 62k LUTs (Appendix~\ref{sec:agent_hardware}).\fi

\paragraph{Routed agent designs.} We placed and routed every correct PPA design with three seeds, and 925 of the 965 complete all three routes. At the achieved clock, 528 attempts meet Equation~\ref{eq:ppa}: 495 of the 569 nominal improvements hold, 74 are lost, and 33 are gained (\ifcompact Figure~\ref{fig:key_main}c\else Figure~\ref{fig:key_hardware}a\fi, Appendix Table~\ref{tab:T05_ppa}).\ifcompact\else{} Fable 5.1 loses the most, 28, and keeps the most improved designs, 168, against 146 for Sol 5.6. On the twelve tasks of the correctness track, accepted designs stay close to their references: over the 1,016 synthesized designs the median LUT ratio per task is 0.97 to 1.10, and in a routed sample of 144 designs the median throughput per task is 0.77 to 1.02 times the reference's (Figure~\ref{fig:key_hardware}d, Appendix~\ref{app:add-hardware}).\fi

\begin{table}[tb]
\centering
\caption{\textbf{The optimized reference schedules raise throughput by 11.4 to 12.0\% and efficiency by 7.5 to 8.5\%, after place and route and on the board.} Post-route and board results for the paired Qwen3.5-0.8B and 4B schedules. Top: baseline; bold: optimized. Means over three implementation seeds.}
\label{tab:physical}
{\small\tablefont\setlength{\tabcolsep}{4pt}\resultrows
\begin{tabular*}{\linewidth}{@{\extracolsep{\fill}}llcccccc@{}}
\toprule
Workload & Stage & {\resultstack\begin{tabular}[b]{@{}c@{}}Throughput\\(tokens/s)\end{tabular}} & {\resultstack\begin{tabular}[b]{@{}c@{}}Throughput\\gain\end{tabular}} & {\resultstack\begin{tabular}[b]{@{}c@{}}Power\\(W)\end{tabular}} & {\resultstack\begin{tabular}[b]{@{}c@{}}Energy\\(J/token)\end{tabular}} & {\resultstack\begin{tabular}[b]{@{}c@{}}Efficiency\\gain\end{tabular}} & {\resultstack\begin{tabular}[b]{@{}c@{}}Seeds at\\126\,MHz\end{tabular}} \\
\midrule
Qwen3.5-0.8B & Post-route & {\resultstack\begin{tabular}[b]{@{}c@{}}3.999\\\textbf{4.455}\end{tabular}} & \cellcolor{tealC!35}\textbf{+11.4\%} & {\resultstack\begin{tabular}[b]{@{}c@{}}8.95\\\textbf{9.27}\end{tabular}} & {\resultstack\begin{tabular}[b]{@{}c@{}}2.237\\\textbf{2.081}\end{tabular}} & \cellcolor{tealC!24}\textbf{+7.5\%} & {\resultstack\begin{tabular}[b]{@{}c@{}}3/3\\\textbf{1/3}\end{tabular}} \\
 & Board & {\resultstack\begin{tabular}[b]{@{}c@{}}3.985\\\textbf{4.457}\end{tabular}} & \cellcolor{tealC!36}\textbf{+11.8\%} & {\resultstack\begin{tabular}[b]{@{}c@{}}8.98\\\textbf{9.27}\end{tabular}} & {\resultstack\begin{tabular}[b]{@{}c@{}}2.254\\\textbf{2.079}\end{tabular}} & \cellcolor{tealC!27}\textbf{+8.5\%} & {\resultstack\begin{tabular}[b]{@{}c@{}}3/3\\\textbf{1/3}\end{tabular}} \\
\addlinespace[3pt]
Qwen3.5-4B & Post-route & {\resultstack\begin{tabular}[b]{@{}c@{}}0.800\\\textbf{0.896}\end{tabular}} & \cellcolor{tealC!36}\textbf{+12.0\%} & {\resultstack\begin{tabular}[b]{@{}c@{}}10.05\\\textbf{10.44}\end{tabular}} & {\resultstack\begin{tabular}[b]{@{}c@{}}12.564\\\textbf{11.650}\end{tabular}} & \cellcolor{tealC!25}\textbf{+7.8\%} & {\resultstack\begin{tabular}[b]{@{}c@{}}3/3\\\textbf{3/3}\end{tabular}} \\
 & Board & {\resultstack\begin{tabular}[b]{@{}c@{}}0.797\\\textbf{0.891}\end{tabular}} & \cellcolor{tealC!35}\textbf{+11.7\%} & {\resultstack\begin{tabular}[b]{@{}c@{}}10.05\\\textbf{10.42}\end{tabular}} & {\resultstack\begin{tabular}[b]{@{}c@{}}12.602\\\textbf{11.703}\end{tabular}} & \cellcolor{tealC!25}\textbf{+7.7\%} & {\resultstack\begin{tabular}[b]{@{}c@{}}3/3\\\textbf{3/3}\end{tabular}} \\
\bottomrule
\end{tabular*}
}
\par\smallskip{\footnotesize\raggedright Power covers the FPGA and attached DRAM during batch-one decode. Post-route rows report implementation-stage values; board rows are measured on the board with five repeats per seed. Throughput uses each implementation's operating clock. Efficiency gain is the baseline-to-optimized ratio of J/token minus one. Resources, slack, and power components: Tables~\ref{tab:T10-results} and~\ref{tab:T10-components}.\par}
\end{table}

\paragraph{Backend results on the reference designs.}
Each reference design has a baseline and an optimized \emph{schedule}; the optimized one changes buffering, overlap, and scheduling and keeps the arithmetic and memory-service model.\ifcompact\else{} It raises throughput by 12.0\% and power by about 3.9\%, an estimated efficiency gain of 7.8\% (Table~\ref{tab:reference_hardware}).\fi{} After place and route with three seeds, every baseline and every optimized 4B implementation closes timing at 126\,MHz; the optimized 0.8B design closes on one of three seeds, so its mean operating clock is 125.3\,MHz. Post-route and board results give throughput gains of 11.4 to 12.0\% and efficiency gains of 7.5 to 8.5\% (Table~\ref{tab:physical}).\ifcompact\else{} In the cycle model, compute takes 65\% of the baseline cycles per token and memory waits 28\% (Appendix~\ref{app:add-schedule}).\fi \ifcompact\else Post-route results also cover the nano, Gemma 4 E4B, Llama 3.2 3B, Gemma 3 4B, and Llama 3.2-1B references, and on eight agent designs and three references board power is within $-0.6$ to $+3.5\%$ of the post-route activity estimate (Figure~\ref{fig:key_hardware}c, Appendix~\ref{app:add-physical}).\fi{} Board power is read from the U250's on-board rail sensors (Appendix~\ref{app:add-physical}).

\subsection{Reference designs beyond the benchmark targets}
\label{sec:correctness}
\ifcompact\else\begin{table}[tb]
\centering
\caption{\textbf{RTL exactness and contract fidelity can each fail while the other holds.} Six reference workloads under the two numerical criteria. Five references pass every RTL check; Qwen3.6-27B has integration failures, and Gemma 2's contract misses the 99\% fidelity threshold.}
\label{tab:extended_reference}\label{tab:bench_overview}
{\small\tablefont\setlength{\tabcolsep}{3pt}\resultrows
\begin{tabular*}{\linewidth}{@{\extracolsep{\fill}}lrrr>{\bfseries}r}
\toprule
 & \multicolumn{2}{c}{\textbf{RTL vs. integer oracle}} & \multicolumn{2}{c}{\textbf{Integer oracle vs. FP32}} \\
\cmidrule(lr){2-3}\cmidrule(lr){4-5}
Target workload & Public exact & Held-out exact & Top-1 matches & Agreement (\%) \\
\midrule
Qwen3.5-0.8B & 3120/3120 & 232/232 & 7931/8000 & 99.14 \\
Qwen3.5-4B & 3680/3680 & 288/288 & 7937/8000 & 99.21 \\
\addlinespace[3pt]
Gemma 2 2B & 1320/1320 & 80/80 & 1574/1600 & \cellcolor{orange!20}98.38 \\
Gemma 3 4B (text) & 1600/1600 & 240/240 & 1584/1600 & 99.00 \\
Qwen3.6-27B & \cellcolor{orange!20}1312/1320 & \cellcolor{orange!20}70/80 & 1584/1600 & 99.00 \\
Llama 3.2-1B & 1600/1600 & 240/240 & 1589/1600 & 99.31 \\
\bottomrule
\end{tabular*}}
\par\smallskip{\footnotesize\raggedright Orange marks incomplete numerical exactness or fidelity below 99\%. Exactness counts test cases; fidelity counts top-1 decisions on common prefixes, which are correlated within a sequence. Gemma 3 4B and Llama 3.2-1B use the 32-step suites of the correctness track; Gemma 2 2B and Qwen3.6-27B use the earlier eight-step suites. Qwen3.5 tests: Appendix~\ref{app:tests}; additional decoders: Appendix~\ref{app:extended_models}.\par}
\end{table}
\fi
To test whether the contract and evaluator carry to decoders outside the benchmark, we built reference designs for the attention-only Gemma 2 2B~\citep{gemma2card} and the 64-layer hybrid Qwen3.6-27B~\citep{qwen36config} (Appendix~\ref{app:extended_models}). \apx{}Table~\ref{tab:extended_reference} compares them on exactness and fidelity with four references that qualify: Qwen3.5-0.8B and 4B, and the correctness-track targets Gemma 3 4B and Llama 3.2-1B (Section~\ref{sec:tasks}). Both meet timing (Table~\ref{tab:T09-gates}), and each fails a different criterion. Gemma 2 2B's RTL matches its contract on every case, but the contract agrees with FP32 on only 98.38\% of decisions.\ifcompact\else{} Seven of eight alternative integer contracts reach the threshold, and higher precision costs hardware: LUTs rise from 58.4k to 84.1k and throughput falls from 1.48 to 0.90 tokens/s (Appendix~\ref{app:add-reference}).\fi{} Qwen3.6-27B's contract meets the threshold, but its RTL fails eight public and ten held-out cases even though every operator and layer check passes, so the defects appear only when layers are integrated. \ifcompact Two later revisions still fail one or two cases, so Qwen3.6-27B does not qualify.\else Two later revisions reduce the failures to one or two cases; neither passes every check, so Qwen3.6-27B does not qualify.\fi

\section{Findings}
\label{sec:findings}
\label{sec:failures}
\begin{figure}[tbp]
\centering
\includegraphics[width=\linewidth]{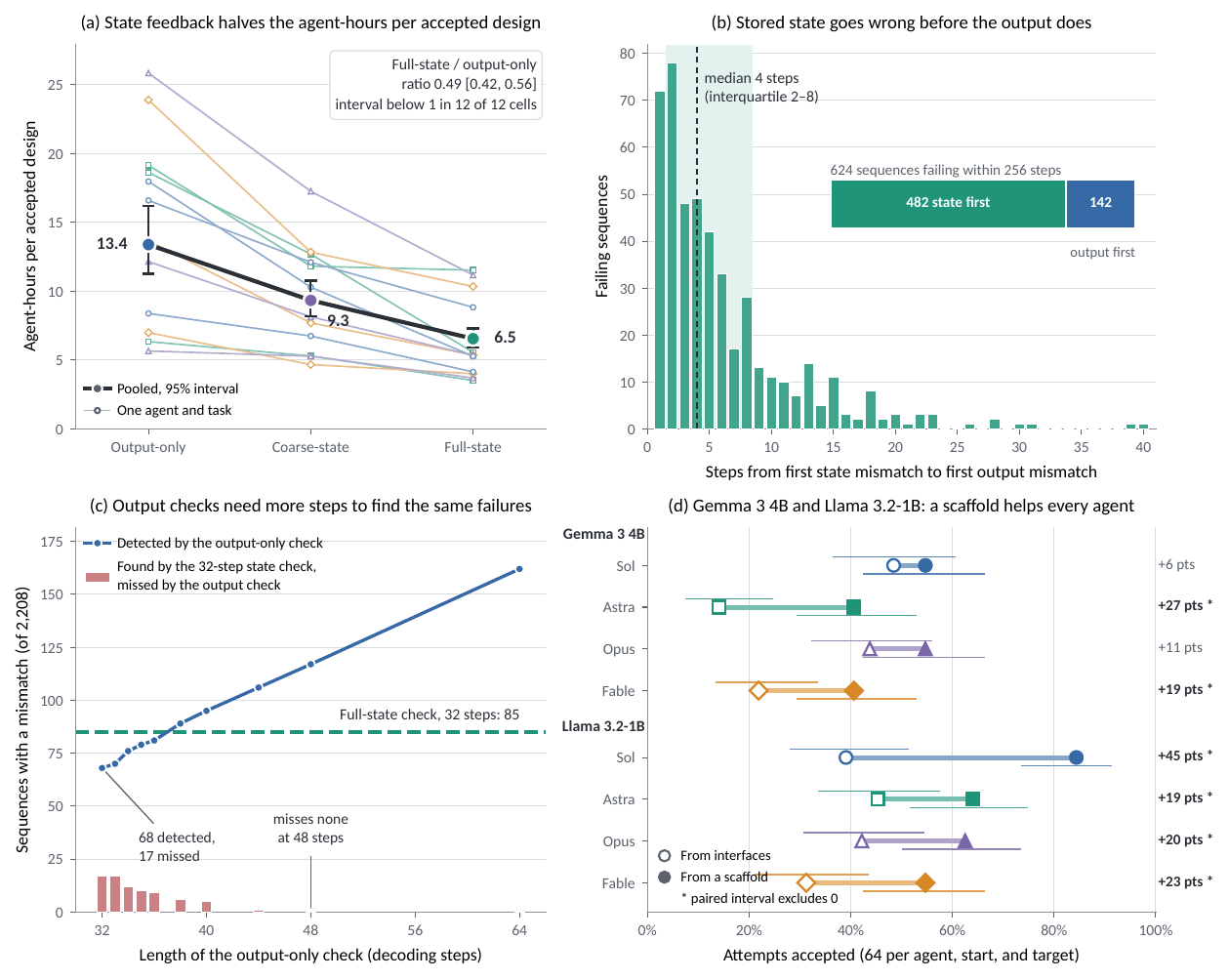}
\caption{\textbf{State feedback, state-first failures, and the starting workspace.} (a) Agent-hours per accepted design in the feedback study: pooled over the 384 attempts of each setting with block-bootstrap 95\% intervals (dark line), and for each of the 12 agent--task cells (thin lines; color and marker identify the agent). (b) Regrading of the 69 accepted submissions: steps from the first state mismatch to the first output mismatch for the 470 state-first sequences whose output also fails within 256 steps; 12 more state-first sequences show no output mismatch by step 256. (c) Sequences with a mismatch when the output-only check runs for 32 to 64 steps, against the 85 that the full-state check finds in 32 steps; bars count sequences that the state check finds and the output check misses. (d) Acceptance on the four Gemma 3 4B and Llama 3.2-1B tasks; thin lines above and below each row are the Wilson 95\% intervals from interfaces and from the scaffold. Labels give the paired difference; * marks a paired interval that excludes zero.}
\label{fig:key_state}
\end{figure}

Five findings recur across the two tracks and the diagnostic studies (\ifcompact Figure~\ref{fig:key_main}\else Figure~\ref{fig:key_state}\fi).

\subsection{Agents adapt a scaffold more reliably than they build a full-size decoder from interfaces}
\label{sec:remaining}
\begin{table}[tb]
\centering
\caption{\textbf{Acceptance by task: building Gemma 4 E4B and Llama 3.2 3B from interfaces is hardest.} Accepted/attempts per agent and task; the last row pools the four agents.}
\label{tab:leaderboard_tasks}
{\footnotesize\tablefont\setlength{\tabcolsep}{1.5pt}\resultrows
\begin{tabular*}{\linewidth}{@{\extracolsep{\fill}}l*{12}{c}@{}}
\toprule
 & \multicolumn{6}{c}{Construction from interfaces} & \multicolumn{6}{c}{Scaffold adaptation} \\
\cmidrule(lr){2-7}\cmidrule(lr){8-13}
Agent & {\resultstack\begin{tabular}[b]{@{}c@{}}Nano\\T2\end{tabular}} & {\resultstack\begin{tabular}[b]{@{}c@{}}Nano\\T3\end{tabular}} & {\resultstack\begin{tabular}[b]{@{}c@{}}Gemma\\4 E4B\end{tabular}} & {\resultstack\begin{tabular}[b]{@{}c@{}}Llama\\3.2 3B\end{tabular}} & {\resultstack\begin{tabular}[b]{@{}c@{}}Gemma\\3 4B\end{tabular}} & {\resultstack\begin{tabular}[b]{@{}c@{}}Llama\\3.2-1B\end{tabular}} & {\resultstack\begin{tabular}[b]{@{}c@{}}Qwen3.5\\0.8B\end{tabular}} & {\resultstack\begin{tabular}[b]{@{}c@{}}Qwen3.5\\4B\end{tabular}} & {\resultstack\begin{tabular}[b]{@{}c@{}}Gemma\\4 E4B\end{tabular}} & {\resultstack\begin{tabular}[b]{@{}c@{}}Llama\\3.2 3B\end{tabular}} & {\resultstack\begin{tabular}[b]{@{}c@{}}Gemma\\3 4B\end{tabular}} & {\resultstack\begin{tabular}[b]{@{}c@{}}Llama\\3.2-1B\end{tabular}} \\
\midrule
\textbf{Fable 5.1} & \cellcolor{blueC!40}25/32 & \cellcolor{blueC!37}23/32 & \cellcolor{blueC!16}6/24 & \cellcolor{blueC!23}10/24 & \cellcolor{blueC!14}14/64 & \cellcolor{blueC!19}20/64 & \cellcolor{blueC!40}25/32 & \cellcolor{blueC!44}21/24 & \cellcolor{blueC!21}9/24 & \cellcolor{blueC!31}14/24 & \cellcolor{blueC!23}26/64 & \cellcolor{blueC!29}35/64 \\
\textbf{Sol 5.6} & \cellcolor{blueC!38}24/32 & \cellcolor{blueC!35}22/32 & \cellcolor{blueC!21}9/24 & \cellcolor{blueC!18}7/24 & \cellcolor{blueC!26}31/64 & \cellcolor{blueC!22}25/64 & \cellcolor{blueC!45}29/32 & \cellcolor{blueC!42}20/24 & \cellcolor{blueC!35}16/24 & \cellcolor{blueC!36}17/24 & \cellcolor{blueC!29}35/64 & \cellcolor{blueC!42}54/64 \\
\textbf{Astra 6} & \cellcolor{blueC!44}28/32 & \cellcolor{blueC!35}22/32 & \cellcolor{blueC!10}3/24 & \cellcolor{blueC!14}5/24 & \cellcolor{blueC!11}9/64 & \cellcolor{blueC!25}29/64 & \cellcolor{blueC!38}24/32 & \cellcolor{blueC!44}21/24 & \cellcolor{blueC!27}12/24 & \cellcolor{blueC!33}15/24 & \cellcolor{blueC!23}26/64 & \cellcolor{blueC!33}41/64 \\
\textbf{Opus 5.5} & \cellcolor{blueC!41}26/32 & \cellcolor{blueC!38}24/32 & \cellcolor{blueC!18}7/24 & \cellcolor{blueC!21}9/24 & \cellcolor{blueC!24}28/64 & \cellcolor{blueC!23}27/64 & \cellcolor{blueC!37}23/32 & \cellcolor{blueC!27}12/24 & \cellcolor{blueC!36}17/24 & \cellcolor{blueC!35}16/24 & \cellcolor{blueC!29}35/64 & \cellcolor{blueC!33}40/64 \\
\midrule
\textbf{Pooled} & \cellcolor{blueC!41}80.5\% & \cellcolor{blueC!36}71.1\% & \cellcolor{blueC!16}26.0\% & \cellcolor{blueC!19}32.3\% & \cellcolor{blueC!19}32.0\% & \cellcolor{blueC!22}39.5\% & \cellcolor{blueC!40}78.9\% & \cellcolor{blueC!39}77.1\% & \cellcolor{blueC!30}56.3\% & \cellcolor{blueC!34}64.6\% & \cellcolor{blueC!26}47.7\% & \cellcolor{blueC!34}66.4\% \\
\bottomrule
\end{tabular*}
}
\par\smallskip{\footnotesize\raggedright Nano T2, nano T3, and Qwen3.5-0.8B come from the full-state setting of the feedback study (32 attempts per agent and task). Qwen3.5-4B, Gemma 4 E4B, and Llama 3.2 3B have 24 attempts per agent and task, and Gemma 3 4B and Llama 3.2-1B have 64. Per-cell intervals and costs: Tables~\ref{tab:feedback_repeated}, \ref{tab:T03-adaptation}--\ref{tab:T03-cost}, and~\ref{tab:T10_gemma3_llama1b}.\par}
\end{table}

Acceptance depends strongly on the starting workspace (Table~\ref{tab:leaderboard_tasks}). Pooled over agents, 47.7 to 78.9\% of scaffold-adaptation attempts are accepted. So are 80.5\% (T2) and 71.1\% (T3) of attempts on the four-layer nano tasks, which start from interfaces. Building a full-size decoder from interfaces succeeds less often: on 26.0\% of attempts for Gemma 4 E4B, 32.3\% for Llama 3.2 3B, 32.0\% for Gemma 3 4B, and 39.5\% for Llama 3.2-1B, and no agent exceeds 10 of 24 on either of the first two. On these four targets every agent does better from the scaffold, 408 of 704 attempts against 239, and construction from interfaces costs 46 to 184 agent-hours per accepted design (Appendices~\ref{sec:cross_architecture} and~\ref{app:gemma3_llama1b}).\ifcompact\else{} On Gemma 3 4B and Llama 3.2-1B, where each agent attempts 64 matched instances from each start, six of the eight paired differences between starts have intervals that exclude zero (Figure~\ref{fig:key_state}d). Appendix~\ref{app:add_target} lists the steps that turn a decoder into a task.\fi

Agents also differ on single tasks. Opus 5.5 adapts Qwen3.5-4B in 12 of 24 attempts against 20 or 21 for the others, and Astra 6 and Fable 5.1 build Gemma 3 4B from interfaces in 9 and 14 of 64 attempts against 28 and 31 for Opus 5.5 and Sol 5.6; matched tests confirm both differences (Holm-adjusted $p=0.015$ and $p<0.001$; Appendix~\ref{app:add-agents}).\ifcompact\else{} Astra 6 builds Gemma 4 E4B from interfaces in 3 of 24.\fi\ifcompact\else{} The ranking also changes between experiments: on the controlled matrix of Section~\ref{sec:state_budget}, Fable 5.1 has 445 of 864 attempts accepted and Sol 5.6, first on the correctness track, has 385 ($p=0.001$).\fi\ifcompact\else{} Outcomes also vary between attempts on one instance. With three attempts on each of eight Gemma 4 E4B and eight Llama 3.2 3B adaptation instances, the three outcomes disagree in 26 of 64 agent--instance groups (Appendix~\ref{app:add-cross}).\fi

\subsection{Agents do much better when told where state diverges}
\label{sec:res_feedback}
On nano T2, nano T3, and Qwen3.5-0.8B, each agent attempted 32 task instances under each feedback setting; the three attempts on one instance form a matched \emph{block}. Full-state feedback raises acceptance over output-only feedback in all 12 agent--task cells, by 25.0 to 46.9 points, and every block-bootstrap interval excludes zero (Appendix Table~\ref{tab:feedback_repeated}). Pooled, 144, 208, and 295 of 384 attempts are accepted under output-only, coarse-state, and full-state feedback, with nearly equal total agent time, so the cost per accepted design falls from 13.4 [11.3, 16.2] to 6.5 [5.9, 7.3] agent-hours, a ratio of 0.49 [0.42, 0.56]. Output-only and full-state feedback differ both in whether state is visible and in whether the mismatch is located. Coarse-state feedback reports a state mismatch without locating it, and full-state feedback exceeds it in all 12 cells, by 9.4 to 37.5 points, with intervals that exclude zero in seven\ifcompact\else. Removing any one of the 96 matched blocks keeps the pooled ratio between 0.48 and 0.50, and in each of the 12 agent--task cells the interval of the full-state ratio lies below one\fi{} (\ifcompact Figure~\ref{fig:key_main}a\else Figure~\ref{fig:key_state}a\fi, Appendix~\ref{app:add-feedback}).\ifcompact\else{} Accepted designs also arrive earlier: by half the wall cap, 82 of 384 full-state attempts had submitted a design that was later accepted, against 67 under coarse-state and 52 under output-only feedback (Appendix~\ref{app:add-outcomes}).\fi

Comparing the two state settings separates finding a state defect from repairing it. Among attempts with a state defect, agents find it slightly more often under full-state feedback, in 151 of 166 after a median of 1.35 agent-hours, than when told only that state disagrees, in 142 of 169 after 1.50 hours. Repair differs more: agents repair the defect in 123 of the 166 under full-state feedback, after a median of 2.69 hours, and in 87 of the 169 under coarse-state feedback, after 3.22 hours (Table~\ref{tab:T05-resources}).\ifcompact\else{} On the constructed decoder of Section~\ref{sec:bg_decoding}, one Sol 5.6 repair under each setting shows what the agent is told. Full-state feedback reports a cache valid count of 4 at transaction 2, where the oracle expects 2; output-only feedback reports logits $[12,12,-6,-5]$ at transaction 5, where the oracle expects $[13,12,-6,-6]$. Both runs make the same one-line change, committing the cache entry only when the output transfer is accepted, and pass the held-out suite (Appendix~\ref{app:debug_trace}).\fi

\ifcompact\else The gain splits between reporting that state disagrees and locating where. Full-state feedback adds 151 pooled acceptances over output-only feedback; 64 arise between output-only and coarse-state feedback and 87 between coarse-state and full-state feedback. Both increments are positive in every cell. Full-state exceeds coarse-state feedback by 9.4 to 37.5 points, with intervals that exclude zero in seven of the 12 cells; for the gain of coarse-state over output-only feedback, the intervals exclude zero in only two cells (Table~\ref{tab:T05-contrasts}).
\fi

\subsection{Passing a test suite does not establish exactness}
\label{sec:evaluator_results}
On the correctness track, 148 designs that pass every public test fail final acceptance, 115 of them on the held-out suite: 25 for Sol, 39 for Opus, 33 for Fable, and 51 for Astra. For the 36 such designs on Gemma 4 E4B, Llama 3.2 3B, and Qwen3.5-4B, the grader's primary diagnoses are 13 integration, 13 state/protocol, and 10 arithmetic (Appendix~\ref{app:add-cross}). The gap between what the agent sees pass and what is accepted is wider under output-only feedback: of 384 output-only attempts, 219 submit a design that passes every check the agent can see, 199 also pass the full public suite, and 144 are accepted.

Accepted designs can also fail under conditions their tests did not cover. We regraded 144 frozen submissions on Qwen3.5-0.8B, Gemma 4 E4B, and Llama 3.2 3B, 69 of which had been accepted, together with the three reference designs. Each design decodes eight new token prefixes under each of four conditions (a new prefix alone, nonzero cache positions, reset after activity, and bounded memory stalls) for up to 256 steps, which gives 32 sequences per design. The references stay exact throughout. Of the 69 accepted submissions, 45 fail at least one sequence within the 32 steps their tests required, although 2,123 of their 2,208 sequences stay exact. The failures concentrate in the conditions that exercise state: of the 85 failing sequences, \ifcompact 76 fail under stalls, reset after activity, or nonzero cache positions\else 29 fail under bounded memory stalls, 28 under reset after activity, 19 at nonzero cache positions,\fi{} and only 9 on a new prefix alone. At 256 steps,\ifcompact\else{} stalls remain the hardest condition, with 60.6 to 69.9\% of sequences exact per target against 75.6 to 86.1\% on new prefixes alone (Table~\ref{tab:T07-conditions}), and\fi{} only 20 of the 69 are exact on all eight new prefixes alone, and 482 of the 624 failing sequences first diverge in persistent state, a median of four steps before the output does (\ifcompact Figure~\ref{fig:key_main}b\else Figure~\ref{fig:key_state}b\fi). Within the required 32 steps, 17 of the 85 failing sequences\ifcompact\else, and 5 of the 45 failing designs,\fi{} are wrong in state only, so an output test of that length would pass them. \ifcompact Output tests extended to 48 steps miss none of them (Appendix~\ref{app:add-horizon}).\else Longer output tests find these defects later: extended to 35 steps, output checks detect 79 sequences and miss 10 that the 32-step state check finds, and at 48 steps they miss none (Figure~\ref{fig:key_state}c, Appendix~\ref{app:add-horizon}).\fi{}

\paragraph{Known defects.}
\label{sec:evaluator_defects}
On implementations with known defects, the complete evaluator rejects all 12 development defects and 43 of 48 held-out defects\ifcompact\else{} (89.6\%)\fi; the public tests alone reject 3 of the 12 and 3 of the 48. It accepts every repaired counterpart and correct control (Appendices~\ref{app:gate_analysis} and~\ref{app:add-faults}).

\subsection{Longer horizons lower acceptance, and the budget binds}
\label{sec:state_budget}
\ifcompact\else\ifcompact\begin{table}[htbp]\else\begin{table}[tb]\fi
\centering
\caption{\textbf{Acceptance falls as the required decoding horizon grows, at both budgets.} Final acceptance by horizon and budget multiplier in the controlled matrix, pooled over four agents and three added state conditions. Every attempt starts from the same Qwen3.5-nano scaffold.}
\label{tab:state_budget}
{\small\tablefont\setlength{\tabcolsep}{11pt}\resultrows
\begin{tabular}{@{}l*{4}{c}}
\toprule
Budget & {\resultstack\begin{tabular}[b]{@{}c@{}}1 step\end{tabular}} & {\resultstack\begin{tabular}[b]{@{}c@{}}4 steps\end{tabular}} & {\resultstack\begin{tabular}[b]{@{}c@{}}8 steps\end{tabular}} & {\resultstack\begin{tabular}[b]{@{}c@{}}16 steps\end{tabular}} \\
\midrule
$1\times$ & \cellcolor{blueC!44}{\resultstack\begin{tabular}[b]{@{}c@{}}\textbf{87.5\%}\\{\scriptsize 252/288}\end{tabular}} & \cellcolor{blueC!34}{\resultstack\begin{tabular}[b]{@{}c@{}}\textbf{64.6\%}\\{\scriptsize 186/288}\end{tabular}} & \cellcolor{blueC!30}{\resultstack\begin{tabular}[b]{@{}c@{}}\textbf{56.3\%}\\{\scriptsize 162/288}\end{tabular}} & \cellcolor{blueC!22}{\resultstack\begin{tabular}[b]{@{}c@{}}\textbf{39.9\%}\\{\scriptsize 115/288}\end{tabular}} \\
$1.5\times$ & \cellcolor{blueC!43}{\resultstack\begin{tabular}[b]{@{}c@{}}\textbf{85.8\%}\\{\scriptsize 247/288}\end{tabular}} & \cellcolor{blueC!38}{\resultstack\begin{tabular}[b]{@{}c@{}}\textbf{74.3\%}\\{\scriptsize 214/288}\end{tabular}} & \cellcolor{blueC!36}{\resultstack\begin{tabular}[b]{@{}c@{}}\textbf{68.8\%}\\{\scriptsize 198/288}\end{tabular}} & \cellcolor{blueC!31}{\resultstack\begin{tabular}[b]{@{}c@{}}\textbf{58.7\%}\\{\scriptsize 169/288}\end{tabular}} \\
\bottomrule
\end{tabular}
}
\par\smallskip{\footnotesize\raggedright Each cell has 288 attempts: 4 agents $\times$ 3 state conditions (none, reset after activity, bounded stalls) $\times$ 24 attempts. A budget multiplier scales the wall-time, tool-call, and output-token caps together; the 0.5$\times$ budget, at which most attempts end at a cap, is omitted here and described in the text of Appendix~\ref{app:add-state}. Public and held-out suites grow from 984 and 68 cases at one step to 1,344 and 128 at 16, so a longer horizon also adds decoding work and test cases. Per-agent and per-condition results: Tables~\ref{tab:T06-horizon}--\ref{tab:T06-effects}.\par}
\end{table}
\fi
The controlled matrix starts every attempt from one Qwen3.5-nano scaffold and varies the required decoding steps (1 to 16), an added state condition (none, reset after activity, or bounded memory stalls), and a multiplier on all three caps. At the 1$\times$ budget, pooled acceptance falls from 252 to 115 of 288 attempts between one and 16 required steps, and with no added condition every agent has fewer attempts accepted at 16 steps than at one (\ifcompact Appendix~\ref{app:add-state}, \fi Table~\ref{tab:state_budget}). A 1.5$\times$ budget lifts the 16-step result to 169 of 288, but acceptance still falls at each longer horizon. On the Gemma 3 4B and Llama 3.2-1B tasks, the same instances at eight and at 32 required steps have 558 and 475 of 1,024 attempts accepted (Appendix~\ref{app:gemma3_llama1b}). \ifcompact\else How an added state condition changes the horizon effect depends on the budget. At 0.5$\times$, where at most 1 of 96 attempts per cell is accepted at 16 steps, the interaction is positive because acceptance has little room to fall; at 1$\times$ it is negative for reset after activity, and the other intervals include zero (Appendix~\ref{app:add-state}).\fi

The budget binds on the correctness track as well: 489 of 1,888 attempts end at a cap without submitting (322 on wall time, 105 on tool calls, and 62 on output tokens). Under full-state feedback, caps account for about half of the failures that remain: on the three feedback tasks, 43 of the 89 failed full-state attempts end at a cap, and 39 of these 43 had no state defect or had already repaired it. Caps also end 260 of the 465 failed attempts to build a full-size decoder from interfaces.

\subsection{Optimization often ends without a correct design}
In the PPA track, 571 of 1,536 attempts end without a correct design, 90 to 185 of 384 per agent: 194 submit an incorrect design, and 377 reach a cap with RTL that fails the correctness checks. Every one of these attempts started from a design that passed every gate\ifcompact{} (Appendix Figure~\ref{fig:failure_records}b).\else. In one such attempt (Figure~\ref{fig:failure_records}b), Sol 5.6 starts from the correct 0.8B design at 3.999 tokens/s and stops at the tool-call cap with a design that fails the correctness checks and runs at 3.803 tokens/s.\fi

\ifcompact\else\ifcompact\begin{figure}[htbp]\else\begin{figure}[tb]\fi
\centering
\newcommand{\recordhead}[2]{{\footnotesize\textbf{(#1)}~#2\par}\vspace{2pt}}
\newcommand{\recordrule}{\noalign{\vskip1pt\hrule\vskip1pt}}
\newcommand{\recordtab}{\scriptsize\setlength{\tabcolsep}{0pt}\renewcommand{\arraystretch}{1.05}}
\begin{minipage}[t]{.31\linewidth}
\recordhead{a}{Sol 5.6, T3, full-state}
{\recordtab\begin{tabular*}{\linewidth}{@{}r@{\extracolsep{\fill}}rl@{}}
Hours & Calls & \hspace{1em}Event\\ \recordrule
0.00 & 0 & \hspace{1em}start\\
2.86 & 78 & \hspace{1em}state defect found\\
4.25 & 116 & \hspace{1em}state defect repaired\\
6.00 & 164 & \hspace{1em}wall cap; not accepted\\
\end{tabular*}}
\end{minipage}\hfill
\begin{minipage}[t]{.31\linewidth}
\recordhead{b}{Sol 5.6, 0.8B optimization}
{\recordtab\begin{tabular*}{\linewidth}{@{}r@{\extracolsep{\fill}}rl@{}}
Hours & Calls & \hspace{1em}Event\\ \recordrule
0.00 & 0 & \hspace{1em}correct; 3.999 tokens/s\\
8.71 & 420 & \hspace{1em}tool cap\\
 & & \hspace{1em}correctness checks fail\\
 & & \hspace{1em}3.803 tokens/s\\
\end{tabular*}}
\end{minipage}\hfill
\begin{minipage}[t]{.31\linewidth}
\recordhead{c}{Sol 5.6, T3, output-only}
{\recordtab\begin{tabular*}{\linewidth}{@{}r@{\extracolsep{\fill}}rl@{}}
Hours & Calls & \hspace{1em}Event\\ \recordrule
5.12 & 118 & \hspace{1em}submitted\\
 & & \hspace{1em}visible checks 1384/1384\\
5.19 & & \hspace{1em}public suite 1378/1384\\
 & & \hspace{1em}held-out suite 95/96\\
 & & \hspace{1em}not accepted\\
\end{tabular*}}
\end{minipage}
\caption{Excerpts from three attempt records, with hours since the start of the attempt and tool calls. (a) A full-state attempt finds and repairs its state defect and then reaches the 6-hour wall cap. (b) An optimization attempt starts from the correct reference design and reaches the 420-call tool cap with RTL that fails the correctness checks. (c) An output-only submission passes every check the agent can see and, after the RTL is frozen, fails the full public suite and the held-out suite (Listing~\ref{lst:sa-output-only}).}
\label{fig:failure_records}
\end{figure}
\fi

\section{Related Work}
\label{sec:related}
\paragraph{Where AIChipBench sits in chip design.}
A digital design passes through several stages: a specification and an architecture; RTL implementation; functional verification; synthesis onto hardware resources; physical implementation, which places and routes the design and analyzes its timing and power; and finally deployment on an FPGA or fabrication as an application-specific integrated circuit (ASIC). Language models have been applied at several of these stages, including RTL generation, verification, EDA scripting, tool orchestration~\citep{chipnemo,chateda}, and the full path from a workload specification to a routed layout~\citep{buildachip}. Given a target decoder and a fixed module structure, AIChipBench's agents write, debug, and optimize RTL; the evaluator runs verification and synthesis, and placement and routing follow submission (Section~\ref{sec:physical_flow}). Architecture design and high-level synthesis lie upstream.

\paragraph{Benchmarks.}
VerilogEval~\citep{liu2023verilogeval}, built from HDLBits~\citep{hdlbits} problems, RTLLM~\citep{rtllm} with its 2.0 release~\citep{openllm}, TuRTLe~\citep{turtle}, and RTL-Repo~\citep{rtlrepo} ask a model to generate or complete a module or design and simulate it against references; RTLLM and TuRTLe also score synthesis or PPA. ChipBench~\citep{chipbench} adds hierarchical and CPU-component generation, debugging, and reference-model generation. CVDP~\citep{cvdp} covers design, verification, debugging, and comprehension, and in its agentic tasks an agent works with tools in a repository before hidden harnesses grade the result; LBC-bench~\citep{lbcbench} reports model and agent results on CVDP tasks. Build-A-Chip-Bench~\citep{buildachip}, described in a blog post, carries agents through architecture and physical design to an ASIC layout scored on PPA. Appendix Table~\ref{tab:related} compares what each benchmark grades. AIChipBench shares executable tools and hidden grading with the agentic benchmarks. Within a fixed module structure, it measures state-exact correctness of decoder accelerators over decoding steps, the effect of what debugging feedback reveals, and each agent's reliability and cost over repeated budgeted attempts. Benchmark sizes count problems, designs, examples, or tasks, units comparable neither across benchmarks nor with agent attempts (Appendix~\ref{app:related_sizes}). The closest feedback study is ChipBench's, which gives models a buggy design with or without simulation waveforms (signals and internal state over time); the waveforms improve results for 8 models and lower them for 13. It differs from ours in information and protocol, so the results are not directly comparable.\ifcompact\else{} Architect Labs~\citep{architect} studies RTL generation with verifiable rewards.\fi

\ifcompact\else
\paragraph{Verification across representations.}
EDASYTH~\citep{edasyth} gates transitions among symbolic calculations, simulation, and circuit artifacts, EEBench~\citep{eebench} evaluates electrical engineering designs through simulation and cost, and the large-circuit-model perspective~\citep{lcm} studies learning across specifications, RTL, netlists, and layouts. AIChipBench checks one transition, from a quantized model to RTL, against an executable state-transition contract.

\fi

\section{Conclusion}
\label{sec:mainend}
AIChipBench tests whether LLM coding agents can build decoder hardware that stays exact, in outputs and committed state, over many decoding steps, and whether they can make a correct design faster without breaking it. Its results locate the difficulty in stored state. When accepted designs fail under conditions their tests did not exercise, most of the failing sequences diverge in state before the output does, and output checks of the required length miss defects that a state check exposes. Agents, in turn, succeed far more often when they can see that state: feedback that locates the first state mismatch doubles pooled acceptance over output-only feedback, after which caps account for about half of the remaining failures. Correctness is also fragile under optimization, since more than a third of attempts to speed up a correct design end without one. An overall score ranks Sol 5.6 and Fable 5.1 above Astra 6 and Opus 5.5, and the order holds across random halves of the instances. Debugging tools for hardware agents should therefore report stored state, and final evaluations should test well beyond the development cases.

\paragraph{Limitations.} Tasks fix the module structure, and agents do not place and route their designs. Graded horizons stop at 32 steps and regrading at 256. Whether feedback on stored state and held-out state conditions helps on other stateful hardware remains to be tested.

\vspace{10ex}

\bibliography{references}
\bibliographystyle{plainnat}
\clearpage
\appendix
\renewcommand{\appendixpagename}{\centering\huge Appendix}
\appendixpage
\counterwithin{theorem}{section}

\startcontents[section]
\begingroup
\small
\setlength{\parskip}{0pt}
\printcontents[section]{l}{1}{\setcounter{tocdepth}{2}}
\endgroup
\clearpage
\section{Benchmark protocol and attempt records}
\label{app:protocol}
Two independently versioned objects identify a reported result: a workload contract and an agent configuration. The manifest fields in Table~\ref{tab:manifest} determine which comparison a reported result supports. Table~\ref{tab:experiments} lists the agent and benchmark studies, the conditions each varies, and the sections that report them, and Figure~\ref{fig:architecture} shows the workflow of one attempt.

\begin{table}[tbp]
\centering
\caption{\textbf{Agent and benchmark studies.} Each agent study varies one set of conditions with the others fixed; the benchmark studies test the evaluator and the reference designs.}
\label{tab:experiments}
{\small\tablefont\setlength{\tabcolsep}{3pt}\resultrows
\begin{tabularx}{\linewidth}{@{}l>{\raggedright\arraybackslash}Xllrl@{}}
\toprule
Study & Varies & Design & Unit & Attempts & Results \\
\midrule
Earlier task series & Task (T1, T2, T3, 0.8B) & 4 agents $\times$ 4 tasks & Attempt & 240 & App.~\ref{sec:agents} \\
Controlled matrix & Horizon, state condition, budget & 4 agents $\times$ 36 cells $\times$ 24 blocks & Attempt & 3,456 & \ref{sec:state_budget} \\
Feedback study & Feedback (3 settings) & 4 agents $\times$ 3 tasks $\times$ 3 settings $\times$ 32 & Block & 1,152 & \ref{sec:res_feedback} \\
Sol feedback pilot & Feedback (2 settings) & Sol 5.6 $\times$ 2 tasks $\times$ 16 pairs & Pair & 32 & App.~\ref{app:sa-comparison} \\
Cross-architecture & Target and starting files & 4 agents $\times$ 5 cells $\times$ 24 & Attempt & 480 & App.~\ref{sec:cross_architecture} \\
Repeated instances & Repeat of one instance & 4 agents $\times$ 16 instances $\times$ 3 & Group & 192 & App.~\ref{app:add-cross} \\
PPA track & Improvement of a correct design & 4 agents $\times$ 4 targets $\times$ 96 & Attempt & 1,536 & \ref{sec:hardware}, \ref{app:add-optimization} \\
\begin{tabular}[t]{@{}l@{}}Gemma 3 4B,\\Llama 3.2-1B\end{tabular} & Target and starting files & 4 agents $\times$ 4 tasks $\times$ 64 & Block & 1,024 & App.~\ref{app:gemma3_llama1b} \\
Horizon comparison & Required decoding steps (8, 32) & 4 agents $\times$ 4 tasks $\times$ 64 & Block & 1,024 & App.~\ref{app:gemma3_llama1b} \\
\midrule
\multicolumn{4}{@{}l}{\textbf{All agent attempts}} & \textbf{9,136} & \\
\midrule
\multicolumn{6}{@{}l}{\textbf{Benchmark studies (no further agent attempts)}} \\
Known defects & Injected RTL defect & 12 development, 48 held-out & Defect & -- & \ref{sec:evaluator_defects} \\
Regrading & New prefixes and conditions & 144 submissions, 3 references & Sequence & -- & \ref{sec:evaluator_results} \\
Reference workloads & Target decoder & 6 reference designs & Test case & -- & \ref{sec:correctness} \\
Reference hardware & Schedule (0.8B, 4B); target (the other five) & 7 reference designs & Design & -- & \ref{sec:hardware} \\
Routed agent designs & Place and route; board power & 965 PPA, 144 correctness & Design & -- & \ref{sec:hardware} \\
Constructed decoder & Seeded commit defect; scripted repairs & 4 snapshots, 9 cases & Transaction & -- & App.~\ref{app:decoder_case} \\
\bottomrule
\end{tabularx}
}
\par\smallskip{\footnotesize\raggedright The unit is what a study's rates count or its intervals resample (block: matched instance block; pair: paired attempt; group: agent--instance group). The earlier task series has 16 attempts per agent on each nano task and 12 on 0.8B. The controlled matrix varies the horizon (1--16 steps), added state condition, and budget (0.5--1.5$\times$) from one scaffold; its blocks pair attempts across state conditions. The feedback study compares output-only, coarse-state, and full-state feedback on T2, T3, and 0.8B, and the Sol pilot compares output-only with full-state feedback on T2 and T3. The cross-architecture cells are Gemma 4 E4B and Llama 3.2 3B from either start and Qwen3.5-4B from a scaffold. Repeated instances cover 8 adaptation instances each on Gemma 4 E4B and Llama 3.2 3B. The Gemma 3 4B and Llama 3.2-1B tasks use 32 decoding steps; their blocks pair the four agents on one instance, and the horizon comparison reruns the same instances at eight steps. The correctness track is the full-state setting of the feedback study, the cross-architecture cells, and the Gemma 3 4B and Llama 3.2-1B tasks: 472 attempts per agent. Regrading decodes up to 256 steps. The Sol 5.6 and Astra 6 attempts on the earlier task series and the Sol pilot total 152 attempts, with the pilot's full-state attempts counted once. Counts by study: Table~\ref{tab:additional_attempts}.\par}
\end{table}

\begin{figure}[htbp]
  \centering
  \includegraphics[width=.9\linewidth]{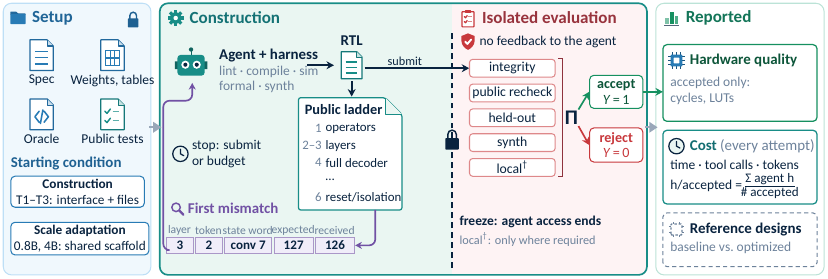}
  \caption{Workflow of one attempt. The agent iterates against public tests until it submits or reaches a cap. Its RTL is then frozen and evaluated in isolation; held-out results never return to the agent. The harness records cost over all attempts.}
  \label{fig:architecture}
\end{figure}

\begin{table}[ht]
\centering\small
\caption{\textbf{Manifest fields required for every attempt.}}
\label{tab:manifest}
\tablefont
\begin{tabularx}{\linewidth}{lX}\toprule
Field & Meaning and purpose \\\midrule
Task revision & Specification, public tests, held-out-test policy and RTL interface version \\
Model revision & Exact checkpoint and configuration; tokenizer and export software versions \\
Arithmetic digest & Bit widths, scales, tile order, nonlinear tables, rounding and saturation \\
Starting point & Construction from interfaces or scaffold adaptation; files available at time zero \\
Agent configuration & Provider model identifier, sampling settings, prompt, harness revision \\
Caps & Wall time, generated tokens, tool/evaluator calls and timeout policy \\
Execution environment & CPU, RAM, operating system, simulator, compiler, synthesis version \\
Hardware constraints & Device/part, clocks, memory interface, resource caps, I/O exclusions \\
Isolation & Held-out assets unavailable during generation; protected files read-only \\
Output identity & Hashes of submitted RTL, reports, and other output files after agent exit \\
Outcome & Each gate separately, first mismatch, cost and termination reason \\
\bottomrule\end{tabularx}
\end{table}

\paragraph{Arithmetic versions.}
The arithmetic digest in Table~\ref{tab:manifest} identifies one version of a numerical contract. A change in precision, a new nonlinear table, or a different accumulation order requires a new arithmetic version, even if next-token agreement improves. The baseline and optimized schedules of a reference design (Section~\ref{sec:hardware}) share one version: they use the same integer functions, parameter arrays, sequence lengths, and external-memory behavior, and both convert cycles to throughput at 126\,MHz. Their RTL may differ in buffering, overlap, and scheduling; approximations and matrix dimensions stay fixed.

\subsection{Task configuration and execution records}
\label{app:bindings}

\paragraph{Starting files and assistance.}
The starting workspace sets how much the agent must implement, so it is part of the task definition.
For construction from interfaces, it holds the supplied module declarations, testbench interfaces, arithmetic components, and any working RTL; for scaffold adaptation, it also holds the initial accelerator tree and the configuration that tree supports.
Agents can query the reference implementation of the numerical contract through the permitted oracle API but cannot read its source.
So that a starting workspace can be reproduced, the protocol requires an inventory, taken before the attempt, of every path the agent can read or edit, including any reference implementation it can access. Separate digests identify the public tests, oracle code, arithmetic tables, held-out evaluator assets, and hardware constraints.
Within a task, all agents start from the same implementation.
A change to the supplied implementation defines a different starting workspace, even when the instruction text is unchanged.
Figure~\ref{fig:task_card} summarizes the Qwen3.5-0.8B adaptation task and its reference design in this form.

\ifcompact\fi

\paragraph{How much RTL the agent writes.}
Table~\ref{tab:rtl_scale} gives, for each of the twelve tasks of the correctness track, the size of the reference design, the RTL in the starting workspace, and how many lines accepted submissions add and delete relative to their starting files. A scaffold supplies 85 to 88\% of the reference's lines, and accepted adaptations add a median of 1,382 to 2,311 lines and delete 729 to 1,262.5. From interfaces, the workspace holds 284 to 746 lines, and accepted designs add a median of 6,154 lines on nano T2 and 34,300 on Gemma 4 E4B; on Gemma 4 E4B and Llama 3.2 3B the median accepted design adds more lines than the reference contains (34,300 against 31,884 and 27,625 against 26,713). The nano tasks share one reference, and T3's workspace supplies 132 more lines than T2's.

\begin{table}[tbp]
\centering
\caption{\textbf{RTL size of the reference designs, the starting workspaces, and accepted submissions.} Net added and deleted lines compare each accepted submission with its starting files; medians with quartiles.}
\label{tab:rtl_scale}
{\tablefont\begingroup
\small
\setlength{\tabcolsep}{4pt}
\noindent{\footnotesize (a) Size of the reference and of the starting workspace}\par\smallskip
\begin{tabularx}{\linewidth}{@{}>{\raggedright\arraybackslash}X>{\raggedleft\arraybackslash}p{0.22\linewidth}>{\raggedleft\arraybackslash}p{0.22\linewidth}>{\raggedleft\arraybackslash}p{0.12\linewidth}@{}}
\toprule
\textbf{Task / start} & \textbf{Reference lines / modules} & \textbf{Provided lines / modules} & \textbf{Accepted $n$} \\
\midrule
Nano T2 / interfaces & 6,843 / 34 & 284 / 6 & 103 \\
Nano T3 / interfaces & 6,843 / 34 & 416 / 8 & 91 \\
Qwen 0.8B / scaffold & 18,742 / 81 & 16,308 / 74 & 101 \\
Qwen 4B / scaffold & 29,615 / 109 & 25,204 / 98 & 74 \\
Gemma E4B / interfaces & 31,884 / 121 & 746 / 17 & 25 \\
Gemma E4B / scaffold & 31,884 / 121 & 27,492 / 108 & 54 \\
Llama 3B / interfaces & 26,713 / 98 & 682 / 15 & 31 \\
Llama 3B / scaffold & 26,713 / 98 & 23,468 / 91 & 62 \\
Gemma 3 4B / interfaces & 29,376 / 112 & 728 / 16 & 82 \\
Gemma 3 4B / scaffold & 29,376 / 112 & 25,018 / 98 & 122 \\
Llama 1B / interfaces & 21,843 / 83 & 594 / 13 & 101 \\
Llama 1B / scaffold & 21,843 / 83 & 18,972 / 75 & 170 \\
\bottomrule
\end{tabularx}
\par\medskip
\noindent{\footnotesize (b) Lines changed by accepted submissions, median (Q1--Q3)}\par\smallskip
\begin{tabularx}{\linewidth}{@{}>{\raggedright\arraybackslash}X>{\raggedleft\arraybackslash}p{0.33\linewidth}>{\raggedleft\arraybackslash}p{0.31\linewidth}@{}}
\toprule
\textbf{Task / start} & \textbf{Net added} & \textbf{Net deleted} \\
\midrule
Nano T2 / interfaces & 6,154 (5,418--7,167) & 37 (22--49) \\
Nano T3 / interfaces & 6,634 (5,693--7,264) & 51 (31.5--71) \\
Qwen 0.8B / scaffold & 1,382 (1,125--1,802) & 729 (486--1,068) \\
Qwen 4B / scaffold & 2,017 (1,559.5--2,767.5) & 1,104 (817.75--1,595.25) \\
Gemma E4B / interfaces & 34,300 (30,344--37,791) & 80 (55--133) \\
Gemma E4B / scaffold & 2,311 (1,786--2,864.75) & 1,237.5 (935.25--1,526.5) \\
Llama 3B / interfaces & 27,625 (23,680.5--30,699.5) & 76 (46--113.5) \\
Llama 3B / scaffold & 2,177.5 (1,749.75--2,549) & 1,262.5 (900.5--1,566.25) \\
Gemma 3 4B / interfaces & 27,895.5 (25,126.75--32,273) & 63.5 (44--90.25) \\
Gemma 3 4B / scaffold & 2,279.5 (1,863--2,639.25) & 1,245 (1,014--1,469.5) \\
Llama 1B / interfaces & 21,384 (18,620--24,593) & 79 (46--104) \\
Llama 1B / scaffold & 1,669.5 (1,388.25--1,936.5) & 910 (721.75--1,133.5) \\
\bottomrule
\end{tabularx}
\endgroup
}
\par\smallskip{\footnotesize\raggedright Sizes come from the authors' records of each reference design, starting workspace, and accepted submission. Accepted $n$ counts the accepted correctness-track attempts per task. Interfaces: module declarations, testbench interfaces, and arithmetic components; scaffold: a working accelerator to adapt.\par}
\end{table}

\paragraph{Reference designs.}
Three engineers built the reference designs of all seven correctness-track targets. Table~\ref{tab:reference_origins} records the five Qwen3.5, Gemma 4, and Llama 3.2 3B references, and Appendix~\ref{app:extended_models} describes the Gemma 3 4B and Llama 3.2-1B references and their test suites. The nano and Gemma 4 E4B references were written by hand and then reviewed independently. For the 0.8B, Llama 3.2 3B, Gemma 3 4B, and Llama 3.2-1B references, an agent drafted individual components, and the engineer reviewed, edited, and integrated them. The 4B reference was derived from a submission that passed every gate. The build-session records name the agent of the 0.8B, Llama 3.2 3B, and 4B work as \texttt{anthropic/claude-opus-4-20251020}, which is not one of the four evaluated agents, so no evaluated agent starts from its own code. On the 4B submission the engineer changed 70 lines: 47 in the reset sequencing, which now clears state in a fixed order and zeroes the valid count, and 23 in the parameter-loading wrapper, which now aligns bursts and checks weight addresses against their bounds. The arithmetic datapath and the commit protocol are unchanged. The public and held-out suites were written from the integer contract before any reference RTL existed and do not depend on the agent-drafted source. Every reference then passed the same gates as an agent submission (Equation~\ref{eq:acceptance}).

\begin{table}[H]
\centering
\caption{\textbf{How the Qwen3.5, Gemma 4, and Llama 3.2 3B reference designs were built.} Three engineers built the five references. An independent reviewer compared each reference with its integer contract and then ran the operator, layer, and integration gates. Later work is quoted from the author records. SHA-256 prefix: the first 12 hexadecimal digits of the SHA-256 content digest of the reviewed reference file.}
\label{tab:reference_origins}
{\small\tablefont\setlength{\tabcolsep}{4pt}\resultrows
\begin{tabularx}{\linewidth}{@{}l>{\centering\arraybackslash}p{0.07\linewidth}>{\raggedright\arraybackslash}p{0.27\linewidth}>{\raggedright\arraybackslash}Xl@{}}
\toprule
Target & Engineer & Construction & Later work & SHA-256 prefix \\
\midrule
Qwen3.5-nano & 1 & Written by hand & Independent contract review and reset/cache-bound fixes & \texttt{e1eda35ea043} \\
Qwen3.5-0.8B & 2 & Agent-drafted components, reviewed, edited, and integrated by the engineer & Builder reviews generated components and integrates integer datapath & \texttt{e5dcfc78b525} \\
Qwen3.5-4B & 2 & Derived from an accepted agent submission & Builder changes reset sequencing and parameter-loading wrapper & \texttt{a1f60be5ec68} \\
Gemma 4 E4B & 3 & Written by hand & Builder implements the contract-specific normalization and cache layout & \texttt{392c996167ba} \\
Llama 3.2 3B & 3 & Agent-drafted components, reviewed, edited, and integrated by the engineer & Builder integrates reviewed attention and parameter-service components & \texttt{d13f1ccbede2} \\
\bottomrule
\end{tabularx}}
\end{table}

\paragraph{Agent configuration and attempt records.}
An agent configuration covers the model and the execution settings that govern its behavior.
The attempt specification requires the model identifier returned by the provider, its revision when available, the complete system and user prompts, the harness revision, reasoning and sampling settings, the context-management policy, stopping criteria, and explicit caps on wall time, tokens, and tool calls.
Termination labels separate timeouts from RTL failure categories. Table~\ref{tab:T12_caps} gives the caps on wall time, tool calls, and output tokens in each study. The benchmark runs use the provider model identifiers \texttt{openai/gpt-5.6-sol-20261015} (Sol 5.6), \texttt{openai/astra-6-20260930} (Astra 6), \texttt{anthropic/claude-opus-5-5-20251101} (Opus 5.5), and \texttt{anthropic/claude-fable-5-1-20261101} (Fable 5.1). All four models run at their provider's default reasoning effort, which is medium.

\begin{table}[H]
\centering
\caption{\textbf{Caps on wall time, tool calls, and output tokens by study.}}
\label{tab:T12_caps}
{\small\tablefont\setlength{\tabcolsep}{3.5pt}\renewcommand{\arraystretch}{1.04}
\begin{tabular*}{\linewidth}{@{\extracolsep{\fill}}lrrrr@{}}
\toprule
Study/view & Attempts & Wall hours & Tool calls & Output tokens (k) \\
\midrule
Earlier task series & 240 & 3--12 & 180--420 & 90--260 \\
Cross-architecture & 480 & 16--30 & 560--900 & 340--600 \\
Repeated instances & 192 & 16--18 & 560--600 & 340--360 \\
Feedback study & 1,152 & 5--12 & 260--420 & 140--260 \\
Controlled matrix & 3,456 & 3--9 & 160--480 & 90--270 \\
Gemma 3 4B and Llama 3.2-1B & 1,024 & 16--30 & 560--900 & 340--600 \\
Gemma 3 4B and Llama 3.2-1B, 8 steps & 1,024 & 16--30 & 560--900 & 340--600 \\
Correctness track & 1,888 & 5--30 & 260--900 & 140--600 \\
Sol feedback pilot & 32 & 5--6 & 260--320 & 140--180 \\
PPA track & 1,536 & 12--18 & 420--600 & 260--360 \\
\bottomrule
\end{tabular*}}
\par\smallskip{\footnotesize\raggedright The correctness track is the full-state setting of the feedback study, the cross-architecture tasks, and the Gemma 3 4B and Llama 3.2-1B tasks, so the rows do not add up.\par}
\end{table}

The results of the earlier task series (Appendix~\ref{app:cohorts}) report the nano construction tasks and the larger scaffold-adaptation task in separate columns because they start from different workspaces: interface files for nano and a supplied scaffold for adaptation. Table~\ref{tab:sa-budgets} gives the task allocation and caps for that series.

\subsection{Attempt counts and statistics}
\label{app:additional}
Table~\ref{tab:additional_attempts} counts the 9,136 agent attempts by study: 8,984 in the eight studies it lists, which Sections~\ref{sec:results} and~\ref{sec:findings} summarize, and 152 by Sol 5.6 and Astra 6 on the earlier task series and its feedback pilot.

Single acceptance rates carry Wilson 95\% intervals. Paired contrasts carry block-bootstrap intervals over matched instances; those of the feedback and state-condition contrasts are computed from the paired attempt records (Appendices~\ref{app:add-feedback} and~\ref{app:add-state}).

\begin{table}[H]
\centering
\caption{\textbf{Agent attempts by study.} Every attempt is a fresh process and workspace with a unique identifier.}
\label{tab:additional_attempts}
{\small\tablefont\setlength{\tabcolsep}{4pt}\resultrows
\begin{tabular}{@{}l>{\raggedright\arraybackslash}p{.44\linewidth}lr@{}}
\toprule
Study & Design & Agents $\times$ cells & Attempts \\
\midrule
Earlier task series & Opus 5.5 and Fable 5.1 on T1--T3 (16 each) and 0.8B adaptation (12) & 2 $\times$ 60 & 120 \\
Cross-architecture & Gemma 4 E4B and Llama 3.2 3B from scaffold and interfaces; Qwen3.5-4B from scaffold & 4 $\times$ 5 $\times$ 24 & 480 \\
Repeated instances & Gemma 4 E4B and Llama 3.2 3B adaptation, 8 instances $\times$ 3 repeats & 4 $\times$ 2 $\times$ 24 & 192 \\
Feedback study & T2, T3, and 0.8B under output-only, coarse-state, and full-state feedback & 4 $\times$ 3 $\times$ 3 $\times$ 32 & 1,152 \\
Controlled matrix & One scaffold; 4 horizons $\times$ 3 state conditions $\times$ 3 budgets & 4 $\times$ 36 $\times$ 24 & 3,456 \\
PPA track & Improvement of a correct Qwen3.5-0.8B, Qwen3.5-4B, Gemma 4 E4B, or Llama 3.2 3B design & 4 $\times$ 4 $\times$ 96 & 1,536 \\
Gemma 3 4B, Llama 3.2-1B & Gemma 3 4B (text) and Llama 3.2-1B from scaffold and interfaces, 64 instances each, 32 steps & 4 $\times$ 4 $\times$ 64 & 1,024 \\
Horizon comparison & The same Gemma 3 4B and Llama 3.2-1B instances at eight steps & 4 $\times$ 4 $\times$ 64 & 1,024 \\
\midrule
\multicolumn{3}{@{}l}{Subtotal} & 8,984 \\
\multicolumn{3}{@{}l}{Sol 5.6 and Astra 6 on the earlier task series, with the Sol feedback pilot (Appendix~\ref{app:cohorts})} & 152 \\
\multicolumn{3}{@{}l}{\textbf{All agent attempts}} & \textbf{9,136} \\
\bottomrule
\end{tabular}
}
\par\smallskip{\footnotesize\raggedright No attempt appears in more than one study; each feedback setting and each repeat is a separate attempt.\par}
\end{table}

\subsection{Acceptance rate and agent-hours per accepted design}
\label{app:cost}
For a fixed task and agent configuration, independent attempts yield binary outcomes $Y_i$, with $Y_i=1$ when attempt $i$ is accepted, and the acceptance rate is $\hat p=\sum_iY_i/N$. The correctness track has 24, 32, or 64 attempts per agent on each task (Table~\ref{tab:benchmark}). The earlier task series has 16 attempts per agent on each nano task and 12 on 0.8B adaptation; Table~\ref{tab:key_results} reports its counts.

For attempt costs $c_i$, the cost per accepted design is
\begin{equation}
\widehat c_{\rm accepted}=\frac{\sum_{i=1}^N c_i}{\sum_{i=1}^N Y_i}.
\end{equation}
The numerator keeps the costs of failed attempts, and the ratio is undefined, not zero, when no attempt is accepted. With agent time as the cost, this ratio is the agent-hours per accepted design reported in Table~\ref{tab:leaderboard}. It summarizes the observed attempts; reading it as the expected cost of future retries requires a model of how repeated attempts are distributed. Dividing the median cost by $\hat p$ generally gives a different value.

\section{Tests, evaluator checks, and RTL components}
\label{app:evaluation}

\subsection{Public and held-out tests}
\label{app:tests}
The public suite shows the agent the interface and gives reproducible feedback. The held-out suite changes seeds and short token sequences and uses only operations the contract supports. Table~\ref{tab:testexamples} pairs each kind of public test with its held-out variation. A test case is one complete stimulus and its checks; the tensor elements within a case are not independent statistical trials. One mismatch fails a case, and one failed required case fails the suite's gate ($I_{\rm public}$ or $I_{\rm heldout}$ in Equation~\ref{eq:acceptance}). The longest decoding tests run 32 steps.

\begin{table}[ht]
\centering\small
\caption{\textbf{Public tests and their held-out variations.} Both suites follow the same documented semantics.}
\label{tab:testexamples}
\tablefont
\begin{tabularx}{\linewidth}{lXX}\toprule
Check & Public tests & Held-out variation and expected behavior \\\midrule
Rounding & Shift by 8: $128\mapsto0$, $384\mapsto2$ & Negative ties: $-128\mapsto0$, $-384\mapsto-2$; ties to even \\
Saturation & Largest representable INT16 result & Positive and negative overflow clamp to 32,767 and $-32,768$ \\
Token sequence & Prefill IDs $[1,7,13]$; decode eight steps & Different in-range IDs and prompt length five; all logits/state exact \\
Weight reload & Load set A into an idle engine & Load shape-compatible quantized set B; no stale A values \\
Recurrent state & Zero initial GDN and convolution state & Small supported nonzero state; check state after every token \\
Reset/isolation & Reset before first sequence & Decode A, reset, decode B; compare B with a fresh instance \\
Backpressure & Always-ready in-order responses & Stall for 1--3 cycles; stable payload, no loss or duplicate commit \\
Cache address & Start from position zero & Nonzero supported position and a short prefill/decode transition \\
Integrity & Read-only oracle and constraints & Reject modifications; run evaluator from an independent clean copy \\
\bottomrule\end{tabularx}
\end{table}

\paragraph{New weights and inputs.}
The held-out suite's new quantized weights are arrays generated within the supported shapes and ranges; they can catch hard-coded vectors and stale weight buffers. The scales are not randomized, because arbitrary scales would violate the frozen arithmetic contract. Fidelity to FP32 is measured with weights derived from the model checkpoint (Section~\ref{sec:correctness}).

\paragraph{Fidelity by decoding position.}
The pooled fidelity of the Qwen3.5-0.8B and 4B contracts rests on 1,000 sequences at each of eight decoding positions: 7,931 of 8,000 top-1 decisions match FP32 for Qwen3.5-0.8B (99.1375\%) and 7,937 for Qwen3.5-4B (99.2125\%). Table~\ref{tab:fidelity_positions} gives the breakdown by position in percent for all six reference targets. Qwen3.5-nano uses 256 sequences per position and matches FP32 on 1,991 of 2,048 decisions. Both executions see the same token histories, but the FP32 and integer states each evolve under their own arithmetic.

\paragraph{Local checks.}
The local checks of Equation~\ref{eq:acceptance} are bounded safety checks on reduced instances of the controller. They assume a single clock, reset at the initial step, in-order memory responses, and response stalls of at most three cycles, and they run for 32 cycles. Five properties are checked: the output payload is stable while blocked, one accepted request commits at most once, reset clears valid state, issued addresses are in the configured range, and reload occurs only while idle. 

\subsection{Development defects}
\label{app:gate_analysis}
To check that the acceptance gates reject known defects, we grade twelve implementations that each contain one specified defect, together with a correct control. The unit is one known-defect implementation. The control must pass every gate, and each known-defect implementation must be rejected by at least one. Table~\ref{tab:mutation_checks} gives the outcome of every gate.

\begin{table}[t]
\centering
\caption{\textbf{Gate outcomes for the twelve development defects and one correct control.} Pass means that a gate accepts the implementation, even if it is defective.}
\label{tab:mutation_checks}
\small
\tablefont
\begin{tabular}{lcccc}
\toprule
Implementation & Public & Held-out & Synthesis & Integrity \\
\midrule
Correct reference & pass & pass & pass & pass \\
Wrong rounding mode & reject & reject & pass & pass \\
Signed accumulator truncation & reject & reject & pass & pass \\
Saturation off by one & pass & reject & pass & pass \\
GDN update before decay & pass & reject & pass & pass \\
Conv-state late writeback & reject & reject & pass & pass \\
KV-cache position offset & pass & reject & pass & pass \\
Reset leakage & pass & reject & pass & pass \\
Public-vector hard-code & pass & reject & pass & reject \\
Held-out group-size hard-code & pass & reject & pass & pass \\
Fixed memory-latency assumption & pass & reject & pass & pass \\
Non-synthesizable lookup & pass & pass & reject & pass \\
Protected evaluator edit & pass & pass & pass & reject \\
\bottomrule
\end{tabular}
\end{table}

The public tests reject three of the twelve: incorrect rounding, signed-accumulator truncation, and late convolution-state writeback. The held-out numerical tests reject these three and seven more: saturation, GDN update ordering, cache position, reset leakage, the public-vector hard-code, the held-out group-size hard-code, and fixed memory latency. The remaining two pass both numerical suites; they are a non-synthesizable lookup, which synthesis rejects, and an edit to the protected evaluator, which the integrity gate rejects. Together the numerical, synthesis, and integrity gates reject all twelve and accept the control.

The rejection sets overlap, so their counts cannot be added; the integrity gate, for instance, also rejects the public-vector hard-code. Each of the three kinds of gate rejects at least one defect that the other two accept, which supports keeping all three in the acceptance rule. Protected-file hashing and test isolation address different risks, and the evaluator needs both: hashing detects edits to protected files, and isolation keeps held-out assets out of the agent's workspace. The held-out defect study (Appendix~\ref{app:add-faults}) measures detection on a broader sample of 48 defects.

The protocol ties any reported first-detection time to an activated defect and an identified stimulus, and it reports a defect that is never activated within the test horizon separately from one that executes and escapes every assertion.

\subsection{Held-out defects}
\label{app:add-faults}
The held-out defect study adds 48 defects in eight families to the twelve development defects (Section~\ref{sec:evaluator_defects}). Each of the 48 has a repaired counterpart, and there are eight independent correct controls (Figure~\ref{fig:F07_evaluator}, Tables~\ref{tab:T08-coverage} and~\ref{tab:T08-development}). The public tests alone reject 3 of the 48 defects, the two numerical suites together 29, the numerical suites and synthesis 34, and the complete evaluator 43 (89.6\%). None of the 56 correct implementations (48 repaired, eight controls) is rejected, so the evaluator separates 43 defect--repair pairs. The five missed defects are one each in wiring/integration, reset/isolation, and synthesis, and two in local properties. 

\begin{figure}[htbp]
\centering
\includegraphics[width=\linewidth,height=.62\textheight,keepaspectratio]{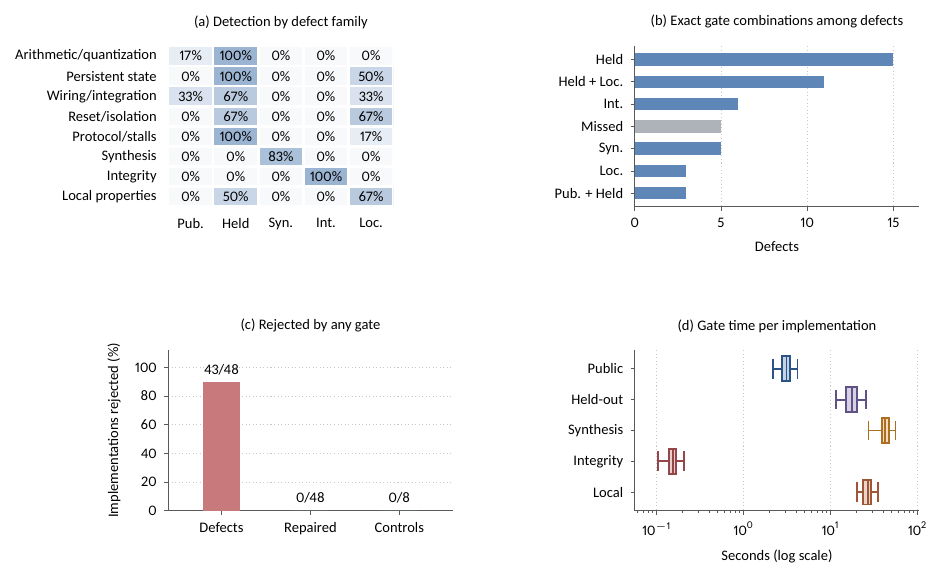}
\caption{Evaluator coverage on 48 held-out defects in eight families (six per family), 48 repaired counterparts, and 8 correct controls. (a) Detection rate by family and gate. (b) Most frequent exact gate combinations among defects, including no detection; each defect falls in one combination (Pub., public; Held, held out; Syn., synthesis; Int., integrity; Loc., local). (c) Rejection by any gate. (d) Gate time per implementation as median, interquartile range, and 1.5-IQR whiskers; outliers stay in the data but are omitted from the plot. The development defects appear in their own table.}
\label{fig:F07_evaluator}
\end{figure}

\begin{table}[htbp]
\centering
\caption{\textbf{Detection of 48 held-out defects and rejection of eight correct controls, with bootstrap 95\% intervals.}}
\label{tab:T08-coverage}
{\small\tablefont\setlength{\tabcolsep}{7pt}\resultrows
\begin{tabular}{@{}lrrr}
\toprule
Gate configuration & Detected (\%) & False reject & Time (s) \\
\midrule
Public & \cellcolor{blueC!7}3/48 (6.3)\,\textsubscript{\tiny(0.0--14.6)} & 0/8 & 3.1 \\
Numerical & \cellcolor{blueC!32}29/48 (60.4)\,\textsubscript{\tiny(35.4--85.4)} & 0/8 & 20.7 \\
Numerical + synthesis & \cellcolor{blueC!37}34/48 (70.8)\,\textsubscript{\tiny(45.8--89.6)} & 0/8 & 64.1 \\
\textbf{Full} & \cellcolor{blueC!45}43/48 (89.6)\,\textsubscript{\tiny(81.2--97.9)} & \textbf{0/8} & \textbf{91.3} \\
\bottomrule
\end{tabular}
}
\par\smallskip{\footnotesize\raggedright A configuration rejects an implementation when any of its gates does, so overlapping detections count once. The complete evaluator (bold) is the acceptance rule. Intervals resample defect families and describe this set of defects only.\par}
\end{table}
\begin{table}[htbp]
\centering
\caption{\textbf{Development defects, reported apart from the held-out defects.}}
\label{tab:T08-development}
{\small\tablefont\setlength{\tabcolsep}{4pt}\resultrows
\begin{tabular*}{\linewidth}{@{\extracolsep{\fill}}lrrrrrr@{}}
\toprule
Population & $n$ & Public & Held out & Synth. & Integr. & Union \\
\midrule
Development defects & 12 & 3 & 10 & 1 & 2 & 12 \\
Development correct control & 1 & 0 & 0 & 0 & 0 & 0 \\
\bottomrule
\end{tabular*}
}
\par\smallskip{\footnotesize\raggedright Each gate column counts rejected implementations. Detection here establishes coverage of these specified defects only; the held-out defects measure detection beyond them.\par}
\end{table}

\subsection{Arithmetic and state-commit RTL components}
\label{app:rtl}
This subsection describes two components the main text refers to (Section~\ref{sec:bg_decoding}): the requantizer (Listing~\ref{lst:requant}) and a buffer that commits state only on accepted transfers. A directed testbench for both covers positive and negative rounding ties, saturation, enqueue without commit, a three-cycle stall, simultaneous pop/push, and reset isolation. Each module is complete at its stated interface.

\paragraph{Nearest-even requantization.}
For shift $k$, write $x=2^kq+r$ with $0\le r<2^k$ and $q=\lfloor x/2^k\rfloor$. Increment $q$ when $r>2^{k-1}$, or on an exact tie when $q$ is odd; then saturate. This formulation also handles negative ties, because an arithmetic right shift computes the floor quotient. Taking the absolute value first would overflow at the most negative representable input and would need special handling. The module's legal domain is $1\le\texttt{SHIFT}<\texttt{IN\_W}$ and $2\le\texttt{OUT\_W}\le\texttt{IN\_W}$.

\begin{lstlisting}[language=Verilog,caption={Signed nearest-even requantizer with output saturation. The one-bit-wider intermediate preserves the rounding increment before comparison with the output limits.},label={lst:requant}]
// Signed power-of-two requantization: round to nearest, ties to even, saturate.
// Legal parameters: 1 <= SHIFT < IN_W, 2 <= OUT_W <= IN_W.
module requant_rne_sat #(
    parameter integer IN_W = 32,
    parameter integer OUT_W = 16,
    parameter integer SHIFT = 8
) (
    input  wire signed [IN_W-1:0] x,
    output reg  signed [OUT_W-1:0] y
);
    localparam signed [IN_W:0] MAX_Y =
        ({{IN_W{1'b0}},1'b1} << (OUT_W-1)) - 1'b1;
    localparam signed [IN_W:0] MIN_Y =
        -({{IN_W{1'b0}},1'b1} << (OUT_W-1));
    localparam [SHIFT-1:0] HALF = {1'b1,{(SHIFT-1){1'b0}}};
    reg signed [IN_W:0] floor_q;
    reg signed [IN_W:0] rounded_q;
    reg [SHIFT-1:0] remainder;
    reg increment;
    always @* begin
        // Arithmetic shift is floor division, also for negative inputs.
        floor_q = $signed({x[IN_W-1],x}) >>> SHIFT;
        remainder = x[SHIFT-1:0];
        increment = (remainder > HALF) ||
                    ((remainder == HALF) && floor_q[0]);
        rounded_q = floor_q + {{IN_W{1'b0}},increment};
        if (rounded_q > MAX_Y)
            y = {1'b0,{(OUT_W-1){1'b1}}};
        else if (rounded_q < MIN_Y)
            y = {1'b1,{(OUT_W-1){1'b0}}};
        else
            y = rounded_q[OUT_W-1:0];
    end
endmodule
\end{lstlisting}

\paragraph{State commit and cancellation.}
The one-entry buffer accepts a new pending state word when it is empty or when the current word is consumed, and architectural state changes on output acceptance. A simultaneous pop/push commits the old payload and retains the new one, using nonblocking assignments. Reset or a new-sequence request cancels pending work and clears the state. A system with a different reset policy needs a different contract and different tests.

\subsection{Delayed state failure in a constructed attention decoder}
\label{app:decoder_case}
\label{app:delayed}
This section shows a delayed state failure in a decoder with a cache, handshakes, and a memory interface, small enough to inspect in full. The authors constructed it for this case study: one attention head over a four-token vocabulary, embedding width 2, an eight-entry key--value cache of 16-bit words, 32-bit accumulators, one memory response per token, and valid/ready handshakes on input, memory, and output. A state word is the cache's valid count or one cache entry. The reference RTL is one module of 143 lines, 134 of which are neither blank nor comment-only, and the testbench has 110 lines (Table~\ref{tab:T14_decoder_source}). An integer oracle computes the expected logits and state. All results in this section come from Icarus Verilog simulation of each snapshot on a fixed, directed public suite of four cases (24 transactions) and a held-out suite of five cases (31 transactions).

The seeded defect writes the cache entry and increments the valid count whenever the output is valid, before the output transfer is accepted. Under output backpressure the design therefore commits one token more than once. In the public case with output stalls, the valid count is first wrong at transaction 2, the logits at transaction 5, and the selected token at transaction 6 (Figure~\ref{fig:F09_decoder_case}). Two scripted repairs follow (Table~\ref{tab:T13_decoder_regressions}). The first guards the cache payload with the output handshake but still increments the valid count; every logit and token of the public suite is then exact, while the state and commit checks still fail, and the held-out suite still fails four of five cases. The second commits the entry and the count only on an accepted output transfer and passes both suites; it reproduces the reference RTL byte for byte. The authors wrote both repairs as patches and replayed them in order, so the sequence shows what each check reports on a known defect. Appendix~\ref{app:debug_trace} gives an agent run on the same defect, and the regrading of 144 frozen agent submissions and the three references (Section~\ref{sec:evaluator_results}) is the corresponding evidence from real submissions.

\begin{figure}[tbp]
\centering
\includegraphics[width=0.92\linewidth]{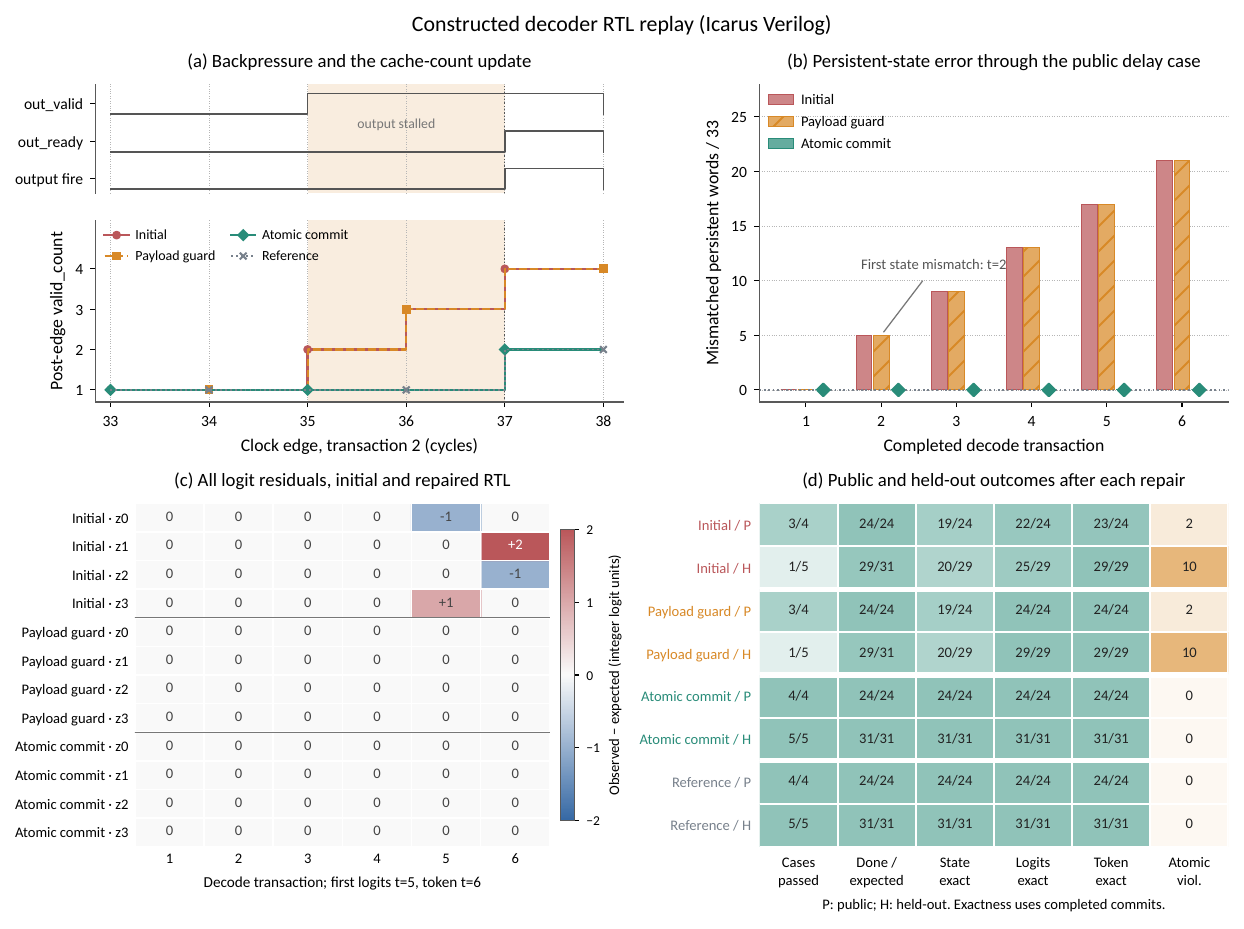}
\caption{Constructed-decoder replay in Icarus Verilog simulation. (a) Logged handshakes before the clock edge and cache counts after it during two output-stall cycles of transaction 2. (b) Mismatched persistent words after each completed transaction in the public delay case. (c) Signed residuals of the four integer logits; the initial snapshot first differs in state at transaction 2, in logits at 5, and in the selected token at 6. The scripted payload guard restores the outputs while the state and commit checks still fail. (d) Public and held-out outcomes for each snapshot and the reference. Exactness ratios use completed transactions; incomplete transactions appear in the completion column. Reset and cancellation events are protocol checks, not output observations.}
\label{fig:F09_decoder_case}
\end{figure}
\begin{table}[H]
\centering
\caption{\textbf{Constructed-decoder outcomes after each scripted repair (Icarus Verilog simulation).}}
\label{tab:T13_decoder_regressions}
{\footnotesize\tablefont\setlength{\tabcolsep}{3.5pt}\renewcommand{\arraystretch}{1.04}
\begin{tabular*}{\linewidth}{@{\extracolsep{\fill}}llrrrrrrr@{}}
\toprule
\multicolumn{2}{c}{} & \multicolumn{2}{c}{Cases and transactions} & \multicolumn{3}{c}{Exact transactions} & \multicolumn{2}{c}{Violations / timeouts} \\
\cmidrule(lr){3-4}\cmidrule(lr){5-7}\cmidrule(lr){8-9}
Snapshot & Suite & Cases passed & Completed & State & Logits & Token & Commit & Input cap \\
\midrule
Initial & Public & 3/4 & 24/24 & 19 & 22 & 23 & \cellcolor{orangeC!14}2 & 0 \\
Initial & Held-out & 1/5 & 29/31 & 20 & 25 & 29 & \cellcolor{orangeC!14}10 & 2 \\
\addlinespace[3pt]
Payload guard & Public & 3/4 & 24/24 & 19 & 24 & 24 & \cellcolor{orangeC!14}2 & 0 \\
Payload guard & Held-out & 1/5 & 29/31 & 20 & 29 & 29 & \cellcolor{orangeC!14}10 & 2 \\
\addlinespace[3pt]
Atomic commit & Public & \cellcolor{blueC!15}4/4 & 24/24 & 24 & 24 & 24 & 0 & 0 \\
Atomic commit & Held-out & \cellcolor{blueC!15}5/5 & 31/31 & 31 & 31 & 31 & 0 & 0 \\
\addlinespace[3pt]
Reference & Public & 4/4 & 24/24 & 24 & 24 & 24 & 0 & 0 \\
Reference & Held-out & 5/5 & 31/31 & 31 & 31 & 31 & 0 & 0 \\
\bottomrule
\end{tabular*}}
\par\smallskip{\footnotesize\raggedright Exactness counts use completed transactions and the independent integer oracle. Commit counts are cycles violating atomic state update. One pending transaction per snapshot is intentionally canceled by reset and excluded from expected outputs. Final repair and reference complete all 55 expected transactions. The suite is fixed and directed. The two repairs are patches written by the authors and replayed in order; they are not an agent trajectory.\par}
\end{table}

\begin{table}[H]
\centering
\caption{\textbf{Constructed-decoder source size and patch size.}}
\label{tab:T14_decoder_source}
{\footnotesize\tablefont\setlength{\tabcolsep}{3.5pt}\renewcommand{\arraystretch}{1.04}
\begin{tabular*}{\linewidth}{@{\extracolsep{\fill}}lrrrrr@{}}
\toprule
\multicolumn{1}{c}{} & \multicolumn{3}{c}{Source size} & \multicolumn{2}{c}{Incoming patch} \\
\cmidrule(lr){2-4}\cmidrule(lr){5-6}
Snapshot / source & File lines & Counted lines & Modules & Added lines & Removed lines \\
\midrule
Reference & 143 & 134 & 1 & -- & -- \\
Initial & 143 & 134 & 1 & -- & -- \\
Payload guard & 145 & 136 & 1 & 6 & 4 \\
Atomic commit & 143 & 134 & 1 & 6 & 8 \\
Testbench & 110 & 106 & 1 & -- & -- \\
\bottomrule
\end{tabular*}}
\par\smallskip{\footnotesize\raggedright File lines count every line of the source file; counted lines exclude blank and comment-only lines. Patch counts include every added/deleted text line in each supplied unified diff. The testbench is reported separately. A dash denotes no incoming patch for that file. Patch application reproduces the adjacent snapshot byte for byte; these counts describe only the constructed decoder.\par}
\end{table}

\subsection{A paired debugging trace}
\label{app:debug_trace}
To show what an agent does with each kind of feedback, Sol 5.6 repaired the seeded decoder once under full-state and once under output-only feedback (Table~\ref{tab:T15_debug_trace}). Both runs start from the same RTL and see the same public suite, which fails one of four cases. Full-state feedback names the first mismatching state word: at transaction 2 the cache's valid count is 4 where the oracle expects 2. Output-only feedback reports the first output mismatch, the logits of transaction 5. In both runs the agent's first and only edit changes the commit condition from output valid to output valid and ready, a one-line change; the public suite then passes, the agent submits, and the frozen RTL passes all five held-out cases. On this small defect both settings lead to the same repair, and full-state feedback points to the premature commit three transactions earlier. One run per setting shows the mechanism; the feedback study of Appendix~\ref{app:add-feedback} measures the rates. The runs use temperature 0.2, top-p 0.95, reasoning disabled, a 131,072-token context that keeps every public message without compaction, and caps of 24 tool calls, 0.5 hours, and 25,600 output tokens.

\begin{table}[H]
\centering
\caption{\textbf{One repair of the constructed decoder under each feedback setting.}}
\label{tab:T15_debug_trace}
{\footnotesize\tablefont\setlength{\tabcolsep}{3pt}\renewcommand{\arraystretch}{1.04}
\begin{tabularx}{\linewidth}{@{}llcrrr>{\raggedright\arraybackslash}X@{}}
\toprule
Feedback & Evaluation & \shortstack{Shown\\to agent} & Cases & \shortstack{State/logits/\\token exact} & \shortstack{Commit\\viol.} & First mismatch reported \\
\midrule
Full-state & Initial RTL, public & yes & 3/4 & 19/22/23 of 24 & 2 & tx 2: valid count 4, exp.\ 2 \\
Full-state & After edit 1, public & yes & 4/4 & 24/24/24 of 24 & 0 & -- \\
Full-state & Frozen, held-out & no & 5/5 & 31/31/31 of 31 & 0 & -- \\
\addlinespace[3pt]
Output-only & Initial RTL, public & yes & 3/4 & 19/22/23 of 24 & 2 & tx 5: logits [12,12,$-$6,$-$5], exp.\ [13,12,$-$6,$-$6] \\
Output-only & After edit 1, public & yes & 4/4 & 24/24/24 of 24 & 0 & -- \\
Output-only & Frozen, held-out & no & 5/5 & 31/31/31 of 31 & 0 & -- \\
\bottomrule
\end{tabularx}}
\par\smallskip{\footnotesize\raggedright Sol 5.6 starts from the seeded RTL of Appendix~\ref{app:decoder_case} and makes one edit under each setting; both edits change one line (1 added, 1 deleted) and yield the same RTL. Exact counts come from the evaluator; the output-only agent is shown output mismatches only. The held-out evaluation uses full-state checks for both runs and is not returned to the agent. Provider-reported use (full-state: 18,420 input and 2,156 output tokens, 18.4 s; output-only: 16,280 input and 1,984 output tokens, 15.7 s). tx: transaction; viol.: violations.\par}
\end{table}

\subsection{Failure labels, attempt records, and two case studies}
\label{app:traces}

\paragraph{Failure labels.}
The grader policy labels a rejected submission by its first failing gate, in the order integrity, compilation, public suite, held-out suite, synthesis, and local checks; an attempt that ends at a cap is not graded. For the case checks of the decoder evaluator, the policy takes the first failing check in a fixed order: input timeout, output timeout, incomplete transactions, reset clearing, output stability under stalls, cache bounds, commit atomicity, state mismatch, logit mismatch, and token mismatch. A label names the category of the first failing check. On the 36 snapshot--case records of the constructed decoder (Appendix~\ref{app:decoder_case}), two annotators labeled each record independently; they agree on 28 (Cohen's $\kappa=0.46$, 95\% interval 0.21 to 0.68), and they agree with the rule on 34 and 28.

\paragraph{Excerpts from attempt records.}
Figure~\ref{fig:failure_records} gives excerpts from the records of three attempts that fail in different ways: at the wall cap after the state defect was repaired, at the tool-call cap during optimization, and on the frozen RTL after every check the agent could see had passed.

\ifcompact\fi

\paragraph{Scaffold-adaptation case studies.}
Two stored trajectories, one for Qwen3.5-0.8B and one for Qwen3.5-4B, each document a single accepted scaffold adaptation (Table~\ref{tab:run_accounting}). Acceptance rates come from repeated attempts: for 0.8B, those of the earlier task series (Appendix~\ref{sec:agents}), and for 4B, the 96 adaptation attempts of the cross-architecture study (Appendix~\ref{sec:cross_architecture}). The 4B case study is not one of the counted attempts.
\begin{table}[H]
\centering
\caption{\textbf{Time and tool calls for the two scaffold-adaptation case studies.} Agent time includes design iterations from a shared scaffold, and grading time begins once the submitted RTL is frozen. A clean pass starts from correct RTL and runs validation and synthesis once each. Call counts include failed calls and exclude grading.}
\label{tab:run_accounting}
\small
\tablefont
\begin{tabular*}{\linewidth}{@{\extracolsep{\fill}}lrrrrrr}
\toprule
Target & Agent h & Grading h & Total h & Calls & Failed & Clean pass (min) \\
\midrule
Qwen3.5-0.8B & 7.25 & 0.35 & 7.60 & 168 & 11 & 162.2 \\
Qwen3.5-4B & 9.75 & 0.85 & 10.60 & 210 & 17 & 225.8 \\
\bottomrule
\end{tabular*}

\end{table}

A clean pass starts from correct RTL and runs validation and synthesis once each. It takes 162.2 minutes for 0.8B and 225.8 minutes for 4B, of which 31 and 44 minutes are synthesis and timing. Agent time in Table~\ref{tab:run_accounting} covers the iterative development within one attempt; agent-hours per accepted design (Appendix~\ref{app:cost}) also count the time of failed attempts.

\section{Benchmark targets}
\label{app:targets}

\subsection{Qwen3.5-nano}
\label{app:nano}
Qwen3.5-nano is our compact Qwen-derived test model. It has four layers (three GDN and one attention layer), hidden width 256, intermediate width 768, and vocabulary 4,096. It is the target of the nano correctness-track tasks (T2 and T3), the controlled matrix, and the earlier task series.

\paragraph{Reference exactness and fidelity.}
The nano reference passes all 1,384 public cases. Its contract agrees with FP32 on 1,991 of 2,048 top-1 decisions over common prefixes (97.22\%), below the 99\% fidelity threshold. During development, the reference moved from 62 of 64 to 64 of 64 exact full-model tests and from 27 of 32 to 32 of 32 exact eight-token sequences, while its operator and single-layer tests stayed exact. The development failures were at integration boundaries, and the arithmetic definition did not change while they were fixed.

\paragraph{Reference hardware.}
The nano target is not part of the PPA track, but its reference design has a baseline and an optimized schedule, as the 0.8B and 4B references do (Section~\ref{sec:hardware}). Table~\ref{tab:nano_hardware} compares the two. The optimized schedule improves throughput by 10.5\% and estimated energy efficiency by 7.8\%, and energy per token falls from 0.0162 to 0.0150\,J.
\begin{table}[htbp]
\centering
\caption{\textbf{Baseline and optimized schedules of the Qwen3.5-nano reference design.}}
\label{tab:nano_hardware}
{\small\tablefont\setlength{\tabcolsep}{3pt}\resultrows
\begin{tabular*}{\linewidth}{@{\extracolsep{\fill}}l*{5}{c}}
\toprule
Workload & {\resultstack\begin{tabular}[b]{@{}c@{}}Throughput\\(tokens/s)\end{tabular}} & {\resultstack\begin{tabular}[b]{@{}c@{}}Throughput\\gain\end{tabular}} & {\resultstack\begin{tabular}[b]{@{}c@{}}Est. efficiency\\gain\end{tabular}} & {\resultstack\begin{tabular}[b]{@{}c@{}}Power\\(W)\end{tabular}} & {\resultstack\begin{tabular}[b]{@{}c@{}}LUTs\\(approx.)\end{tabular}} \\
\midrule
Qwen3.5-nano & {\resultstack\begin{tabular}[b]{@{}c@{}}365.12\\\textbf{403.43}\end{tabular}} & \cellcolor{tealC!32}\textbf{+10.5\%} & \cellcolor{tealC!25}\textbf{+7.8\%} & {\resultstack\begin{tabular}[b]{@{}c@{}}5.90\\\textbf{6.05}\end{tabular}} & {\resultstack\begin{tabular}[b]{@{}c@{}}47.1k\\\textbf{47.3k}\end{tabular}} \\
\bottomrule
\end{tabular*}
}
\par\smallskip{\footnotesize\raggedright Top: baseline; bold: optimized. Within each pair: same arithmetic and memory model, 126\,MHz, batch size one, eight-token prompt. Power is FPGA plus DRAM; LUTs are approximate. Gains are relative to baseline; efficiency is throughput/power.\par}
\end{table}

\subsection{Qwen3.5-0.8B and 4B configurations}
\label{app:qwen-configs}
Qwen3.5-0.8B and Qwen3.5-4B are the targets of two scaffold-adaptation tasks and of the PPA track. Qwen3.5-0.8B has 24 layers, hidden width 1,024, intermediate width 3,584, and vocabulary 248,320; Qwen3.5-4B has 32 layers, hidden width 2,560, intermediate width 9,216, and vocabulary 248,320. Both text configurations include attention output gating, convolution length four, and a full-attention layer every fourth layer.

\subsection{Qualification of Gemma 4 E4B and Llama 3.2 3B}
\label{app:add-qualification}
Agents attempt only targets whose reference design passes every gate (Section~\ref{sec:tasks}). Gemma 4 E4B~\citep{gemma4card} and Llama 3.2 3B~\citep{llama32card} were evaluated under two contract conditions, labeled Main and QF in Table~\ref{tab:T02-new}. Under the Main contracts, both pass all 1,600 public and 240 held-out cases, synthesis, integrity, and local checks, and their contracts agree with FP32 on 99.78\% and 99.76\% of 4,096 pooled decisions. Every staged check passes, from operators through 32-step decoding, reset, stalls, and parameter reload (Table~\ref{tab:T02-stages}). Under QF, Gemma 4 E4B falls to 98.49\% agreement, and Llama 3.2 3B fails three held-out cases and its local check, so neither target qualifies there. All agent attempts use the Main contracts. For comparison, Table~\ref{tab:T02-current} lists the same reference checks for the six targets of Table~\ref{tab:extended_reference}.

\begin{table}[htbp]
\centering
\caption{\textbf{Reference checks for six targets.}}
\label{tab:T02-current}
{\small\tablefont\setlength{\tabcolsep}{7pt}\resultrows
\begin{tabular}{@{}lrrrrr}
\toprule
Target & Public & Held out & FP32 matches & \% & $\geq$99\% \\
\midrule
Qwen 0.8B & 3120/3120 & 232/232 & 7931/8000 & 99.138 & pass \\
Qwen 4B & 3680/3680 & 288/288 & 7937/8000 & 99.213 & pass \\
Gemma 2 2B & 1320/1320 & 80/80 & 1574/1600 & 98.375 & \cellcolor{orange!18}fail \\
Gemma 3 4B & 1320/1320 & 80/80 & 1584/1600 & 99.000 & pass \\
Qwen 27B & 1312/1320 & 70/80 & 1584/1600 & 99.000 & pass \\
Llama 1B & 1320/1320 & 80/80 & 1589/1600 & 99.312 & pass \\
\bottomrule
\end{tabular}
}
\par\smallskip{\footnotesize\raggedright Exactness is measured against each target's integer contract. The 99\% fidelity threshold applies to pooled FP32 agreement; positions within a sequence are correlated.\par}
\end{table}
\begin{table}[htbp]
\centering
\caption{\textbf{Qualification of Gemma 4 E4B and Llama 3.2 3B under two contract conditions.}}
\label{tab:T02-new}
{\small\tablefont\setlength{\tabcolsep}{4pt}\resultrows
\begin{tabular*}{\linewidth}{@{\extracolsep{\fill}}llrrrrr@{}}
\toprule
Cond. & Target & FP32 matches & \% & Public & Held out & Eligible \\
\midrule
Main & Gemma 4 E4B & 4087/4096 & 99.780 & pass & pass & pass \\
Main & Llama 3B & 4086/4096 & 99.756 & pass & pass & pass \\
QF & Gemma 4 E4B & 4034/4096 & 98.486 & pass & pass & \cellcolor{orange!18}fail \\
QF & Llama 3B & 4080/4096 & 99.609 & pass & \cellcolor{orange!18}fail & \cellcolor{orange!18}fail \\
\bottomrule
\end{tabular*}
}
\par\smallskip{\footnotesize\raggedright Eligibility also requires synthesis, integrity, and local checks. Main is the contract used in the agent studies. QF is a second contract condition; under it neither target meets every gate.\par}
\end{table}
\begin{table}[htbp]
\centering
\caption{\textbf{Qualification checks by stage for the Main contracts.}}
\label{tab:T02-stages}
{\small\tablefont\setlength{\tabcolsep}{7pt}\resultrows
\begin{tabular}{@{}lrr}
\toprule
Check & Gemma 4 E4B & Llama 3B \\
\midrule
Operators & 1024/1024 (100.0) & 1024/1024 (100.0) \\
Individual layer & 192/192 (100.0) & 192/192 (100.0) \\
Decoder block & 192/192 (100.0) & 192/192 (100.0) \\
One-step decoder & 64/64 (100.0) & 64/64 (100.0) \\
32-step decoder & 64/64 (100.0) & 64/64 (100.0) \\
Reset isolation & 64/64 (100.0) & 64/64 (100.0) \\
Held-out prefixes & 48/48 (100.0) & 48/48 (100.0) \\
Held-out cache positions & 48/48 (100.0) & 48/48 (100.0) \\
Held-out reset & 48/48 (100.0) & 48/48 (100.0) \\
Held-out stalls & 48/48 (100.0) & 48/48 (100.0) \\
Held-out reload & 48/48 (100.0) & 48/48 (100.0) \\
Synthesis & 1/1 (100.0) & 1/1 (100.0) \\
Integrity & 1/1 (100.0) & 1/1 (100.0) \\
Local properties & 1/1 (100.0) & 1/1 (100.0) \\
\bottomrule
\end{tabular}
}
\par\smallskip{\footnotesize\raggedright Cells give passed checks out of required checks (percent).\par}
\end{table}

\subsection{Adding a target}
\label{app:add_target}
A decoder becomes an agent task after four steps, which the Gemma 3 4B and Llama 3.2-1B tasks of Appendix~\ref{app:gemma3_llama1b} also followed. First, an integer contract for the decoder configuration fixes the quantized parameters, every rounding and saturation decision, and the state words the oracle checks after each step (Equation~\ref{eq:transition}). Second, the contract must agree with the FP32 model on at least 99\% of top-1 next-token decisions; Gemma 2 2B misses this threshold (98.38\%) and is not an agent task, and Gemma 4 E4B misses it under its QF contract (98.49\%), so its tasks use the Main contract. Qwen3.5-nano is the one exception: its contract reaches 97.22\%, and it is kept because its tasks measure RTL exactness and its small size allows repeated attempts (Section~\ref{sec:tasks}). Third, a reference design must pass every gate of Equation~\ref{eq:acceptance}: the public suite, the held-out suite with new weights, prefixes, cache positions, reset after activity, memory stalls, and parameter reload, synthesis, integrity, and the local checks. Qwen3.6-27B's reference fails eight public and ten held-out cases and is not an agent task. Fourth, the task fixes the starting workspace (interfaces or a scaffold), the required horizon, and the caps, and the agents attempt matched instances. Gemma 3 4B and Llama 3.2-1B passed the first three steps and became four tasks of the correctness track (Section~\ref{sec:correctness}).

\section{Correctness track: detailed results}
\label{app:correctness}
\ifcompact\begin{figure}[tbp]
\centering
\includegraphics[width=\linewidth]{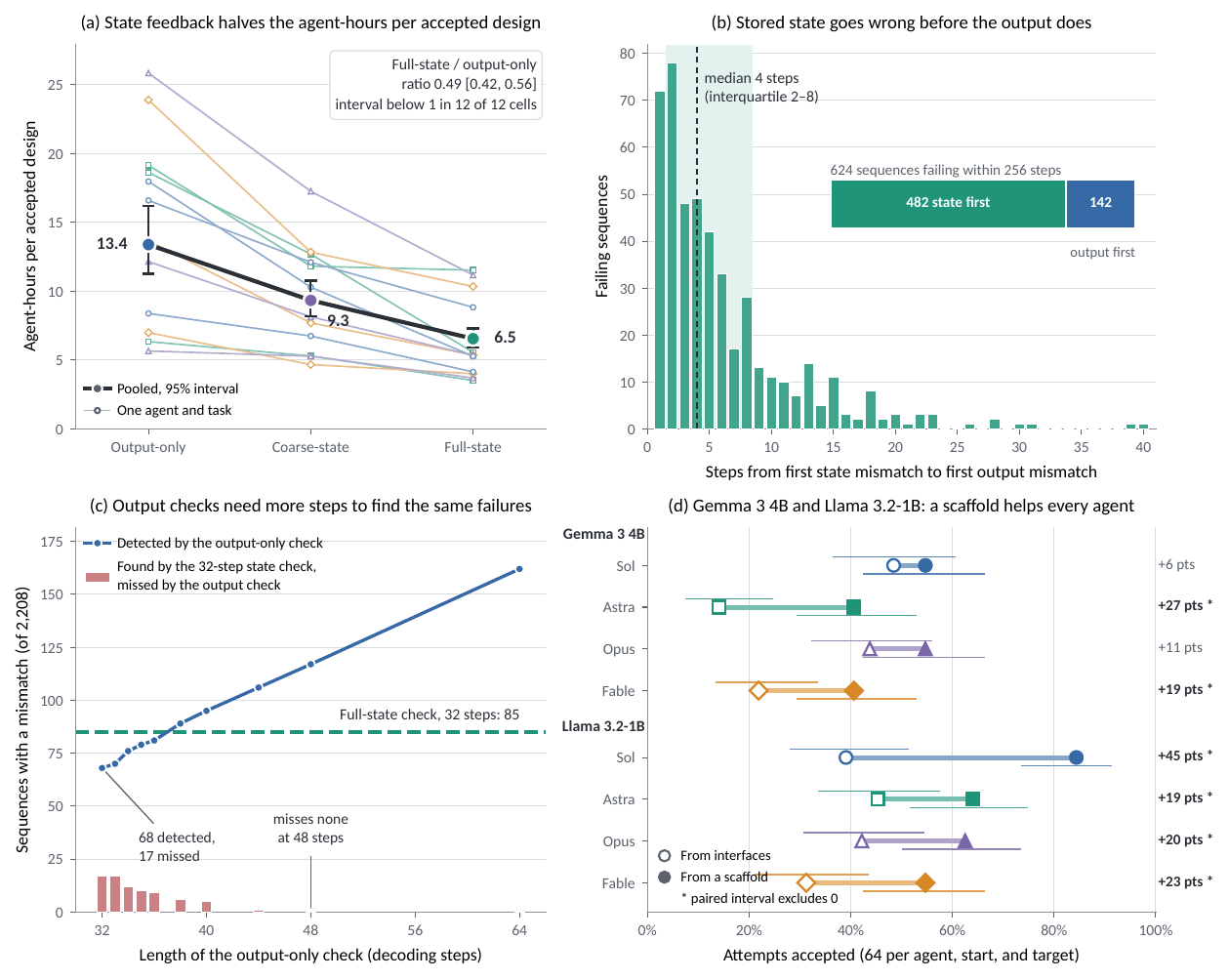}
\caption{\textbf{State feedback, state-first failures, and the starting workspace.} (a) Agent-hours per accepted design in the feedback study: pooled over the 384 attempts of each setting with block-bootstrap 95\% intervals (dark line), and for each of the 12 agent--task cells (thin lines; color and marker identify the agent). (b) Regrading of the 69 accepted submissions: steps from the first state mismatch to the first output mismatch for the 470 state-first sequences whose output also fails within 256 steps; 12 more state-first sequences show no output mismatch by step 256. (c) Sequences with a mismatch when the output-only check runs for 32 to 64 steps, against the 85 that the full-state check finds in 32 steps; bars count sequences that the state check finds and the output check misses. (d) Acceptance on the four Gemma 3 4B and Llama 3.2-1B tasks; thin lines above and below each row are the Wilson 95\% intervals from interfaces and from the scaffold. Labels give the paired difference; * marks a paired interval that excludes zero.}
\label{fig:key_state}
\end{figure}
\fi

\subsection{Scaffold adaptation and construction from interfaces}
\label{app:add-cross}
\label{sec:cross_architecture}
\begin{table}[tb]
\centering
\caption{\textbf{On Gemma 4 E4B and Llama 3.2 3B, every agent has more attempts accepted from a scaffold than from interfaces.} Final acceptance out of 24 attempts per agent, target, and start.}
\label{tab:cross_architecture}
{\small\tablefont\setlength{\tabcolsep}{4pt}\resultrows
\begin{tabular*}{0.8\linewidth}{@{\extracolsep{\fill}}llc*{4}{c}@{}}
\toprule
Target & Start & {\resultstack\begin{tabular}[b]{@{}c@{}}Reference\\FP32 agr.\end{tabular}} & Sol 5.6 & Astra 6 & Opus 5.5 & Fable 5.1 \\
\midrule
Gemma 4 E4B & Scaffold & 99.78\% & \cellcolor{blueC!35}16/24 & \cellcolor{blueC!27}12/24 & \cellcolor{blueC!37}17/24 & \cellcolor{blueC!21}9/24 \\
Gemma 4 E4B & Interfaces & 99.78\% & \cellcolor{blueC!21}9/24 & \cellcolor{blueC!10}3/24 & \cellcolor{blueC!17}7/24 & \cellcolor{blueC!16}6/24 \\
Llama 3.2 3B & Scaffold & 99.76\% & \cellcolor{blueC!37}17/24 & \cellcolor{blueC!33}15/24 & \cellcolor{blueC!35}16/24 & \cellcolor{blueC!31}14/24 \\
Llama 3.2 3B & Interfaces & 99.76\% & \cellcolor{blueC!17}7/24 & \cellcolor{blueC!14}5/24 & \cellcolor{blueC!21}9/24 & \cellcolor{blueC!23}10/24 \\
Qwen3.5-4B & Scaffold & qualified & \cellcolor{blueC!42}20/24 & \cellcolor{blueC!44}21/24 & \cellcolor{blueC!27}12/24 & \cellcolor{blueC!44}21/24 \\
\bottomrule
\end{tabular*}
}
\par\smallskip{\footnotesize\raggedright Each row uses a reference qualified as in Section~\ref{sec:contract} (Table~\ref{tab:T02-new}); Qwen3.5-4B is the reference of Section~\ref{sec:correctness}. Within a row, agents share starting files, tools, feedback, and caps. Pooled over agents, 116 of 192 attempts from a scaffold and 56 of 192 from interfaces are accepted on Gemma 4 E4B and Llama 3.2 3B. Intervals, costs, and gate outcomes: Tables~\ref{tab:T03-adaptation}--\ref{tab:T03-cost} and~\ref{tab:T04-gates-E03}.\par}
\end{table}

Five tasks of the correctness track use Gemma 4 E4B, Llama 3.2 3B, and Qwen3.5-4B: scaffold adaptation on all three targets and construction from interfaces on the first two. We call them the \emph{cross-architecture tasks}. Each agent makes 24 attempts on each (Table~\ref{tab:cross_architecture}, Figure~\ref{fig:F02_cross_architecture}, Tables~\ref{tab:T03-adaptation}--\ref{tab:T03-cost}). On Gemma 4 E4B and Llama 3.2 3B, 116 of 192 scaffold-adaptation attempts are accepted against 56 of 192 from interfaces, and every agent does better from the scaffold on both targets. On Qwen3.5-4B, between 12 of 24 (Opus 5.5) and 21 of 24 (Astra 6 and Fable 5.1) attempts are accepted. Construction from interfaces costs 53.8 to 184.0 agent-hours per accepted design, against 12.2 to 37.6 for adaptation.

Tables~\ref{tab:T04-gates-E03} and~\ref{tab:T04-intersections} follow all 480 attempts through the gates, by agent. A quarter of the attempts (120 of 480) end without an eligible submission. The public suite rejects 78 of the remaining 360 and the held-out suite 36 more, and every attempt that passes both suites also passes synthesis and the local checks. For the 36 rejected designs that pass the public suite, the grader's primary diagnosis is integration in 13, state/protocol in 13, and arithmetic in 10.

A separate set of repeated attempts measures how much outcomes vary on one instance. Each agent makes three attempts on each of eight Gemma 4 E4B and eight Llama 3.2 3B adaptation instances, 192 attempts in all, which are not counted in the 24-attempt cells above. Of the 64 agent--instance groups, all three attempts are accepted in 27 and none in 11; in the other 26 (41\%), one or two are accepted (Figure~\ref{fig:F02_cross_architecture}d). 

\begin{figure}[htbp]
\centering
\includegraphics[width=\linewidth,height=.62\textheight,keepaspectratio]{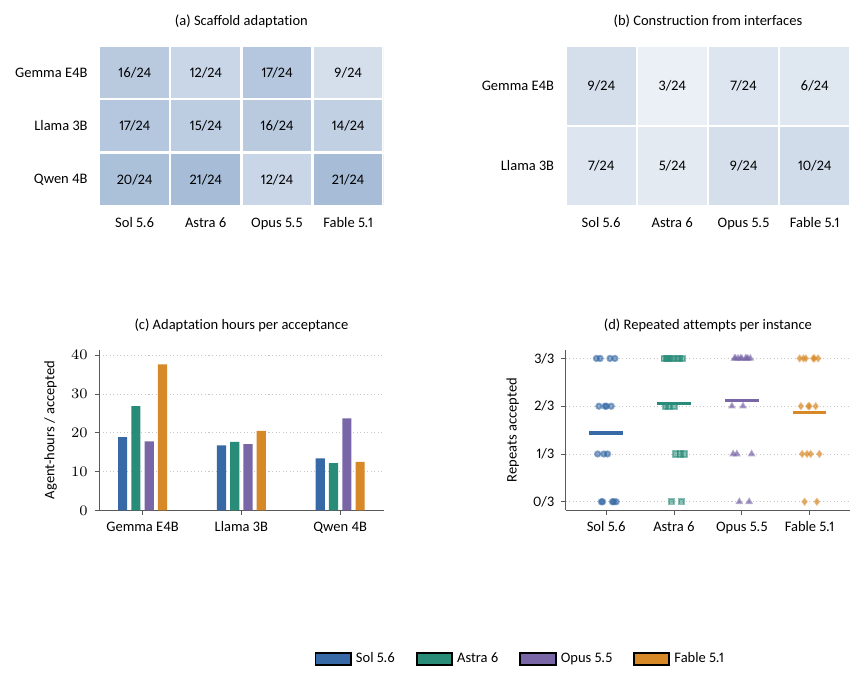}
\caption{Cross-architecture tasks with 24 attempts per agent, target, and start. (a) Final acceptance after scaffold adaptation. (b) Final acceptance when construction starts from interfaces. (c) Agent-hours per accepted design over all adaptation attempts. (d) Repeated attempts on Gemma 4 E4B and Llama 3.2 3B: each point is the share of 3 repeats accepted on one instance, and bars mark the means. These 192 repeated attempts are separate from those in (a)--(c) and stay out of their denominators.}
\label{fig:F02_cross_architecture}
\end{figure}

\begin{table}[htbp]
\centering
\caption{\textbf{Cross-architecture tasks: scaffold adaptation, with Wilson 95\% intervals.}}
\label{tab:T03-adaptation}
{\small\tablefont\setlength{\tabcolsep}{7pt}\resultrows
\begin{tabular}{@{}llrrr}
\toprule
Target & Agent & Public (\%) & Final (\%) & h/accepted \\
\midrule
Gemma 4 E4B & Sol 5.6 & 17/24 (70.8) & \cellcolor{blueC!35}16/24 (66.7)\,\textsubscript{\tiny(46.7--82.0)} & 18.9 \\
Gemma 4 E4B & Astra 6 & 15/24 (62.5) & \cellcolor{blueC!27}12/24 (50.0)\,\textsubscript{\tiny(31.4--68.6)} & 26.9 \\
Gemma 4 E4B & Opus 5.5 & 18/24 (75.0) & \cellcolor{blueC!37}17/24 (70.8)\,\textsubscript{\tiny(50.8--85.1)} & 17.8 \\
Gemma 4 E4B & Fable 5.1 & 13/24 (54.2) & \cellcolor{blueC!21}9/24 (37.5)\,\textsubscript{\tiny(21.2--57.3)} & 37.6 \\
Llama 3B & Sol 5.6 & 17/24 (70.8) & \cellcolor{blueC!37}17/24 (70.8)\,\textsubscript{\tiny(50.8--85.1)} & 16.7 \\
Llama 3B & Astra 6 & 17/24 (70.8) & \cellcolor{blueC!33}15/24 (62.5)\,\textsubscript{\tiny(42.7--78.8)} & 17.6 \\
Llama 3B & Opus 5.5 & 17/24 (70.8) & \cellcolor{blueC!35}16/24 (66.7)\,\textsubscript{\tiny(46.7--82.0)} & 17.1 \\
Llama 3B & Fable 5.1 & 14/24 (58.3) & \cellcolor{blueC!31}14/24 (58.3)\,\textsubscript{\tiny(38.8--75.5)} & 20.5 \\
Qwen 4B & Sol 5.6 & 22/24 (91.7) & \cellcolor{blueC!42}20/24 (83.3)\,\textsubscript{\tiny(64.1--93.3)} & 13.4 \\
Qwen 4B & Astra 6 & 23/24 (95.8) & \cellcolor{blueC!44}21/24 (87.5)\,\textsubscript{\tiny(69.0--95.7)} & 12.2 \\
Qwen 4B & Opus 5.5 & 17/24 (70.8) & \cellcolor{blueC!27}12/24 (50.0)\,\textsubscript{\tiny(31.4--68.6)} & 23.7 \\
Qwen 4B & Fable 5.1 & 23/24 (95.8) & \cellcolor{blueC!44}21/24 (87.5)\,\textsubscript{\tiny(69.0--95.7)} & 12.5 \\
\bottomrule
\end{tabular}
}
\par\smallskip{\footnotesize\raggedright Each target and agent has 24 task instances. Within a task, all agents receive the same starting files, tools, feedback, and caps on wall time, tool calls, and output tokens. Every target has a qualified reference.\par}
\end{table}
\begin{table}[htbp]
\centering
\caption{\textbf{Cross-architecture tasks: construction from interfaces, with Wilson 95\% intervals.}}
\label{tab:T03-interface}
{\small\tablefont\setlength{\tabcolsep}{7pt}\resultrows
\begin{tabular}{@{}llrrr}
\toprule
Target & Agent & Public (\%) & Final (\%) & h/accepted \\
\midrule
Gemma 4 E4B & Sol 5.6 & 11/24 (45.8) & \cellcolor{blueC!21}9/24 (37.5)\,\textsubscript{\tiny(21.2--57.3)} & 62.0 \\
Gemma 4 E4B & Astra 6 & 6/24 (25.0) & \cellcolor{blueC!10}3/24 (12.5)\,\textsubscript{\tiny(4.3--31.0)} & 184.0 \\
Gemma 4 E4B & Opus 5.5 & 7/24 (29.2) & \cellcolor{blueC!17}7/24 (29.2)\,\textsubscript{\tiny(14.9--49.2)} & 80.7 \\
Gemma 4 E4B & Fable 5.1 & 9/24 (37.5) & \cellcolor{blueC!16}6/24 (25.0)\,\textsubscript{\tiny(12.0--44.9)} & 92.7 \\
Llama 3B & Sol 5.6 & 7/24 (29.2) & \cellcolor{blueC!17}7/24 (29.2)\,\textsubscript{\tiny(14.9--49.2)} & 75.2 \\
Llama 3B & Astra 6 & 9/24 (37.5) & \cellcolor{blueC!14}5/24 (20.8)\,\textsubscript{\tiny(9.2--40.5)} & 94.5 \\
Llama 3B & Opus 5.5 & 10/24 (41.7) & \cellcolor{blueC!21}9/24 (37.5)\,\textsubscript{\tiny(21.2--57.3)} & 59.4 \\
Llama 3B & Fable 5.1 & 10/24 (41.7) & \cellcolor{blueC!23}10/24 (41.7)\,\textsubscript{\tiny(24.5--61.2)} & 53.8 \\
\bottomrule
\end{tabular}
}
\par\smallskip{\footnotesize\raggedright Each target and agent has 24 task instances. Within a task, all agents receive the same starting files, tools, feedback, and caps on wall time, tool calls, and output tokens. Every target has a qualified reference.\par}
\end{table}
\begin{table}[htbp]
\centering
\caption{\textbf{Resource use and state/protocol diagnoses on the cross-architecture tasks.}}
\label{tab:T03-cost}
{\small\tablefont\setlength{\tabcolsep}{4pt}\resultrows
\begin{tabular*}{\linewidth}{@{\extracolsep{\fill}}lllrrrr@{}}
\toprule
Target & Start & Agent & Calls mean & Output k & Reject h \% & State diag. \\
\midrule
Gemma 4 E4B & adapt. & Sol 5.6 & 389.1 & 180.3 & 39.8 & 1 \\
Gemma 4 E4B & adapt. & Astra 6 & 396.5 & 183.1 & 48.4 & 3 \\
Gemma 4 E4B & adapt. & Opus 5.5 & 373.0 & 183.9 & 33.6 & 2 \\
Gemma 4 E4B & adapt. & Fable 5.1 & 363.5 & 174.8 & 69.7 & 3 \\
Llama 3B & adapt. & Sol 5.6 & 347.8 & 165.8 & 35.2 & 1 \\
Llama 3B & adapt. & Astra 6 & 374.3 & 169.5 & 41.2 & 1 \\
Llama 3B & adapt. & Opus 5.5 & 358.3 & 177.1 & 39.1 & 1 \\
Llama 3B & adapt. & Fable 5.1 & 302.8 & 155.0 & 51.9 & 0 \\
Qwen 4B & adapt. & Sol 5.6 & 345.2 & 196.5 & 21.1 & 0 \\
Qwen 4B & adapt. & Astra 6 & 339.7 & 219.8 & 15.2 & 0 \\
Qwen 4B & adapt. & Opus 5.5 & 329.7 & 209.6 & 52.6 & 3 \\
Qwen 4B & adapt. & Fable 5.1 & 347.3 & 192.2 & 12.2 & 2 \\
Gemma 4 E4B & interface & Sol 5.6 & 586.7 & 314.5 & 70.6 & 1 \\
Gemma 4 E4B & interface & Astra 6 & 551.5 & 331.2 & 90.8 & 4 \\
Gemma 4 E4B & interface & Opus 5.5 & 600.2 & 336.7 & 76.3 & 1 \\
Gemma 4 E4B & interface & Fable 5.1 & 602.5 & 355.4 & 78.0 & 2 \\
Llama 3B & interface & Sol 5.6 & 573.4 & 335.5 & 74.0 & 2 \\
Llama 3B & interface & Astra 6 & 615.2 & 320.7 & 80.2 & 2 \\
Llama 3B & interface & Opus 5.5 & 572.0 & 328.9 & 66.3 & 2 \\
Llama 3B & interface & Fable 5.1 & 593.4 & 294.3 & 59.0 & 2 \\
\bottomrule
\end{tabular*}
}
\par\smallskip{\footnotesize\raggedright State diag.\ counts rejected attempts whose primary diagnosis is state/protocol. Resource means include every attempt.\par}
\end{table}

\begin{table}[htbp]
\centering
\caption{\textbf{Marginal gate outcomes, cross-architecture tasks.}}
\label{tab:T04-gates-E03}
{\small\tablefont\setlength{\tabcolsep}{4pt}\resultrows
\begin{tabular*}{\linewidth}{@{\extracolsep{\fill}}lllrrrrrrrr@{}}
\toprule
Task & Start & Agent & $n$ & Elig. & Public & Held out & Synth. & Integr. & Local & Final \\
\midrule
Gemma E4B & adapt. & Sol 5.6 & 24 & 20 & 17 & 19 & 20 & 20 & 20 & 16 \\
Gemma E4B & adapt. & Astra 6 & 24 & 18 & 15 & 13 & 18 & 18 & 17 & 12 \\
Gemma E4B & adapt. & Opus 5.5 & 24 & 19 & 18 & 17 & 19 & 19 & 19 & 17 \\
Gemma E4B & adapt. & Fable 5.1 & 24 & 17 & 13 & 9 & 15 & 17 & 16 & 9 \\
Gemma E4B & interface & Sol 5.6 & 24 & 15 & 11 & 10 & 15 & 15 & 14 & 9 \\
Gemma E4B & interface & Astra 6 & 24 & 15 & 6 & 10 & 15 & 15 & 14 & 3 \\
Gemma E4B & interface & Opus 5.5 & 24 & 15 & 7 & 11 & 14 & 15 & 11 & 7 \\
Gemma E4B & interface & Fable 5.1 & 24 & 15 & 9 & 10 & 14 & 15 & 14 & 6 \\
Llama 3B & adapt. & Sol 5.6 & 24 & 20 & 17 & 17 & 20 & 20 & 20 & 17 \\
Llama 3B & adapt. & Astra 6 & 24 & 18 & 17 & 16 & 18 & 18 & 18 & 15 \\
Llama 3B & adapt. & Opus 5.5 & 24 & 20 & 17 & 18 & 20 & 20 & 20 & 16 \\
Llama 3B & adapt. & Fable 5.1 & 24 & 16 & 14 & 16 & 16 & 16 & 16 & 14 \\
Llama 3B & interface & Sol 5.6 & 24 & 13 & 7 & 9 & 13 & 13 & 13 & 7 \\
Llama 3B & interface & Astra 6 & 24 & 16 & 9 & 8 & 16 & 16 & 16 & 5 \\
Llama 3B & interface & Opus 5.5 & 24 & 17 & 10 & 13 & 17 & 17 & 17 & 9 \\
Llama 3B & interface & Fable 5.1 & 24 & 16 & 10 & 12 & 15 & 16 & 16 & 10 \\
Qwen 4B & adapt. & Sol 5.6 & 24 & 22 & 22 & 20 & 22 & 22 & 22 & 20 \\
Qwen 4B & adapt. & Astra 6 & 24 & 23 & 23 & 21 & 23 & 23 & 23 & 21 \\
Qwen 4B & adapt. & Opus 5.5 & 24 & 21 & 17 & 14 & 21 & 21 & 21 & 12 \\
Qwen 4B & adapt. & Fable 5.1 & 24 & 24 & 23 & 22 & 24 & 24 & 24 & 21 \\
\bottomrule
\end{tabular*}
}
\par\smallskip{\footnotesize\raggedright Entries count passes among all $n$ attempts. Eligibility failures are counted apart from numerical failures, and marginal counts need not be nested. Local checks are required on all of these tasks.\par}
\end{table}
\begin{table}[htbp]
\centering
\caption{\textbf{Cumulative gate intersections by agent.}}
\label{tab:T04-intersections}
{\small\tablefont\setlength{\tabcolsep}{4pt}\resultrows
\begin{tabular*}{\linewidth}{@{\extracolsep{\fill}}llrrrrrrr@{}}
\toprule
Tasks & Agent & $n$ & Eligible & +Integrity & +Public & +Held out & +Synthesis & +Local \\
\midrule
Earlier series & Opus 5.5 & 60 & 45 & 45 & 42 & 36 & 36 & 36 \\
Earlier series & Fable 5.1 & 60 & 53 & 53 & 46 & 45 & 45 & 45 \\
Cross-arch. & Sol 5.6 & 120 & 90 & 90 & 74 & 69 & 69 & 69 \\
Cross-arch. & Astra 6 & 120 & 90 & 90 & 70 & 56 & 56 & 56 \\
Cross-arch. & Opus 5.5 & 120 & 92 & 92 & 69 & 61 & 61 & 61 \\
Cross-arch. & Fable 5.1 & 120 & 88 & 88 & 69 & 60 & 60 & 60 \\
\bottomrule
\end{tabular*}
}
\par\smallskip{\footnotesize\raggedright Each column adds one gate to those on its left, so counts can only fall from left to right. The last column equals final acceptance. Earlier-series rows cover Opus 5.5 and Fable 5.1 on T1--T3 and Qwen3.5-0.8B adaptation.\par}
\end{table}

\subsection{Gemma 3 4B and Llama 3.2-1B tasks}
\label{app:gemma3_llama1b}
Four tasks of the correctness track use Gemma 3 4B (text) and Llama 3.2-1B (Section~\ref{sec:tasks}), each from a scaffold and from interfaces. They require 32 decoding steps and have 1,600 public and 240 held-out cases each. Each agent makes one attempt per task on each of 64 instances of a target, and the eight attempts on one target instance (four agents, two starts) form a matched block: 1,024 attempts in 128 blocks (Figure~\ref{fig:F07_gemma3_llama1b_tasks}, Table~\ref{tab:T10_gemma3_llama1b}). The two starts carry different caps, 16 to 18 hours, 560 to 600 tool calls, and 340k to 360k output tokens from a scaffold against 28 to 30 hours, 860 to 900 calls, and 560k to 600k tokens from interfaces, so a contrast between starts compares complete protocols.

Pooled over agents, 122 of 256 scaffold attempts are accepted on Gemma 3 4B and 170 on Llama 3.2-1B, against 82 and 101 from interfaces, at 29.0 and 18.2 agent-hours per accepted design from a scaffold against 72.5 and 55.9 from interfaces. Every agent has more attempts accepted from the scaffold on both targets. The paired difference is 6.3 to 45.3 points, and six of the eight intervals exclude zero, three after a Holm adjustment. Caps end 326 of the 1,024 attempts: 205 on wall time, 76 on tool calls, and 45 on output tokens. Sol 5.6 has the most accepted attempts on Gemma 3 4B from interfaces and on Llama 3.2-1B from a scaffold and ties Opus 5.5 on Gemma 3 4B from a scaffold; Astra 6 has the most on Llama 3.2-1B from interfaces. Four of the 24 pairwise agent contrasts stay below 0.05 after a Holm adjustment, three of them on Gemma 3 4B from interfaces, where Astra 6 and Fable 5.1 are accepted on 9 and 14 of 64 instances against 28 and 31 for Opus 5.5 and Sol 5.6.

The same 128 blocks were first run with eight required decoding steps and suites of 1,320 public and 80 held-out cases, under the same caps (1,024 further attempts, Table~\ref{tab:horizon_comparison}). These runs are kept as a horizon comparison and do not enter the correctness track. Pooled acceptance is 558 of 1,024 at eight steps and 475 at 32. It falls in 14 of the 16 agent--task cells, most for Sol 5.6 from the Gemma 3 4B scaffold (48 to 35 of 64), and rises for Astra 6 on both Llama 3.2-1B tasks. Caps end 263 attempts at eight steps and 326 at 32, and the mean agent time per attempt rises from 12.8 to 17.8 hours.

\begin{table}[tbp]
\centering
\caption{\textbf{Gemma 3 4B and Llama 3.2-1B tasks at eight and at 32 required decoding steps.} The same 64 instances per task under both protocols; pooled, 558 and 475 of 1,024 attempts are accepted.}
\label{tab:horizon_comparison}
{\small\tablefont\setlength{\tabcolsep}{3pt}\renewcommand{\arraystretch}{1.04}
\begin{tabular*}{\linewidth}{@{\extracolsep{\fill}}llrrrrrrr@{}}
\toprule
\multicolumn{2}{c}{} & \multicolumn{3}{c}{Accepted} & \multicolumn{4}{c}{8 / 32 steps} \\
\cmidrule(lr){3-5}\cmidrule(lr){6-9}
Task & Agent & 8 steps & 32 steps & Points & \shortstack{Accepted\\at one only} & Caps & \shortstack{Mean\\hours} & \shortstack{Holm\\$p$} \\
\midrule
G3 / S & Sol 5.6 & 48/64 & 35/64 & -20.3 & 17 / 4 & 6 / 15 & 10.8 / 13.5 & 0.108 \\
G3 / S & Astra 6 & 34/64 & 26/64 & -12.5 & 12 / 4 & 18 / 27 & 8.5 / 14.2 & 0.999 \\
G3 / S & Opus 5.5 & 42/64 & 35/64 & -10.9 & 12 / 5 & 17 / 19 & 10.1 / 13.7 & 1.000 \\
G3 / S & Fable 5.1 & 37/64 & 26/64 & -17.2 & 12 / 1 & 11 / 20 & 10.1 / 13.9 & 0.055 \\
G3 / I & Sol 5.6 & 36/64 & 31/64 & -7.8 & 12 / 7 & 16 / 16 & 15.6 / 22.4 & 1.000 \\
G3 / I & Astra 6 & 15/64 & 9/64 & -9.4 & 10 / 4 & 18 / 19 & 16.2 / 22.6 & 1.000 \\
G3 / I & Opus 5.5 & 32/64 & 28/64 & -6.2 & 16 / 12 & 21 / 22 & 16.2 / 23.4 & 1.000 \\
G3 / I & Fable 5.1 & 19/64 & 14/64 & -7.8 & 13 / 8 & 25 / 36 & 17.3 / 24.6 & 1.000 \\
\addlinespace[2pt]
L1 / S & Sol 5.6 & 55/64 & 54/64 & -1.6 & 6 / 5 & 7 / 9 & 9.4 / 12.0 & 1.000 \\
L1 / S & Astra 6 & 37/64 & 41/64 & +6.2 & 10 / 14 & 17 / 14 & 9.7 / 12.4 & 1.000 \\
L1 / S & Opus 5.5 & 43/64 & 40/64 & -4.7 & 15 / 12 & 6 / 5 & 8.2 / 11.4 & 1.000 \\
L1 / S & Fable 5.1 & 42/64 & 35/64 & -10.9 & 13 / 6 & 22 / 27 & 10.8 / 12.4 & 1.000 \\
L1 / I & Sol 5.6 & 28/64 & 25/64 & -4.7 & 10 / 7 & 20 / 26 & 16.0 / 22.4 & 1.000 \\
L1 / I & Astra 6 & 28/64 & 29/64 & +1.6 & 9 / 10 & 20 / 18 & 14.6 / 21.4 & 1.000 \\
L1 / I & Opus 5.5 & 33/64 & 27/64 & -9.4 & 15 / 9 & 13 / 21 & 15.3 / 22.0 & 1.000 \\
L1 / I & Fable 5.1 & 29/64 & 20/64 & -14.1 & 14 / 5 & 26 / 32 & 16.4 / 22.4 & 0.890 \\
\bottomrule
\end{tabular*}}
\par\smallskip{\footnotesize\raggedright G3: Gemma 3 4B text; L1: Llama 3.2-1B. S: scaffold; I: interfaces. The 8-step protocol uses 1,320 public and 80 held-out cases per task, the 32-step protocol 1,600 and 240; caps are the same. Each attempt is a separate run, so a change compares the complete protocols. Accepted at one only: instances accepted at eight steps and not at 32, and the reverse. Holm $p$: exact McNemar test on the matched instances, adjusted over the 16 cells.\par}
\end{table}

\begin{figure}[tbp]
\centering
\includegraphics[width=\linewidth]{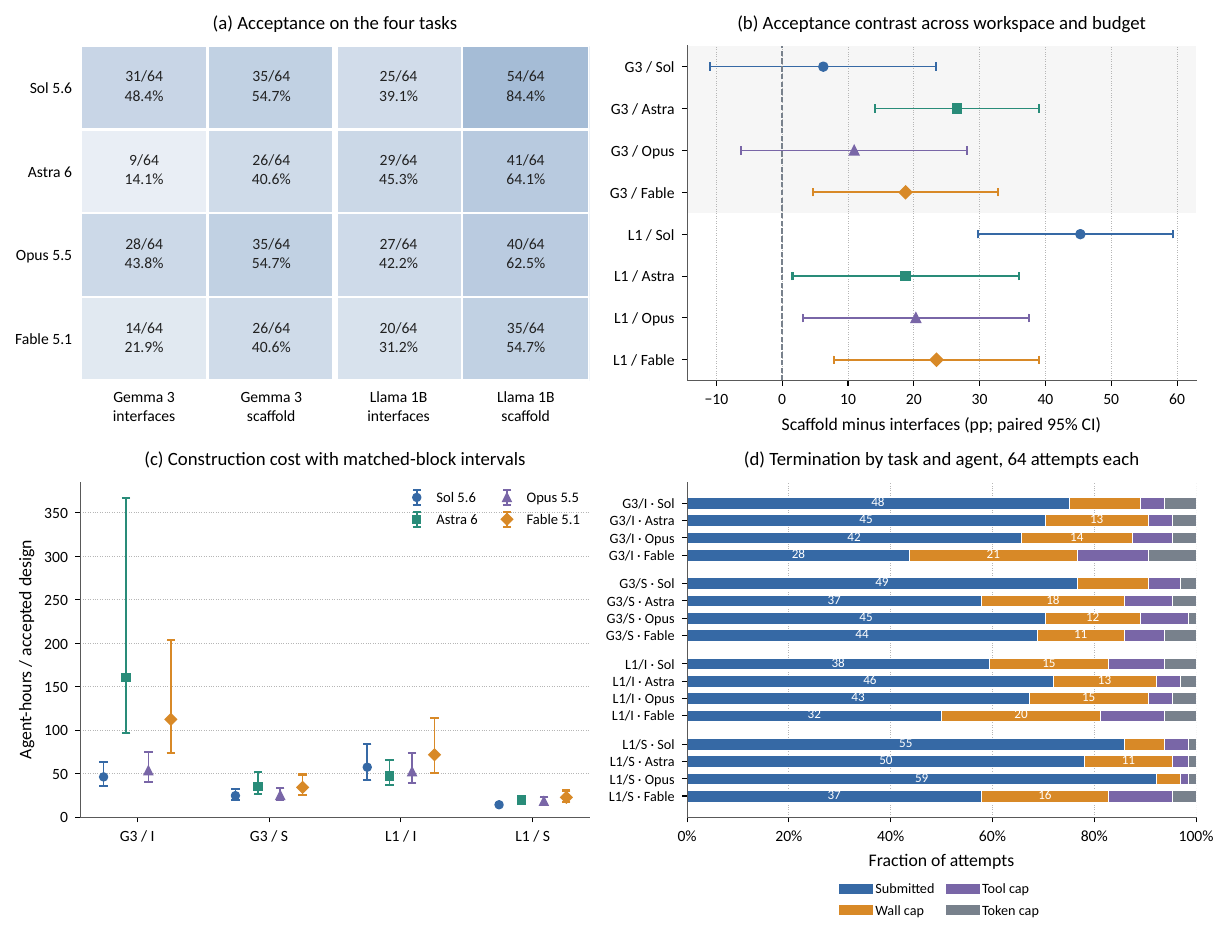}
\caption{The four tasks on Gemma 3 4B (text) and Llama 3.2-1B. (a) Accepted attempts per agent and task, 64 matched instances each. (b) Scaffold minus interfaces acceptance with paired intervals; the two starts have different caps. (c) Agent-hours per accepted design with matched-block intervals. (d) How the 64 attempts of each agent and task end.}
\label{fig:F07_gemma3_llama1b_tasks}
\end{figure}
\begin{table}[H]
\centering
\caption{\textbf{Gemma 3 4B and Llama 3.2-1B tasks: acceptance with Wilson 95\% intervals, cost with paired bootstrap 95\% intervals, and cap exits by agent.}}
\label{tab:T10_gemma3_llama1b}
{\small\tablefont\setlength{\tabcolsep}{3.5pt}\renewcommand{\arraystretch}{1.04}
\begin{tabular*}{\linewidth}{@{\extracolsep{\fill}}llrrrr@{}}
\toprule
\multicolumn{2}{c}{} & \multicolumn{2}{c}{Correctness outcomes} & \multicolumn{2}{c}{Construction cost and caps} \\
\cmidrule(lr){3-4}\cmidrule(lr){5-6}
Task & Agent & Accepted & \shortstack{Acceptance\\\%} & \shortstack{Hours /\\accepted} & Caps \\
\midrule
G3 / I & Fable 5.1 & 14/64 & 21.9\,\textsubscript{\tiny(13.5--33.4)} & 112.4\,\textsubscript{\tiny(73.8--203.5)} & 36 \\
G3 / I & Sol 5.6 & 31/64 & 48.4\,\textsubscript{\tiny(36.6--60.4)} & 46.2\,\textsubscript{\tiny(36.0--63.6)} & 16 \\
G3 / I & Astra 6 & 9/64 & 14.1\,\textsubscript{\tiny(7.6--24.6)} & 160.5\,\textsubscript{\tiny(96.7--367.4)} & 19 \\
G3 / I & Opus 5.5 & 28/64 & 43.8\,\textsubscript{\tiny(32.3--55.9)} & 53.4\,\textsubscript{\tiny(40.7--75.2)} & 22 \\
\addlinespace[3pt]
G3 / S & Fable 5.1 & 26/64 & 40.6\,\textsubscript{\tiny(29.5--52.9)} & 34.3\,\textsubscript{\tiny(25.9--49.0)} & 20 \\
G3 / S & Sol 5.6 & 35/64 & 54.7\,\textsubscript{\tiny(42.6--66.3)} & 24.7\,\textsubscript{\tiny(19.8--32.7)} & 15 \\
G3 / S & Astra 6 & 26/64 & 40.6\,\textsubscript{\tiny(29.5--52.9)} & 35.0\,\textsubscript{\tiny(26.2--51.4)} & 27 \\
G3 / S & Opus 5.5 & 35/64 & 54.7\,\textsubscript{\tiny(42.6--66.3)} & 25.0\,\textsubscript{\tiny(19.9--33.2)} & 19 \\
\addlinespace[3pt]
L1 / I & Fable 5.1 & 20/64 & 31.2\,\textsubscript{\tiny(21.2--43.4)} & 71.8\,\textsubscript{\tiny(51.0--113.5)} & 32 \\
L1 / I & Sol 5.6 & 25/64 & 39.1\,\textsubscript{\tiny(28.1--51.3)} & 57.4\,\textsubscript{\tiny(42.2--83.6)} & 26 \\
L1 / I & Astra 6 & 29/64 & 45.3\,\textsubscript{\tiny(33.7--57.4)} & 47.2\,\textsubscript{\tiny(36.5--65.6)} & 18 \\
L1 / I & Opus 5.5 & 27/64 & 42.2\,\textsubscript{\tiny(30.9--54.4)} & 52.2\,\textsubscript{\tiny(39.5--73.9)} & 21 \\
\addlinespace[3pt]
L1 / S & Fable 5.1 & 35/64 & 54.7\,\textsubscript{\tiny(42.6--66.3)} & 22.7\,\textsubscript{\tiny(17.8--30.6)} & 27 \\
L1 / S & Sol 5.6 & 54/64 & 84.4\,\textsubscript{\tiny(73.6--91.3)} & 14.2\,\textsubscript{\tiny(12.5--16.5)} & 9 \\
L1 / S & Astra 6 & 41/64 & 64.1\,\textsubscript{\tiny(51.8--74.7)} & 19.4\,\textsubscript{\tiny(16.1--24.2)} & 14 \\
L1 / S & Opus 5.5 & 40/64 & 62.5\,\textsubscript{\tiny(50.3--73.3)} & 18.2\,\textsubscript{\tiny(15.2--22.8)} & 5 \\
\bottomrule
\end{tabular*}}
\par\smallskip{\footnotesize\raggedright G3: Gemma 3 4B text; L1: Llama 3.2-1B. I: interfaces; S: scaffold. Each task uses 32 decode steps. Starts have different resource caps, so contrasts compare complete protocols. Cost intervals resample instances. The 1,024 attempts span 128 underlying matched target-instance blocks.\par}
\end{table}

\subsection{Hardware of accepted correctness-track designs}
\label{app:add-hardware}
All 1,016 designs accepted on the twelve tasks of the correctness track, the 246 of the cross-architecture tasks, the 295 full-state designs of the feedback study, and the 475 of the Gemma 3 4B and Llama 3.2-1B tasks, were synthesized, and three accepted designs per agent and task, 144 in all, were placed and routed with three seeds each (Figure~\ref{fig:F05_correctness_hardware}, Table~\ref{tab:T07_correctness_hardware}). Each ratio divides by a reference at the same implementation stage and workload. At synthesis, the median LUT ratio per task is 0.97 to 1.10. Construction from interfaces gives both the largest ratios, 1.08 on Gemma 4 E4B and 1.10 on Llama 3.2 3B, and the smallest, 0.97 on Gemma 3 4B and 0.98 on Llama 3.2-1B. Of the 432 route runs, 421 complete, and 134 of the 144 designs complete all three. Among complete designs, the median routed throughput per task is 0.77 to 1.02 times the reference's, lowest from interfaces on Llama 3.2 3B and Gemma 4 E4B (0.77 and 0.81). Between 11 and 33 of each task's 36 route runs close timing at 126\,MHz, fewest on Qwen3.5-4B. Nano throughput uses one transaction from reset. 

\begin{figure}[tbp]
\centering
\includegraphics[width=\linewidth]{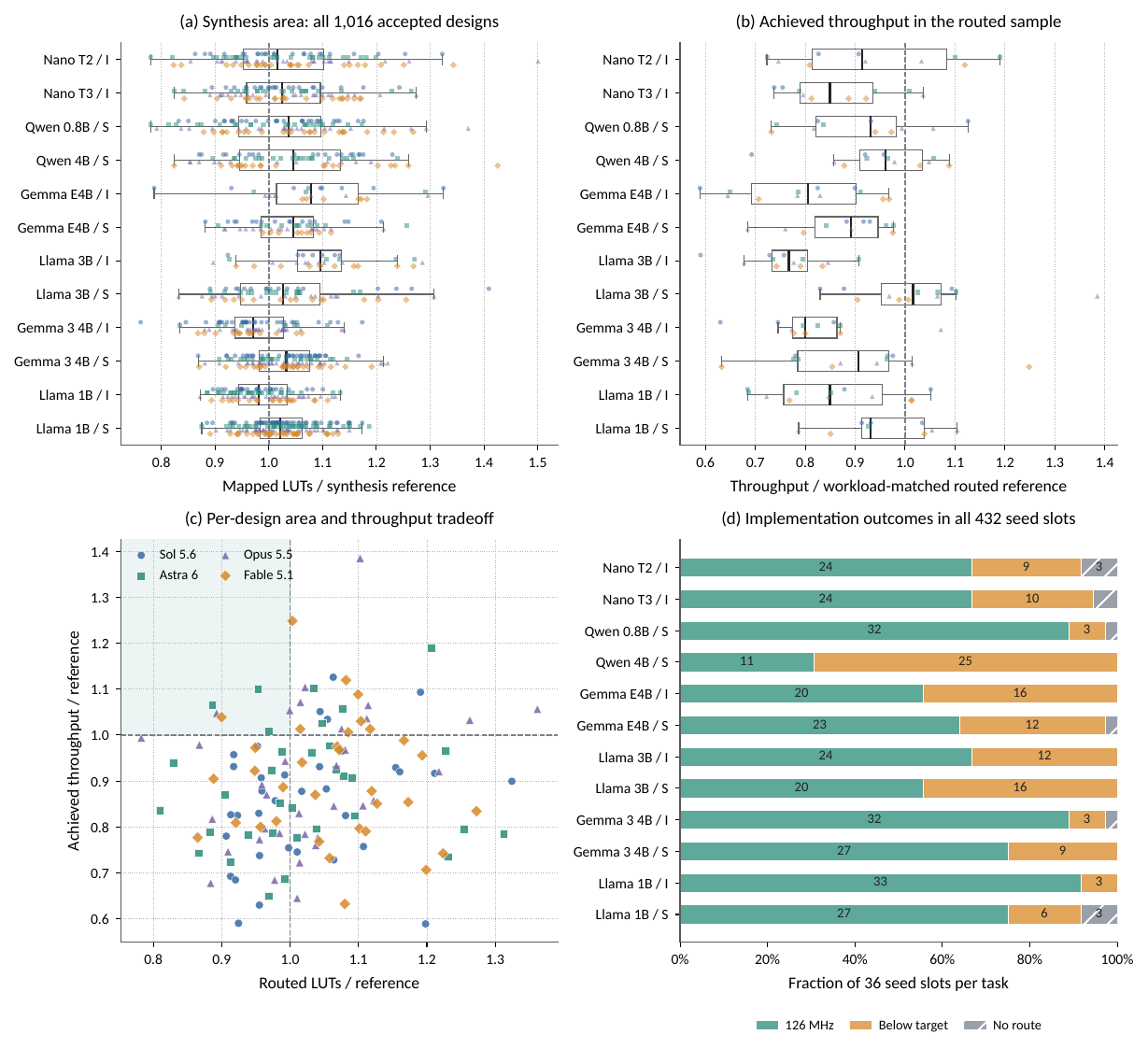}
\caption{Hardware of accepted correctness-track designs. Every synthesis point and every routed design that completes all three seeds is shown; vertical lanes identify agents. Boxes show medians and quartiles, with whiskers to 1.5 interquartile ranges. The route sample has three designs per agent and task, and all 432 route runs appear in the implementation-outcome panel. Ratios use references at the same stage and workload. Nano rates come from single-transaction probes.}
\label{fig:F05_correctness_hardware}
\end{figure}
\begin{table}[H]
\centering
\caption{\textbf{Hardware quality conditional on correctness-track acceptance.}}
\label{tab:T07_correctness_hardware}
{\small\tablefont\setlength{\tabcolsep}{3.5pt}\renewcommand{\arraystretch}{1.04}
\begin{tabular*}{\linewidth}{@{\extracolsep{\fill}}llrrrrrr@{}}
\toprule
\multicolumn{2}{c}{} & \multicolumn{2}{c}{Accepted-design census} & \multicolumn{4}{c}{Balanced implementation sample} \\
\cmidrule(lr){3-4}\cmidrule(lr){5-8}
Task & Start & Synth. & \shortstack{Synthesis\\LUT ratio} & \shortstack{Complete\\sample} & \shortstack{Routed\\rate ratio} & \shortstack{Routed\\LUT ratio} & \shortstack{126 MHz\\slots} \\
\midrule
Gemma 3 4B & S & 122 & 1.03 & 12/12 & 0.91 & 1.00 & 27/36 \\
Gemma 3 4B & I & 82 & 0.97 & 11/12 & 0.80 & 0.98 & 32/36 \\
Gemma E4B & S & 54 & 1.05 & 11/12 & 0.89 & 1.05 & 23/36 \\
Gemma E4B & I & 25 & 1.08 & 12/12 & 0.81 & 1.08 & 20/36 \\
Llama 3B & S & 62 & 1.03 & 12/12 & 1.02 & 1.06 & 20/36 \\
Llama 3B & I & 31 & 1.10 & 12/12 & 0.77 & 1.10 & 24/36 \\
Llama 1B & S & 170 & 1.02 & 9/12 & 0.93 & 1.02 & 27/36 \\
Llama 1B & I & 101 & 0.98 & 12/12 & 0.85 & 1.01 & 33/36 \\
Qwen 0.8B & S & 101 & 1.04 & 11/12 & 0.93 & 0.92 & 32/36 \\
Qwen 4B & S & 74 & 1.05 & 12/12 & 0.96 & 1.09 & 11/36 \\
Nano T2 & I & 103 & 1.02 & 10/12 & 0.91 & 0.96 & 24/36 \\
Nano T3 & I & 91 & 1.02 & 10/12 & 0.85 & 0.97 & 24/36 \\
\bottomrule
\end{tabular*}}
\par\smallskip{\footnotesize\raggedright I: interfaces; S: scaffold. Ratios use references at the same implementation stage and workload; medians condition on complete routes. Each task has 12 sampled designs and 36 seed slots. Nano rates use one matched transaction from reset.\par}
\end{table}

\subsection{Feedback study}
\label{app:add-feedback}
\begin{table}[tb]
\centering
\caption{\textbf{Feedback that locates the first state mismatch raises acceptance over output-only feedback in all 12 agent--task cells.} Accepted attempts out of 32 matched blocks per setting, and the full-state minus output-only difference with paired bootstrap 95\% intervals; within a block, the three settings share the task instance, starting files, final evaluator, and caps.}
\label{tab:feedback_repeated}
{\small\tablefont\setlength{\tabcolsep}{4pt}\resultrows
\begin{tabular*}{0.8\linewidth}{@{\extracolsep{\fill}}llccccc@{}}
\toprule
Task & Agent & {\resultstack\begin{tabular}[b]{@{}c@{}}Output\\only\end{tabular}} & {\resultstack\begin{tabular}[b]{@{}c@{}}Coarse\\state\end{tabular}} & {\resultstack\begin{tabular}[b]{@{}c@{}}Full\\state\end{tabular}} & {\resultstack\begin{tabular}[b]{@{}c@{}}Full $-$ output\\(pp)\end{tabular}} & {\resultstack\begin{tabular}[b]{@{}c@{}}Discordant\\full/output\end{tabular}} \\
\midrule
Nano T2 & Sol 5.6 & \cellcolor{blueC!21}12/32 & \cellcolor{blueC!27}16/32 & \cellcolor{blueC!38}24/32 & \cellcolor{tealC!38}\textbf{+37.5}\,\textsubscript{\tiny(15.6--59.4)} & 15/3 \\
 & Astra 6 & \cellcolor{blueC!28}17/32 & \cellcolor{blueC!33}20/32 & \cellcolor{blueC!44}28/32 & \cellcolor{tealC!35}\textbf{+34.4}\,\textsubscript{\tiny(15.6--53.1)} & 12/1 \\
 & Opus 5.5 & \cellcolor{blueC!30}18/32 & \cellcolor{blueC!31}19/32 & \cellcolor{blueC!41}26/32 & \cellcolor{tealC!27}\textbf{+25.0}\,\textsubscript{\tiny(6.2--43.8)} & 10/2 \\
 & Fable 5.1 & \cellcolor{blueC!26}15/32 & \cellcolor{blueC!36}22/32 & \cellcolor{blueC!40}25/32 & \cellcolor{tealC!32}\textbf{+31.2}\,\textsubscript{\tiny(9.4--50.0)} & 12/2 \\
\addlinespace[2pt]
Nano T3 & Sol 5.6 & \cellcolor{blueC!14}7/32 & \cellcolor{blueC!21}12/32 & \cellcolor{blueC!36}22/32 & \cellcolor{tealC!46}\textbf{+46.9}\,\textsubscript{\tiny(31.2--62.5)} & 15/0 \\
 & Astra 6 & \cellcolor{blueC!14}7/32 & \cellcolor{blueC!18}10/32 & \cellcolor{blueC!36}22/32 & \cellcolor{tealC!46}\textbf{+46.9}\,\textsubscript{\tiny(28.1--65.6)} & 16/1 \\
 & Opus 5.5 & \cellcolor{blueC!18}10/32 & \cellcolor{blueC!24}14/32 & \cellcolor{blueC!38}24/32 & \cellcolor{tealC!43}\textbf{+43.8}\,\textsubscript{\tiny(25.0--62.5)} & 15/1 \\
 & Fable 5.1 & \cellcolor{blueC!17}9/32 & \cellcolor{blueC!28}17/32 & \cellcolor{blueC!37}23/32 & \cellcolor{tealC!43}\textbf{+43.8}\,\textsubscript{\tiny(25.0--62.5)} & 15/1 \\
\addlinespace[2pt]
Qwen 0.8B & Sol 5.6 & \cellcolor{blueC!27}16/32 & \cellcolor{blueC!36}22/32 & \cellcolor{blueC!46}29/32 & \cellcolor{tealC!40}\textbf{+40.6}\,\textsubscript{\tiny(21.9--59.4)} & 14/1 \\
 & Astra 6 & \cellcolor{blueC!23}13/32 & \cellcolor{blueC!34}21/32 & \cellcolor{blueC!38}24/32 & \cellcolor{tealC!35}\textbf{+34.4}\,\textsubscript{\tiny(12.5--56.2)} & 14/3 \\
 & Opus 5.5 & \cellcolor{blueC!18}10/32 & \cellcolor{blueC!26}15/32 & \cellcolor{blueC!37}23/32 & \cellcolor{tealC!40}\textbf{+40.6}\,\textsubscript{\tiny(21.9--59.4)} & 14/1 \\
 & Fable 5.1 & \cellcolor{blueC!18}10/32 & \cellcolor{blueC!33}20/32 & \cellcolor{blueC!40}25/32 & \cellcolor{tealC!46}\textbf{+46.9}\,\textsubscript{\tiny(28.1--68.8)} & 17/2 \\
\bottomrule
\end{tabular*}
}
\par\smallskip{\footnotesize\raggedright The intervals are percentile intervals over the 32 blocks (2,000 resamples); every one excludes zero under each of 100 bootstrap seeds. Discordant counts are blocks accepted only under full-state / only under output-only feedback. Full-state minus coarse-state and coarse-state minus output-only contrasts: Table~\ref{tab:T05-contrasts}.\par}
\end{table}

In the feedback study, each of the four agents attempts 32 task instances on nano T2, nano T3, and Qwen3.5-0.8B under each of the three feedback settings, and the three attempts on one instance form a matched block (Figure~\ref{fig:F03_feedback}, Tables~\ref{tab:feedback_repeated}, \ref{tab:T05-contrasts}, and~\ref{tab:T05-diagnosis}). Full-state feedback raises acceptance over output-only feedback in all 12 agent--task cells, by 25.0 to 46.9 points, and every block-bootstrap interval excludes zero. Pooled, 295, 208, and 144 of 384 attempts are accepted under full-state, coarse-state, and output-only feedback. Total agent time is nearly equal across the settings, about 1,930 agent-hours each, so the cost per accepted design falls from 13.4 agent-hours under output-only feedback to 9.3 under coarse-state and 6.5 under full-state feedback. An earlier comparison with Sol 5.6 alone, on 16 instances each of T2 and T3, points the same way: full-state feedback adds two accepted attempts per task (Appendix~\ref{app:cohorts}).

\ifcompact The gain splits between reporting that state disagrees and locating where. Full-state feedback adds 151 pooled acceptances over output-only feedback; 64 arise between output-only and coarse-state feedback and 87 between coarse-state and full-state feedback. Both increments are positive in every cell. Full-state exceeds coarse-state feedback by 9.4 to 37.5 points, with intervals that exclude zero in seven of the 12 cells; for the gain of coarse-state over output-only feedback, the intervals exclude zero in only two cells (Table~\ref{tab:T05-contrasts}).

\fi
Full-state feedback exceeds coarse-state feedback in all 12 cells, by 9.4 to 37.5 points, and these intervals exclude zero in seven of 12 cells; the intervals for coarse-state against output-only feedback exclude zero in two. Intervals are percentile block bootstraps (2,000 resamples of whole blocks, fixed seed), computed from the paired attempt records. Across 100 bootstrap seeds, the full-minus-output intervals exclude zero in all 12 cells under every seed. The full-minus-coarse intervals exclude zero in six to eight cells and the coarse-minus-output intervals in two to four, because several of their bounds lie at zero.

Output-only feedback reports no state values, so whether an agent has found a state defect can be observed only under the other two settings. Among attempts with a state defect, agents found it within budget in 142 of 169 coarse-state attempts and 151 of 166 full-state attempts, and repaired it in 87 and 123 (Table~\ref{tab:T05-resources}). Measured from the start of the attempt, the median discovery took 1.50 agent-hours and 58 tool calls under coarse-state feedback and 1.35 hours and 57 calls under full-state feedback. The median completed repair took 3.22 hours (118 calls, 51.4k output tokens) and 2.69 hours (117 calls, 52.6k tokens).

\paragraph{Uncertainty of the cost per accepted design.}
Table~\ref{tab:T01_cost} adds intervals to the pooled costs. Each bootstrap draw resamples whole matched blocks, which keeps the four agents and three settings of an instance together; failed and capped attempts count toward agent time, and grading time does not. Pooled over agents and tasks, output-only feedback costs 13.38 [11.25, 16.19] agent-hours per accepted design, coarse-state 9.33 [8.16, 10.77], and full-state 6.55 [5.92, 7.31]; as ratios to output-only, 0.70 [0.60, 0.81] and 0.49 [0.42, 0.56]. Per agent and task (Table~\ref{tab:T02_cost_influence}), every full-state ratio has an interval below one, from 0.29 [0.12, 0.50] for Sol 5.6 on T3 to 0.65 [0.44, 0.90] for Opus 5.5 on T2, while eight of the 12 coarse-state intervals include one. To check whether a few attempts drive the ratios, we removed one matched block at a time. Over all 96 deletions the pooled full-state ratio stays between 0.48 and 0.50, and no deletion raises an agent--task full-state ratio above 0.71 (Figure~\ref{fig:F01_cost_robustness}). Removing single attempts, over all 1,152 deletions, keeps the pooled ratio between 0.486 and 0.492.

\begin{figure}[htbp]
\centering
\includegraphics[width=\linewidth,height=.62\textheight,keepaspectratio]{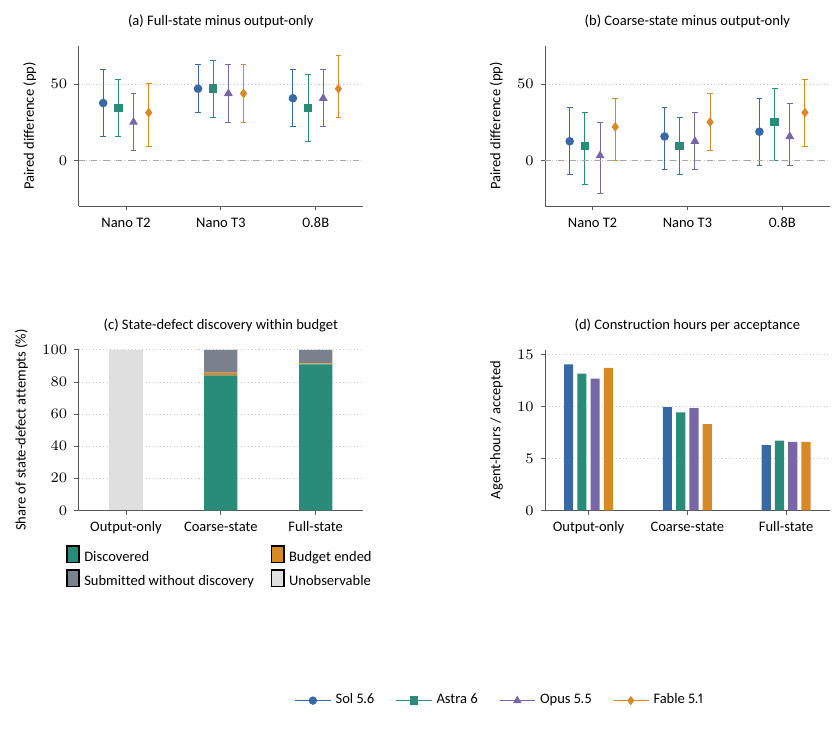}
\caption{Feedback study with 32 matched blocks per agent and task (T2, T3, 0.8B). (a) Full-state minus output-only and (b) coarse-state minus output-only final acceptance, with 2,000-resample block-bootstrap 95\% intervals. (c) Outcomes of attempts that contained a state defect, pooled over agents and tasks; output-only feedback reports no state values, so discovery there is unobservable. (d) Agent-hours per accepted design over the three tasks, including failed attempts.}
\label{fig:F03_feedback}
\end{figure}

\begin{table}[htbp]
\centering
\caption{\textbf{Paired feedback contrasts, coarse-state minus output-only and full-state minus coarse-state, with paired bootstrap 95\% intervals.}}
\label{tab:T05-contrasts}
{\small\tablefont\setlength{\tabcolsep}{7pt}\resultrows
\begin{tabular}{@{}llrrrr}
\toprule
Task & Agent & Disc. & Coarse $-$ output & Disc. & Full $-$ coarse \\
\midrule
T2 & Sol 5.6 & 9/5 & \cellcolor{tealC!16}+12.5\,\textsubscript{\tiny($-$9.4 to 34.4)} & 12/4 & \cellcolor{tealC!25}+25.0\,\textsubscript{\tiny(0.0--46.9)} \\
 & Astra 6 & 9/6 & \cellcolor{tealC!14}+9.4\,\textsubscript{\tiny($-$15.6 to 31.3)} & 8/0 & \cellcolor{tealC!25}+25.0\,\textsubscript{\tiny\textbf{(12.4--40.6)}} \\
 & Opus 5.5 & 8/7 & \cellcolor{tealC!10}+3.1\,\textsubscript{\tiny($-$21.9 to 25.0)} & 7/0 & \cellcolor{tealC!23}+21.9\,\textsubscript{\tiny\textbf{(9.4--37.5)}} \\
 & Fable 5.1 & 10/3 & \cellcolor{tealC!23}+21.9\,\textsubscript{\tiny(0.0--40.6)} & 8/5 & \cellcolor{tealC!14}+9.4\,\textsubscript{\tiny($-$12.5 to 31.2)} \\
\addlinespace[2pt]
T3 & Sol 5.6 & 9/4 & \cellcolor{tealC!19}+15.6\,\textsubscript{\tiny($-$6.2 to 34.4)} & 14/4 & \cellcolor{tealC!29}+31.2\,\textsubscript{\tiny\textbf{(6.2--53.1)}} \\
 & Astra 6 & 6/3 & \cellcolor{tealC!14}+9.4\,\textsubscript{\tiny($-$9.4 to 28.1)} & 16/4 & \cellcolor{tealC!34}+37.5\,\textsubscript{\tiny\textbf{(12.5--59.4)}} \\
 & Opus 5.5 & 8/4 & \cellcolor{tealC!16}+12.5\,\textsubscript{\tiny($-$6.2 to 31.2)} & 11/1 & \cellcolor{tealC!29}+31.2\,\textsubscript{\tiny\textbf{(12.5--50.0)}} \\
 & Fable 5.1 & 10/2 & \cellcolor{tealC!25}+25.0\,\textsubscript{\tiny\textbf{(6.2--43.8)}} & 10/4 & \cellcolor{tealC!21}+18.8\,\textsubscript{\tiny($-$6.2 to 40.6)} \\
\addlinespace[2pt]
0.8B adaptation & Sol 5.6 & 10/4 & \cellcolor{tealC!21}+18.8\,\textsubscript{\tiny($-$3.1 to 40.6)} & 9/2 & \cellcolor{tealC!23}+21.9\,\textsubscript{\tiny\textbf{(3.1--40.6)}} \\
 & Astra 6 & 12/4 & \cellcolor{tealC!25}+25.0\,\textsubscript{\tiny(0.0--46.9)} & 5/2 & \cellcolor{tealC!14}+9.4\,\textsubscript{\tiny($-$6.2 to 25.0)} \\
 & Opus 5.5 & 9/4 & \cellcolor{tealC!19}+15.6\,\textsubscript{\tiny($-$3.1 to 37.5)} & 12/4 & \cellcolor{tealC!25}+25.0\,\textsubscript{\tiny\textbf{(3.1--46.9)}} \\
 & Fable 5.1 & 13/3 & \cellcolor{tealC!29}+31.2\,\textsubscript{\tiny\textbf{(9.4--53.1)}} & 8/3 & \cellcolor{tealC!19}+15.6\,\textsubscript{\tiny($-$3.1 to 34.4)} \\
\bottomrule
\end{tabular}
}
\par\smallskip{\footnotesize\raggedright Effects are differences in final acceptance (percentage points) over 32 matched instance blocks; intervals resample whole blocks (2,000 bootstrap resamples) and are bold when they exclude zero. Disc.\ gives discordant blocks accepted only under the first-named setting / only under the second. Full-state minus output-only contrasts are in Table~\ref{tab:feedback_repeated}. Every setting uses the same grader and caps.\par}
\end{table}
\begin{table}[htbp]
\centering
\caption{\textbf{Qwen3.5-0.8B adaptation under three feedback settings.}}
\label{tab:T05-diagnosis}
{\small\tablefont\setlength{\tabcolsep}{4pt}\resultrows
\begin{tabular*}{\linewidth}{@{\extracolsep{\fill}}llrrrrrr@{}}
\toprule
Agent & Feedback & Visible (\%) & Public (\%) & Final (\%) & h/accepted & Found & Repaired \\
\midrule
Sol 5.6 & output-only & 18/32 (56.3) & 18/32 (56.3) & \cellcolor{blueC!27}16/32 (50.0) & 16.6 & 0 & 0 \\
Astra 6 & output-only & 21/32 (65.6) & 19/32 (59.4) & \cellcolor{blueC!23}13/32 (40.6) & 19.1 & 0 & 0 \\
Opus 5.5 & output-only & 19/32 (59.4) & 18/32 (56.3) & \cellcolor{blueC!18}10/32 (31.3) & 25.8 & 0 & 0 \\
Fable 5.1 & output-only & 19/32 (59.4) & 18/32 (56.3) & \cellcolor{blueC!18}10/32 (31.3) & 23.9 & 0 & 0 \\
Sol 5.6 & coarse-state & 25/32 (78.1) & 25/32 (78.1) & \cellcolor{blueC!36}22/32 (68.8) & 12.1 & 12 & 6 \\
Astra 6 & coarse-state & 22/32 (68.8) & 22/32 (68.8) & \cellcolor{blueC!34}21/32 (65.6) & 11.8 & 17 & 8 \\
Opus 5.5 & coarse-state & 18/32 (56.3) & 18/32 (56.3) & \cellcolor{blueC!26}15/32 (46.9) & 17.3 & 19 & 14 \\
Fable 5.1 & coarse-state & 25/32 (78.1) & 25/32 (78.1) & \cellcolor{blueC!33}20/32 (62.5) & 12.8 & 16 & 11 \\
Sol 5.6 & full-state & 29/32 (90.6) & 29/32 (90.6) & \cellcolor{blueC!46}29/32 (90.6) & 8.8 & 15 & 14 \\
Astra 6 & full-state & 24/32 (75.0) & 24/32 (75.0) & \cellcolor{blueC!38}24/32 (75.0) & 11.5 & 13 & 12 \\
Opus 5.5 & full-state & 25/32 (78.1) & 25/32 (78.1) & \cellcolor{blueC!37}23/32 (71.9) & 11.2 & 15 & 12 \\
Fable 5.1 & full-state & 27/32 (84.4) & 27/32 (84.4) & \cellcolor{blueC!40}25/32 (78.1) & 10.3 & 13 & 8 \\
\bottomrule
\end{tabular*}
}
\par\smallskip{\footnotesize\raggedright Found and Repaired count attempts in which the agent located or repaired a state defect within budget. Output-only feedback reports no state values, so its zero counts carry no information about state errors. Visible gives attempts that pass the checks reported to the agent, and Public those that pass the full public suite rerun with state checks.\par}
\end{table}
\begin{table}[htbp]
\centering
\caption{\textbf{Resources to discover and repair a state defect, by feedback setting.}}
\label{tab:T05-resources}
{\small\tablefont\setlength{\tabcolsep}{3pt}\resultrows
\begin{tabular*}{\linewidth}{@{\extracolsep{\fill}}llrrrrrrrr@{}}
\toprule
Feedback & Agent & Found (\%) & h & Calls & Tok.\ k & Repaired (\%) & h & Calls & Tok.\ k \\
\midrule
coarse-state & Sol 5.6 & \cellcolor{blueC!42}29/35 (82.9) & 1.29 & 52 & 20.0 & \cellcolor{blueC!24}15/35 (42.9) & 3.14 & 105 & 49.7 \\
 & Astra 6 & \cellcolor{blueC!43}39/46 (84.8) & 1.31 & 62 & 22.2 & \cellcolor{blueC!24}20/46 (43.5) & 3.03 & 122 & 47.8 \\
 & Opus 5.5 & \cellcolor{blueC!41}34/42 (81.0) & 1.75 & 64 & 32.0 & \cellcolor{blueC!31}25/42 (59.5) & 3.29 & 118 & 62.2 \\
 & Fable 5.1 & \cellcolor{blueC!44}40/46 (87.0) & 1.31 & 58 & 20.8 & \cellcolor{blueC!31}27/46 (58.7) & 3.51 & 112 & 55.7 \\
 & \textbf{All agents} & \cellcolor{blueC!43}\textbf{142/169 (84.0)} & \textbf{1.50} & \textbf{58} & \textbf{24.6} & \cellcolor{blueC!28}\textbf{87/169 (51.5)} & \textbf{3.22} & \textbf{118} & \textbf{51.4} \\
\midrule
full-state & Sol 5.6 & \cellcolor{blueC!43}36/43 (83.7) & 1.67 & 74 & 29.7 & \cellcolor{blueC!39}33/43 (76.7) & 2.88 & 131 & 60.4 \\
 & Astra 6 & \cellcolor{blueC!45}36/40 (90.0) & 1.26 & 46 & 20.1 & \cellcolor{blueC!40}31/40 (77.5) & 2.57 & 103 & 50.9 \\
 & Opus 5.5 & \cellcolor{blueC!47}40/43 (93.0) & 1.47 & 62 & 30.1 & \cellcolor{blueC!38}32/43 (74.4) & 2.67 & 126 & 58.3 \\
 & Fable 5.1 & \cellcolor{blueC!49}39/40 (97.5) & 1.33 & 59 & 26.4 & \cellcolor{blueC!35}27/40 (67.5) & 2.69 & 107 & 47.2 \\
 & \textbf{All agents} & \cellcolor{blueC!46}\textbf{151/166 (91.0)} & \textbf{1.35} & \textbf{57} & \textbf{26.5} & \cellcolor{blueC!38}\textbf{123/166 (74.1)} & \textbf{2.69} & \textbf{117} & \textbf{52.6} \\
\bottomrule
\end{tabular*}
}
\par\smallskip{\footnotesize\raggedright Rows cover attempts that contained a state defect, pooled over T2, T3, and Qwen3.5-0.8B. Found and Repaired count attempts in which the agent located or repaired the defect within budget; output-only feedback reports no state values, so it has no observable discovery. Resource columns are medians over those events, measured from the start of the attempt: elapsed agent-hours, tool calls, and generated output tokens (thousands).\par}
\end{table}

\begin{figure}[tbp]
\centering
\includegraphics[width=0.92\linewidth]{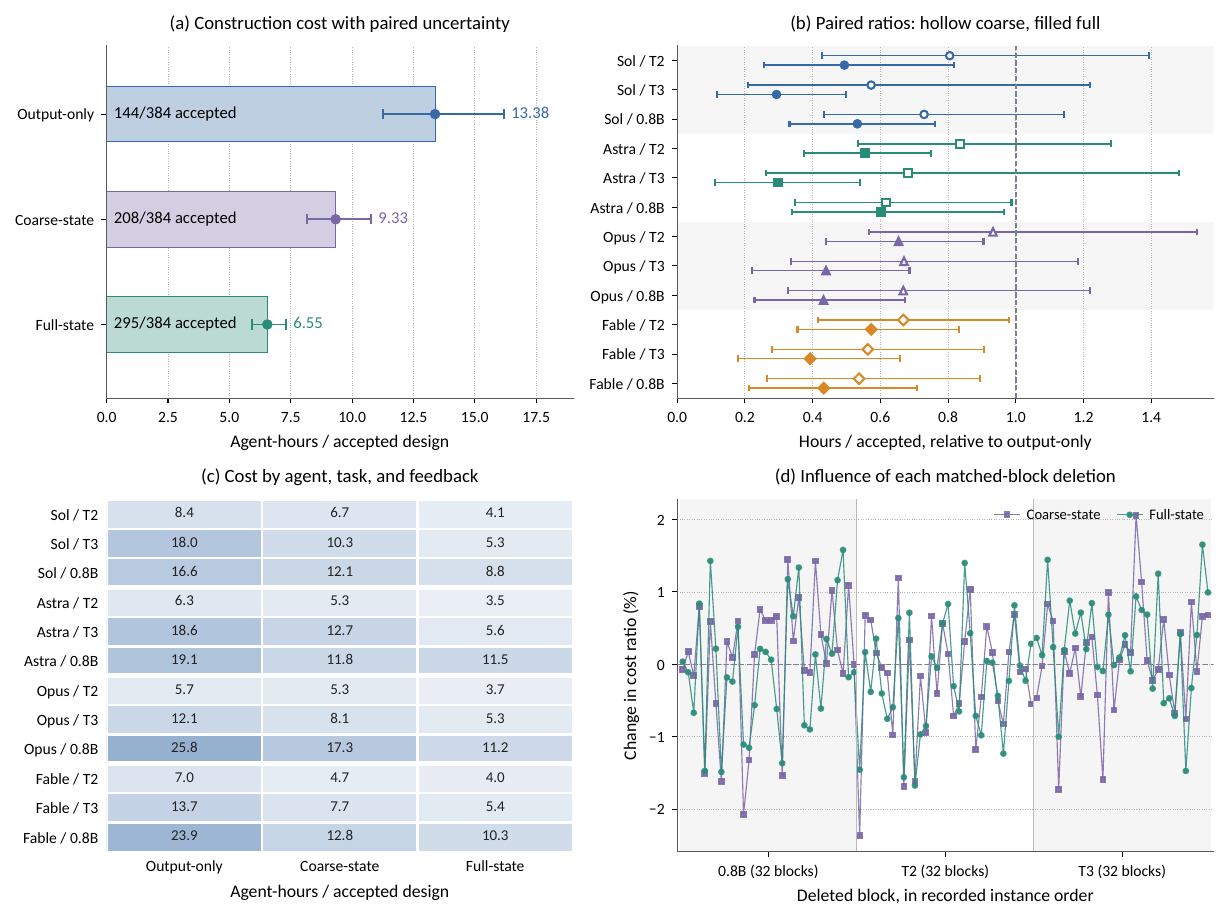}
\caption{Construction cost in the feedback study. Costs include all attempts. Paired intervals keep the matched blocks of an instance together across agents and settings. Hollow and filled markers compare coarse-state and full-state costs with output-only. The deletion panel shows the relative change in each pooled cost ratio after removing one matched block.}
\label{fig:F01_cost_robustness}
\end{figure}
\begin{table}[H]
\centering
\caption{\textbf{Construction cost with paired bootstrap 95\% intervals.}}
\label{tab:T01_cost}
{\footnotesize\tablefont\setlength{\tabcolsep}{3.5pt}\renewcommand{\arraystretch}{1.04}
\begin{tabular*}{\linewidth}{@{\extracolsep{\fill}}lrrrr@{}}
\toprule
\multicolumn{1}{c}{} & \multicolumn{1}{c}{Outcomes} & \multicolumn{2}{c}{Construction cost} & \multicolumn{1}{c}{Versus output-only} \\
\cmidrule(lr){2-2}\cmidrule(lr){3-4}\cmidrule(lr){5-5}
Feedback & Accepted & Total hours & Hours / accepted & Ratio \\
\midrule
Output-only & 144/384 & 1926.8 & 13.38\,\textsubscript{\tiny(11.25--16.19)} & Reference \\
Coarse-state & 208/384 & 1941.1 & 9.33\,\textsubscript{\tiny(8.16--10.77)} & \cellcolor{blueC!15}0.70\,\textsubscript{\tiny(0.60--0.81)} \\
Full-state & 295/384 & 1931.5 & 6.55\,\textsubscript{\tiny(5.92--7.31)} & \cellcolor{blueC!20}0.49\,\textsubscript{\tiny(0.42--0.56)} \\
\bottomrule
\end{tabular*}}
\par\smallskip{\footnotesize\raggedright All failed and capped attempts contribute to construction time; grader time is excluded. The intervals keep matched instances, agents, and settings together. Blue intensity encodes the reduction in cost ratio, not statistical significance.\par}
\end{table}

\begin{table}[H]
\centering
\caption{\textbf{Feedback cost ratios with paired bootstrap 95\% intervals, and influence by agent and task.}}
\label{tab:T02_cost_influence}
{\footnotesize\tablefont\setlength{\tabcolsep}{3.5pt}\renewcommand{\arraystretch}{1.04}
\begin{tabular*}{\linewidth}{@{\extracolsep{\fill}}llrrr@{}}
\toprule
\multicolumn{2}{c}{} & \multicolumn{2}{c}{Cost ratio} & \multicolumn{1}{c}{Full-state influence} \\
\cmidrule(lr){3-4}\cmidrule(lr){5-5}
Agent & Task & Coarse-state & Full-state & Block deletion range \\
\midrule
Fable 5.1 & T2 & 0.67\,\textsubscript{\tiny(0.42--0.98)} & 0.57\,\textsubscript{\tiny(0.36--0.83)} & 0.530--0.614 \\
Fable 5.1 & T3 & 0.56\,\textsubscript{\tiny(0.28--0.91)} & 0.39\,\textsubscript{\tiny(0.18--0.66)} & 0.347--0.418 \\
Fable 5.1 & 0.8B & 0.54\,\textsubscript{\tiny(0.26--0.90)} & 0.43\,\textsubscript{\tiny(0.21--0.71)} & 0.387--0.460 \\
\addlinespace[3pt]
Sol 5.6 & T2 & 0.80\,\textsubscript{\tiny(0.43--1.39)} & 0.49\,\textsubscript{\tiny(0.26--0.82)} & 0.443--0.528 \\
Sol 5.6 & T3 & 0.57\,\textsubscript{\tiny(0.21--1.22)} & 0.29\,\textsubscript{\tiny(0.12--0.50)} & 0.261--0.315 \\
Sol 5.6 & 0.8B & 0.73\,\textsubscript{\tiny(0.43--1.14)} & 0.53\,\textsubscript{\tiny(0.33--0.76)} & 0.490--0.562 \\
\addlinespace[3pt]
Astra 6 & T2 & 0.84\,\textsubscript{\tiny(0.53--1.28)} & 0.55\,\textsubscript{\tiny(0.38--0.75)} & 0.518--0.583 \\
Astra 6 & T3 & 0.68\,\textsubscript{\tiny(0.26--1.48)} & 0.30\,\textsubscript{\tiny(0.11--0.54)} & 0.253--0.318 \\
Astra 6 & 0.8B & 0.62\,\textsubscript{\tiny(0.35--0.99)} & 0.60\,\textsubscript{\tiny(0.34--0.96)} & 0.544--0.644 \\
\addlinespace[3pt]
Opus 5.5 & T2 & 0.93\,\textsubscript{\tiny(0.57--1.53)} & 0.65\,\textsubscript{\tiny(0.44--0.90)} & 0.608--0.701 \\
Opus 5.5 & T3 & 0.67\,\textsubscript{\tiny(0.34--1.18)} & 0.44\,\textsubscript{\tiny(0.22--0.69)} & 0.399--0.466 \\
Opus 5.5 & 0.8B & 0.67\,\textsubscript{\tiny(0.33--1.22)} & 0.43\,\textsubscript{\tiny(0.23--0.67)} & 0.386--0.461 \\
\bottomrule
\end{tabular*}}
\par\smallskip{\footnotesize\raggedright Both ratios use output-only cost as denominator. The final column removes one matched block at a time from the full-state/output-only comparison. Complete individual-deletion results accompany the table.\par}
\end{table}

\subsection{Controlled horizon, added state condition, and budget}
\label{app:add-state}
Nano tasks T1 to T3 differ in several properties at once. The controlled matrix instead starts every attempt from one Qwen3.5-nano scaffold, grades it with the same evaluator, and varies three factors: the number of decoding steps that must match (1, 4, 8, or 16), an added state condition (none, reset after activity, or bounded memory stalls), and a budget multiplier that scales all three caps (0.5, 1, or 1.5). The public and held-out suites grow with the horizon, from 984 and 68 cases at one step to 1,344 and 128 at 16. Each agent has 24 blocks per cell (Figure~\ref{fig:F04_state_budget}, Tables~\ref{tab:T06-horizon}--\ref{tab:T06-effects}).

At the 1$\times$ budget with no added condition, every agent has fewer attempts accepted at 16 steps than at one: Sol 5.6 falls from 21 to 8, Astra 6 from 23 to 15, Opus 5.5 from 21 to 10, and Fable 5.1 from 23 to 17, out of 24. Pooled over agents and added conditions, acceptance at the 1$\times$ budget falls from 252 to 115 of 288 between one and 16 steps (Table~\ref{tab:state_budget}). The starting implementation, evaluator, and caps are the same in every cell, so the drop follows from the longer required horizon. 

At the 0.5$\times$ budget, most attempts cannot finish within their caps: 192 of 288 one-step attempts and 286 of 288 sixteen-step attempts end at a cap, and 83 and one are accepted. Table~\ref{tab:state_budget} omits this budget for that reason, but the result shows that the 1$\times$ budget is not slack. At one step, only 8 of 288 attempts reach a cap at the 1$\times$ budget, and a 1.5$\times$ budget does not raise acceptance (247 against 252 of 288).

Bounded stalls lower pooled acceptance more than reset after activity at most horizons. Of the 144 per-agent intervals for a single added condition, 16 exclude zero (15 to 18 across 100 bootstrap seeds), against about seven expected under no effect. In 15 of the 16, acceptance is lower with the added condition, mostly stalls. 

\paragraph{Horizon by added condition.}
Table~\ref{tab:T11_interactions} and Figure~\ref{fig:F08_controlled_interactions} ask whether an added state condition changes how acceptance falls with the horizon. For each budget, the horizon effect is acceptance at 16 steps minus acceptance at one step, and the interaction is the horizon effect with an added condition minus the effect without one. At 0.5$\times$, both interactions are positive, 15.6 points [5.2, 26.0] for reset after activity and 27.1 [18.8, 35.4] for stalls: at 16 steps, at most 1 of 96 attempts per condition is accepted, so conditions that start lower at one step have less room to fall. At 1$\times$, the interaction is $-15.6$ [$-28.1$, $-2.1$] for reset and $-8.3$ [$-24.0$, 7.3] for stalls, and at 1.5$\times$ both intervals include zero. The intervals are pointwise and clustered on the 24 shared instances.

\ifcompact\fi

\begin{figure}[htbp]
\centering
\includegraphics[width=\linewidth,height=.62\textheight,keepaspectratio]{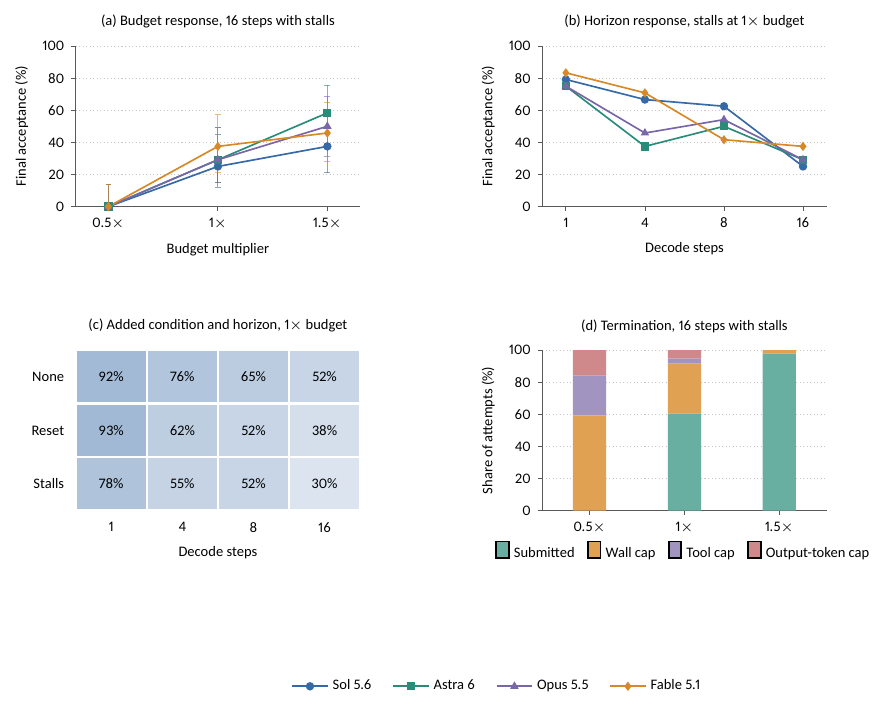}
\caption{Controlled matrix with one starting implementation: horizons of 1, 4, 8, and 16 decode steps; three added state conditions (none, reset after activity, bounded memory stalls); and budget multipliers of 0.5, 1, and 1.5 that scale all three caps together. Each cell has 24 blocks. (a) Acceptance by budget at 16 steps with stalls, with Wilson 95\% intervals. (b) Acceptance by horizon with stalls at the 1$\times$ budget. (c) Acceptance pooled over the four agents at the 1$\times$ budget. (d) Termination reasons at 16 steps with stalls. Each budget tier runs independent attempts, so the curves need not be monotone.}
\label{fig:F04_state_budget}
\end{figure}

\begin{table}[htbp]
\centering
\caption{\textbf{Acceptance by decode horizon at the 1$\times$ budget with no added state condition, with Wilson 95\% intervals.}}
\label{tab:T06-horizon}
{\small\tablefont\setlength{\tabcolsep}{7pt}\resultrows
\begin{tabular}{@{}rlrrrr}
\toprule
Horizon & Agent & Final (\%) & State diag. & Cap exits & h/accepted \\
\midrule
1 & Sol 5.6 & \cellcolor{blueC!44}21/24 (87.5)\,\textsubscript{\tiny(69.0--95.7)} & 0 & 1 & 3.5 \\
1 & Astra 6 & \cellcolor{blueC!48}23/24 (95.8)\,\textsubscript{\tiny(79.8--99.3)} & 1 & 0 & 3.0 \\
1 & Opus 5.5 & \cellcolor{blueC!44}21/24 (87.5)\,\textsubscript{\tiny(69.0--95.7)} & 0 & 1 & 3.1 \\
1 & Fable 5.1 & \cellcolor{blueC!48}23/24 (95.8)\,\textsubscript{\tiny(79.8--99.3)} & 0 & 1 & 2.9 \\
4 & Sol 5.6 & \cellcolor{blueC!37}17/24 (70.8)\,\textsubscript{\tiny(50.8--85.1)} & 0 & 2 & 5.4 \\
4 & Astra 6 & \cellcolor{blueC!37}17/24 (70.8)\,\textsubscript{\tiny(50.8--85.1)} & 5 & 1 & 5.1 \\
4 & Opus 5.5 & \cellcolor{blueC!42}20/24 (83.3)\,\textsubscript{\tiny(64.1--93.3)} & 0 & 1 & 4.5 \\
4 & Fable 5.1 & \cellcolor{blueC!40}19/24 (79.2)\,\textsubscript{\tiny(59.5--90.8)} & 1 & 1 & 4.9 \\
8 & Sol 5.6 & \cellcolor{blueC!33}15/24 (62.5)\,\textsubscript{\tiny(42.7--78.8)} & 1 & 4 & 6.8 \\
8 & Astra 6 & \cellcolor{blueC!33}15/24 (62.5)\,\textsubscript{\tiny(42.7--78.8)} & 0 & 2 & 6.4 \\
8 & Opus 5.5 & \cellcolor{blueC!33}15/24 (62.5)\,\textsubscript{\tiny(42.7--78.8)} & 2 & 5 & 6.5 \\
8 & Fable 5.1 & \cellcolor{blueC!37}17/24 (70.8)\,\textsubscript{\tiny(50.8--85.1)} & 1 & 3 & 5.6 \\
16 & Sol 5.6 & \cellcolor{blueC!19}8/24 (33.3)\,\textsubscript{\tiny(18.0--53.3)} & 1 & 5 & 13.3 \\
16 & Astra 6 & \cellcolor{blueC!33}15/24 (62.5)\,\textsubscript{\tiny(42.7--78.8)} & 3 & 3 & 6.9 \\
16 & Opus 5.5 & \cellcolor{blueC!23}10/24 (41.7)\,\textsubscript{\tiny(24.5--61.2)} & 2 & 6 & 10.1 \\
16 & Fable 5.1 & \cellcolor{blueC!37}17/24 (70.8)\,\textsubscript{\tiny(50.8--85.1)} & 0 & 5 & 6.3 \\
\bottomrule
\end{tabular}
}
\par\smallskip{\footnotesize\raggedright Every horizon starts from the same implementation with the same wall-time, tool-call, and output-token caps; longer horizons require more decoding work. Each cell has 24 task instances. State diag.\ counts rejections whose primary diagnosis is state/protocol, and Cap exits counts attempts that ended at a cap.\par}
\end{table}
\begin{table}[htbp]
\centering
\caption{\textbf{Contrasts between added state conditions at eight steps and the 1$\times$ budget, with paired bootstrap 95\% intervals.}}
\label{tab:T06-effects}
{\small\tablefont\setlength{\tabcolsep}{7pt}\resultrows
\begin{tabular}{@{}lllrr}
\toprule
Agent & A & B & Blocks & B $-$ A, pp \\
\midrule
Sol 5.6 & none & reset after activity & 24 & \cellcolor{orange!16}-12.5\,\textsubscript{\tiny($-$37.5 to 12.5)} \\
Astra 6 & none & reset after activity & 24 & +0.0\,\textsubscript{\tiny($-$25.0 to 25.0)} \\
Opus 5.5 & none & reset after activity & 24 & \cellcolor{orange!19}-16.7\,\textsubscript{\tiny($-$41.7 to 8.3)} \\
Fable 5.1 & none & reset after activity & 24 & \cellcolor{orange!22}-20.8\,\textsubscript{\tiny($-$45.8 to 4.2)} \\
Sol 5.6 & reset after activity & bounded stalls & 24 & \cellcolor{tealC!16}+12.5\,\textsubscript{\tiny($-$16.7 to 41.7)} \\
Sol 5.6 & none & bounded stalls & 24 & +0.0\,\textsubscript{\tiny($-$25.0 to 20.8)} \\
Astra 6 & reset after activity & bounded stalls & 24 & \cellcolor{orange!16}-12.5\,\textsubscript{\tiny($-$37.6 to 12.5)} \\
Astra 6 & none & bounded stalls & 24 & \cellcolor{orange!16}-12.5\,\textsubscript{\tiny($-$33.3 to 12.5)} \\
Opus 5.5 & reset after activity & bounded stalls & 24 & \cellcolor{tealC!14}+8.3\,\textsubscript{\tiny($-$20.8 to 33.3)} \\
Opus 5.5 & none & bounded stalls & 24 & \cellcolor{orange!14}-8.3\,\textsubscript{\tiny($-$37.5 to 20.8)} \\
Fable 5.1 & reset after activity & bounded stalls & 24 & \cellcolor{orange!14}-8.3\,\textsubscript{\tiny($-$37.5 to 16.7)} \\
Fable 5.1 & none & bounded stalls & 24 & \cellcolor{orange!28}-29.2\,\textsubscript{\tiny($-$54.2 to 0.0)} \\
\bottomrule
\end{tabular}
}
\par\smallskip{\footnotesize\raggedright Each contrast changes only the added state condition; horizon, starting implementation, feedback, and budget stay fixed. Bold intervals exclude zero.\par}
\end{table}

\begin{figure}[tbp]
\centering
\includegraphics[width=\linewidth]{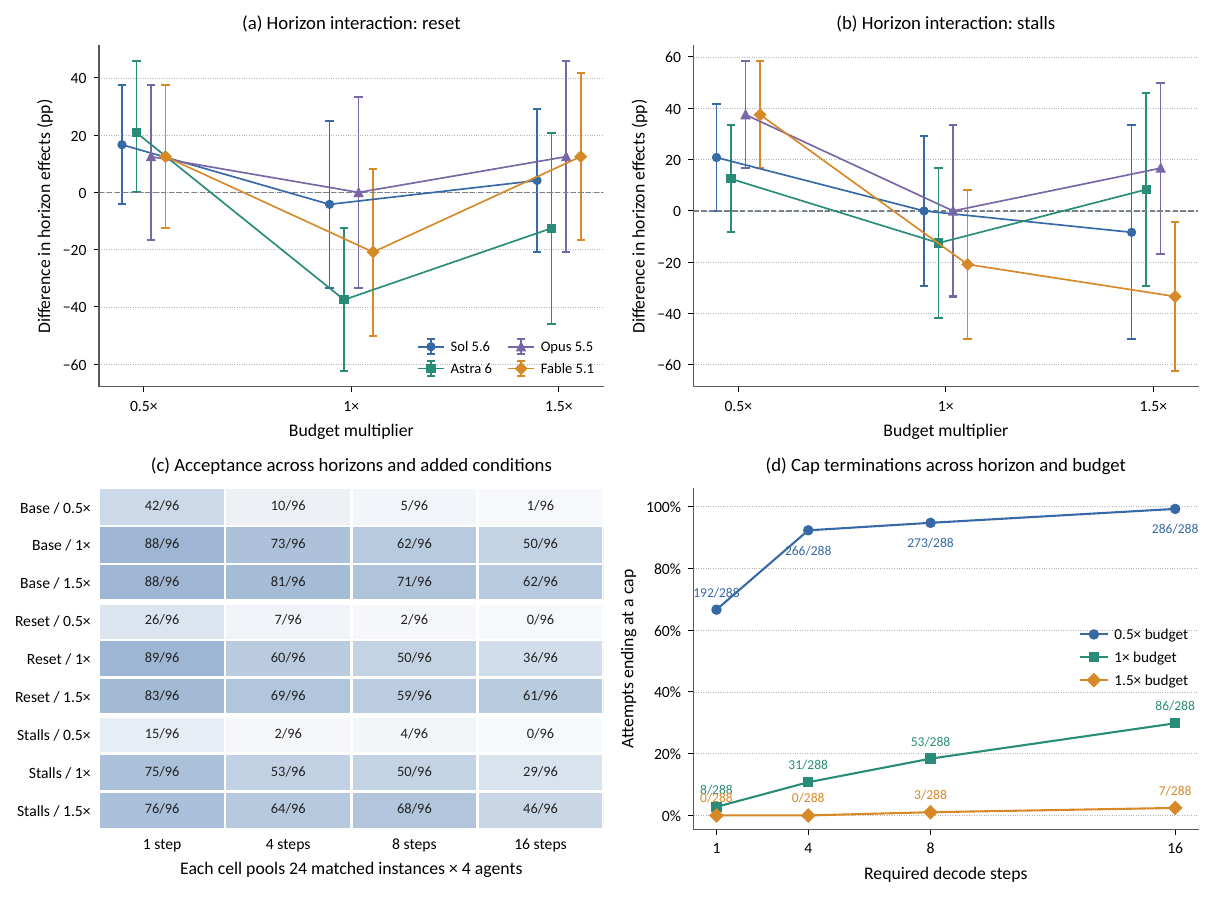}
\caption{Horizon-by-condition contrasts in the controlled matrix. Upper panels keep agent-specific paired intervals. Each acceptance cell pools 24 matched instances across four agents; cap totals also pool the three state conditions. A positive interaction is a smaller decline with the horizon and does not mean that an added condition makes a task easier.}
\label{fig:F08_controlled_interactions}
\end{figure}
\begin{table}[H]
\centering
\caption{\textbf{Horizon-by-state-condition contrasts in the controlled matrix, with paired bootstrap 95\% intervals.}}
\label{tab:T11_interactions}
{\small\tablefont\setlength{\tabcolsep}{3.5pt}\renewcommand{\arraystretch}{1.04}
\begin{tabular*}{\linewidth}{@{\extracolsep{\fill}}llrrr@{}}
\toprule
Budget & Condition & \shortstack{Base horizon\\effect (pp)} & \shortstack{Added-condition\\effect (pp)} & \shortstack{Interaction\\(pp)} \\
\midrule
0.5$\times$ & Reset & -42.7 & -27.1 & 15.6\,\textsubscript{\tiny(5.2--26.0)} \\
0.5$\times$ & Stalls & -42.7 & -15.6 & 27.1\,\textsubscript{\tiny(18.8--35.4)} \\
1.0$\times$ & Reset & -39.6 & -55.2 & -15.6\,\textsubscript{\tiny($-$28.1 to $-$2.1)} \\
1.0$\times$ & Stalls & -39.6 & -47.9 & -8.3\,\textsubscript{\tiny($-$24.0 to 7.3)} \\
1.5$\times$ & Reset & -27.1 & -22.9 & 4.2\,\textsubscript{\tiny($-$12.5 to 20.8)} \\
1.5$\times$ & Stalls & -27.1 & -31.2 & -4.2\,\textsubscript{\tiny($-$22.9 to 14.6)} \\
\bottomrule
\end{tabular*}}
\par\smallskip{\footnotesize\raggedright A horizon effect is acceptance at 16 steps minus acceptance at one step. The interaction subtracts the base effect from the added-condition effect. Positive values can reflect floor effects. The intervals are pointwise and clustered on 24 shared instances.\par}
\end{table}

\subsection{Long-horizon regrading}
\label{app:add-horizon}
Section~\ref{sec:evaluator_results} reports the regrading of 144 frozen submissions on Qwen3.5-0.8B, Gemma 4 E4B, and Llama 3.2 3B, 12 per agent and target, together with the three reference designs. Figure~\ref{fig:F06_horizons} and Tables~\ref{tab:T07-prefix} and~\ref{tab:T07-conditions} give the results at every horizon. Regrading involves no new attempts and changes no acceptance decision. The 69 accepted submissions are all exact at eight steps. At 32 steps, the horizon their tasks required, 85 of their 2,208 sequences fail, in 45 submissions. On new prefixes alone, 454 of their 552 sequences are still exact at 256 steps. Stalls remain the hardest condition at 256 steps, with 60.6 to 69.9\% of sequences exact per target against 75.6 to 86.1\% on new prefixes alone (Table~\ref{tab:T07-conditions}). Of the 624 sequences that fail at 256 steps, 482 first diverge in persistent state and 142 in output arithmetic. In 470 of the 482, the output also mismatches within 256 steps, a median of four decode positions after the first state mismatch (interquartile range two to eight, at most 40); in the other 12, no output mismatch appears within 256 steps. Counted by design instead of by sequence, 24 of the 69 accepted submissions are exact on all 32 sequences at 32 steps, seven at 64 steps, and none at 128 steps.

Table~\ref{tab:T19-state-output} separates the failing sequences by what is wrong. At the 32 steps the tasks required, 85 sequences of the accepted submissions fail: 51 in both stored state and output, 17 in state only, and 17 in output only. A test of that length that compared outputs alone would pass the 17 state-only sequences, nine of which are under bounded memory stalls and five under reset after activity. Longer runs do not add such sequences: at 256 steps, 12 of the 624 failing sequences are wrong in state only. Counted by design (Table~\ref{tab:T03_design_detection}), 45 accepted designs fail within 32 steps and 5 of them fail only in state; from 128 steps on, all 69 fail in output. Of the 482 sequences that first diverge in state, 470 show a wrong output a median of four steps later (quartiles 2 and 8, maximum 40), and 12 show none within 256 steps (Figure~\ref{fig:F02_state_detection}).

\paragraph{Longer output-only checks.}
Table~\ref{tab:T04_output_horizon} compares the 32-step full-state check with output-only checks of 32 to 64 steps on the same 2,208 sequences. The full-state check detects 85 sequences in 45 designs, and the 32-step output check 68 in 40. At 35 steps the output check detects 79 sequences; it still misses 10 that the full-state check finds and finds 4 that it misses, a difference of 0.27 points [$-0.05$, 0.63]. From 38 steps on, the output check detects more sequences than the 32-step full-state check, and at 48 steps it detects all 85, among 117 in 53 designs. On these designs, a full-state check finds the defects at a shorter horizon, and an output check that runs 50\% longer finds the same ones. 

\begin{figure}[htbp]
\centering
\includegraphics[width=\linewidth,height=.62\textheight,keepaspectratio]{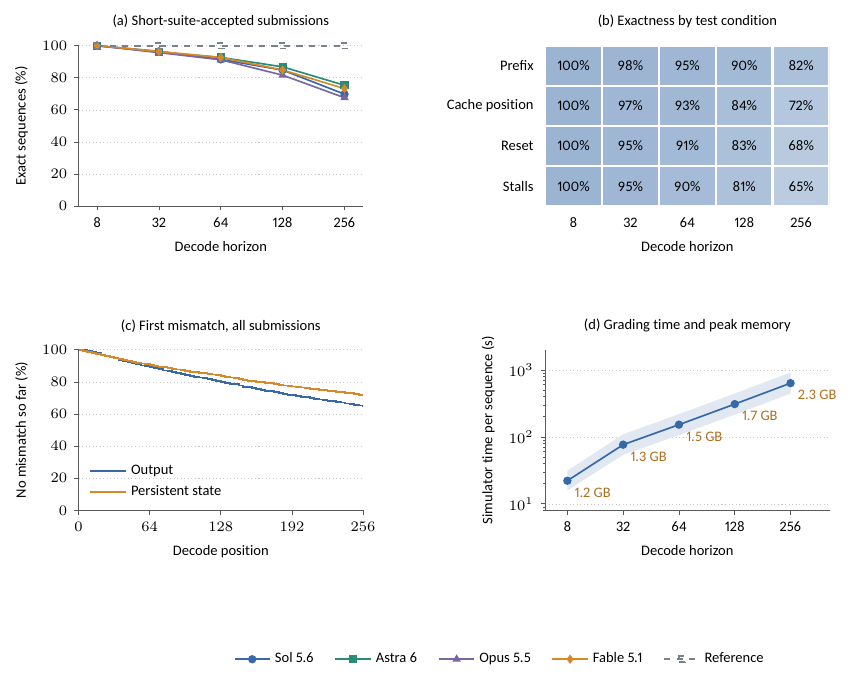}
\caption{Regrading of 144 submissions and 3 references for up to 256 steps, under four test conditions with eight prefixes per condition. (a, b) The 69 accepted submissions; references appear as a dashed line in (a). (c) All submissions: share with no output or state mismatch up to each position, over nested prefixes. (d) Median simulator time per sequence with a 10th--90th percentile band; labels give the median peak memory.}
\label{fig:F06_horizons}
\end{figure}

\begin{table}[htbp]
\centering
\caption{\textbf{Exact sequences of accepted submissions on new prefixes alone.}}
\label{tab:T07-prefix}
{\small\tablefont\setlength{\tabcolsep}{3pt}\resultrows
\begin{tabular*}{\linewidth}{@{\extracolsep{\fill}}llrrrrr@{}}
\toprule
Target & Agent & Designs & 32 steps & 64 steps & 128 steps & 256 steps \\
\midrule
Qwen 0.8B & Sol 5.6 & 7 & \cellcolor{blueC!50}56/56 (100.0) & \cellcolor{blueC!46}51/56 (91.1) & \cellcolor{blueC!46}51/56 (91.1) & \cellcolor{blueC!39}43/56 (76.8) \\
 & Astra 6 & 4 & \cellcolor{blueC!50}32/32 (100.0) & \cellcolor{blueC!50}32/32 (100.0) & \cellcolor{blueC!49}31/32 (96.9) & \cellcolor{blueC!49}31/32 (96.9) \\
 & Opus 5.5 & 4 & \cellcolor{blueC!49}31/32 (96.9) & \cellcolor{blueC!47}30/32 (93.8) & \cellcolor{blueC!44}28/32 (87.5) & \cellcolor{blueC!43}27/32 (84.4) \\
 & Fable 5.1 & 7 & \cellcolor{blueC!48}54/56 (96.4) & \cellcolor{blueC!47}52/56 (92.9) & \cellcolor{blueC!44}49/56 (87.5) & \cellcolor{blueC!43}47/56 (83.9) \\
\addlinespace[2pt]
Gemma 4 E4B & Sol 5.6 & 3 & \cellcolor{blueC!50}24/24 (100.0) & \cellcolor{blueC!48}23/24 (95.8) & \cellcolor{blueC!44}21/24 (87.5) & \cellcolor{blueC!38}18/24 (75.0) \\
 & Astra 6 & 6 & \cellcolor{blueC!50}48/48 (100.0) & \cellcolor{blueC!49}47/48 (97.9) & \cellcolor{blueC!49}47/48 (97.9) & \cellcolor{blueC!44}42/48 (87.5) \\
 & Opus 5.5 & 5 & \cellcolor{blueC!48}38/40 (95.0) & \cellcolor{blueC!45}36/40 (90.0) & \cellcolor{blueC!38}30/40 (75.0) & \cellcolor{blueC!33}25/40 (62.5) \\
 & Fable 5.1 & 7 & \cellcolor{blueC!48}54/56 (96.4) & \cellcolor{blueC!46}51/56 (91.1) & \cellcolor{blueC!44}49/56 (87.5) & \cellcolor{blueC!38}42/56 (75.0) \\
\addlinespace[2pt]
Llama 3B & Sol 5.6 & 5 & \cellcolor{blueC!50}40/40 (100.0) & \cellcolor{blueC!49}39/40 (97.5) & \cellcolor{blueC!48}38/40 (95.0) & \cellcolor{blueC!47}37/40 (92.5) \\
 & Astra 6 & 8 & \cellcolor{blueC!49}63/64 (98.4) & \cellcolor{blueC!48}61/64 (95.3) & \cellcolor{blueC!46}59/64 (92.2) & \cellcolor{blueC!45}57/64 (89.1) \\
 & Opus 5.5 & 7 & \cellcolor{blueC!49}55/56 (98.2) & \cellcolor{blueC!48}54/56 (96.4) & \cellcolor{blueC!45}50/56 (89.3) & \cellcolor{blueC!41}45/56 (80.4) \\
 & Fable 5.1 & 6 & \cellcolor{blueC!50}48/48 (100.0) & \cellcolor{blueC!49}47/48 (97.9) & \cellcolor{blueC!48}46/48 (95.8) & \cellcolor{blueC!42}40/48 (83.3) \\
\bottomrule
\end{tabular*}
}
\par\smallskip{\footnotesize\raggedright Each submission is graded on eight new prefixes; cells count sequences whose outputs and persistent state match the integer oracle through the horizon. The same submissions appear at every horizon; their tasks required 32 steps.\par}
\end{table}
\begin{table}[htbp]
\centering
\caption{\textbf{Exact sequences at 256 steps by test condition.}}
\label{tab:T07-conditions}
{\small\tablefont\setlength{\tabcolsep}{7pt}\resultrows
\begin{tabular}{@{}lrrrr}
\toprule
Target & New prefix alone & Nonzero cache position & Reset after activity & Bounded stalls \\
\midrule
Qwen 0.8B & \cellcolor{blueC!43}148/176 (84.1) & \cellcolor{blueC!37}127/176 (72.2) & \cellcolor{blueC!37}126/176 (71.6) & \cellcolor{blueC!36}123/176 (69.9) \\
Gemma 4 E4B & \cellcolor{blueC!39}127/168 (75.6) & \cellcolor{blueC!38}124/168 (73.8) & \cellcolor{blueC!35}113/168 (67.3) & \cellcolor{blueC!34}108/168 (64.3) \\
Llama 3B & \cellcolor{blueC!44}179/208 (86.1) & \cellcolor{blueC!36}146/208 (70.2) & \cellcolor{blueC!34}137/208 (65.9) & \cellcolor{blueC!32}126/208 (60.6) \\
\bottomrule
\end{tabular}
}
\par\smallskip{\footnotesize\raggedright Rows pool the accepted submissions of the four agents. Regrading does not change any acceptance decision.\par}
\end{table}

\begin{table}[htbp]
\centering
\caption{\textbf{State and output exactness per regraded sequence.} Each sequence is exact or wrong in stored state and in output over the given number of steps.}
\label{tab:T19-state-output}
{\small\tablefont\setlength{\tabcolsep}{5pt}\resultrows
\begin{tabular}{@{}lrrrrrr@{}}
\toprule
Submissions & Steps & Sequences & Both exact & State only & Output only & Both wrong \\
\midrule
Accepted (69) & 32 & 2,208 & 2,123 & 17 & 17 & 51 \\
 & 64 & 2,208 & 2,033 & 13 & 41 & 121 \\
 & 128 & 2,208 & 1,868 & 12 & 81 & 247 \\
 & 256 & 2,208 & 1,584 & 12 & 142 & 470 \\
\addlinespace[2pt]
All (144) & 32 & 4,608 & 4,317 & 44 & 53 & 194 \\
 & 64 & 4,608 & 4,069 & 36 & 108 & 395 \\
 & 128 & 4,608 & 3,649 & 39 & 203 & 717 \\
 & 256 & 4,608 & 2,969 & 28 & 342 & 1,269 \\
\bottomrule
\end{tabular}
}
\par\smallskip{\footnotesize\raggedright Accepted and All: the accepted submissions and all regraded submissions. State only: the stored state mismatches the contract within the given steps and every output matches, so a test of that length that compares outputs alone passes the sequence. Output only: an output mismatches and the stored state stays exact. The number of submissions is in parentheses; each has 32 sequences.\par}
\end{table}

\begin{figure}[tbp]
\centering
\includegraphics[width=\linewidth]{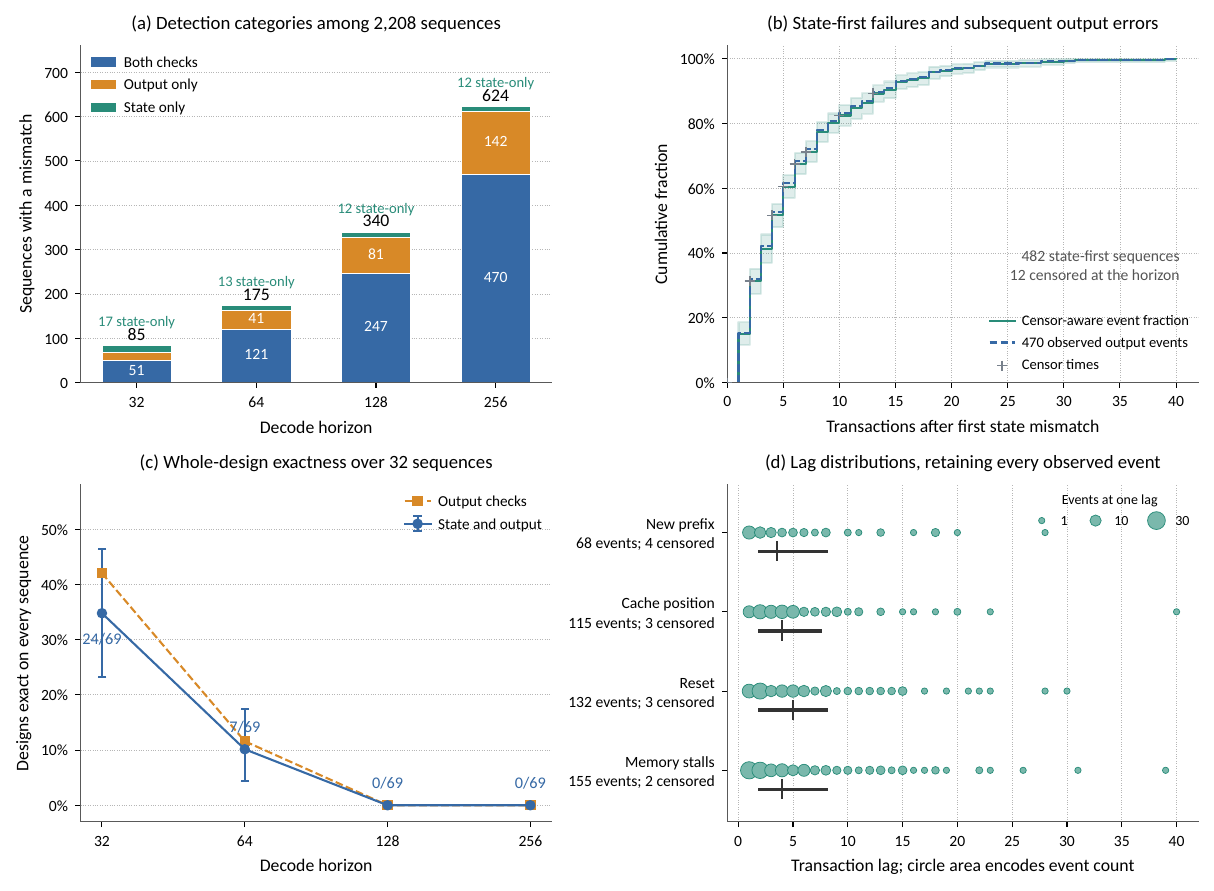}
\caption{State and output detection in the regraded accepted designs. Lag curves show the conditional ECDF of the 470 observed output events and one minus the Kaplan--Meier survival estimate, which keeps the 12 censored sequences; crosses mark censoring times. In the lag-frequency panel, circle area encodes event counts, horizontal bars show interquartile ranges, and vertical ticks show medians. Design and sequence denominators are separate.}
\label{fig:F02_state_detection}
\end{figure}
\begin{table}[H]
\centering
\caption{\textbf{Sequence-level and whole-design detection in the accepted subset.}}
\label{tab:T03_design_detection}
{\small\tablefont\setlength{\tabcolsep}{3.5pt}\renewcommand{\arraystretch}{1.04}
\begin{tabular*}{\linewidth}{@{\extracolsep{\fill}}rrrrrrr@{}}
\toprule
\multicolumn{1}{c}{} & \multicolumn{3}{c}{Sequences with a mismatch} & \multicolumn{3}{c}{Designs with a mismatch} \\
\cmidrule(lr){2-4}\cmidrule(lr){5-7}
Steps & \shortstack{Any-check\\sequences} & \shortstack{Output\\sequences} & \shortstack{State-only\\sequences} & \shortstack{Any-check\\designs} & \shortstack{Output\\designs} & \shortstack{Additional\\designs} \\
\midrule
32 & 85 & 68 & 17 & 45 & 40 & 5 \\
64 & 175 & 162 & 13 & 62 & 61 & 1 \\
128 & 340 & 328 & 12 & 69 & 69 & 0 \\
256 & 624 & 612 & 12 & 69 & 69 & 0 \\
\bottomrule
\end{tabular*}}
\par\smallskip{\footnotesize\raggedright There are 2,208 sequences from 69 selected designs at every horizon. An additional design has no output failure across its 32 sequences but has a state failure. Selection of the original replay pool is undocumented; these are conditional counts.\par}
\end{table}

\begin{table}[H]
\centering
\caption{\textbf{Full-state checks at 32 steps against longer output-only checks, with paired bootstrap 95\% intervals.}}
\label{tab:T04_output_horizon}
{\small\tablefont\setlength{\tabcolsep}{3.5pt}\renewcommand{\arraystretch}{1.04}
\begin{tabular*}{\linewidth}{@{\extracolsep{\fill}}rrrrrr@{}}
\toprule
\multicolumn{1}{c}{} & \multicolumn{2}{c}{Output-only check} & \multicolumn{2}{c}{Detected by one check only} & \multicolumn{1}{c}{Paired contrast} \\
\cmidrule(lr){2-3}\cmidrule(lr){4-5}\cmidrule(lr){6-6}
\shortstack{Output-only\\steps} & \shortstack{Detected\\sequences} & \shortstack{Detected\\designs} & \shortstack{Full-state\\32 steps} & \shortstack{Output-only\\longer} & \shortstack{Difference\\(pp)} \\
\midrule
32 & 68 & 40/69 & 17 & 0 & 0.77\,\textsubscript{\tiny(0.45--1.13)} \\
35 & 79 & 42/69 & 10 & 4 & 0.27\,\textsubscript{\tiny($-$0.05 to 0.63)} \\
38 & 89 & 48/69 & 6 & 10 & -0.18\,\textsubscript{\tiny($-$0.54 to 0.18)} \\
40 & 95 & 49/69 & 5 & 15 & -0.45\,\textsubscript{\tiny($-$0.82 to $-$0.09)} \\
48 & 117 & 53/69 & 0 & 32 & -1.45\,\textsubscript{\tiny($-$1.95 to $-$1.00)} \\
64 & 162 & 61/69 & 0 & 77 & -3.49\,\textsubscript{\tiny($-$4.35 to $-$2.67)} \\
\bottomrule
\end{tabular*}}
\par\smallskip{\footnotesize\raggedright The full-state check at 32 steps detects 85 of the 2,208 sequences of the 69 accepted designs, in 45 designs. Each row extends the output-only check on the same frozen sequences; counts come from the recorded first-failure steps. Difference is full-state minus output-only detection; its interval resamples designs within target.\par}
\end{table}

\subsection{Submission timing and failure outcomes}
\label{app:add-outcomes}
Figure~\ref{fig:F15_time_outcomes} and Table~\ref{tab:T15-outcomes} classify every attempt in the feedback study, on Gemma 4 E4B and Llama 3.2 3B, and in the controlled matrix at the 1$\times$ budget by how it ended: accepted, submitted and rejected (with the grader's primary diagnosis), or stopped at the wall-time, tool-call, or output-token cap. From output-only to full-state feedback, state/protocol rejections fall from 81 to 17 of 384 attempts and cap exits from 58 to 43. Under full-state feedback, caps end 43 of the 89 failed attempts: 26 had no state defect, 13 had already repaired one, and four reached the cap before the repair was complete. Caps also end 70 of the 136 failed attempts at construction from interfaces and 86 of the 173 failed 16-step attempts in the controlled matrix. Figure~\ref{fig:failure_records} gives excerpts from three attempt records.

Under every feedback setting, accepted designs were submitted at a median of 56 to 59\% of the wall cap, but the settings separate early. By half the cap, 52, 67, and 82 of 384 attempts under output-only, coarse-state, and full-state feedback had submitted a design that was later accepted, and by three quarters of the cap, 122, 164, and 232 had. Only 6, 10, and 20 accepted designs were submitted after 90\% of the cap. On Gemma 4 E4B and Llama 3.2 3B, accepted designs arrived at medians of 67\% (scaffold adaptation) and 66\% (construction from interfaces) of the cap, and 85 and 36 of 192 attempts had submitted one by three quarters of it. The controlled matrix varies the caps directly (Appendix~\ref{app:add-state}).

Table~\ref{tab:T15-visible} follows each submission in the feedback study from the checks the agent could see to final acceptance. Under output-only feedback, 219 submissions pass every visible check, 199 pass the full public suite with its state checks, and 144 are accepted. Under coarse-state and full-state feedback, every submission that passes the visible checks also passes the full public suite (252 and 314), and 208 and 295 are accepted.

\begin{figure}[htbp]
\centering
\includegraphics[width=\linewidth,height=.62\textheight,keepaspectratio]{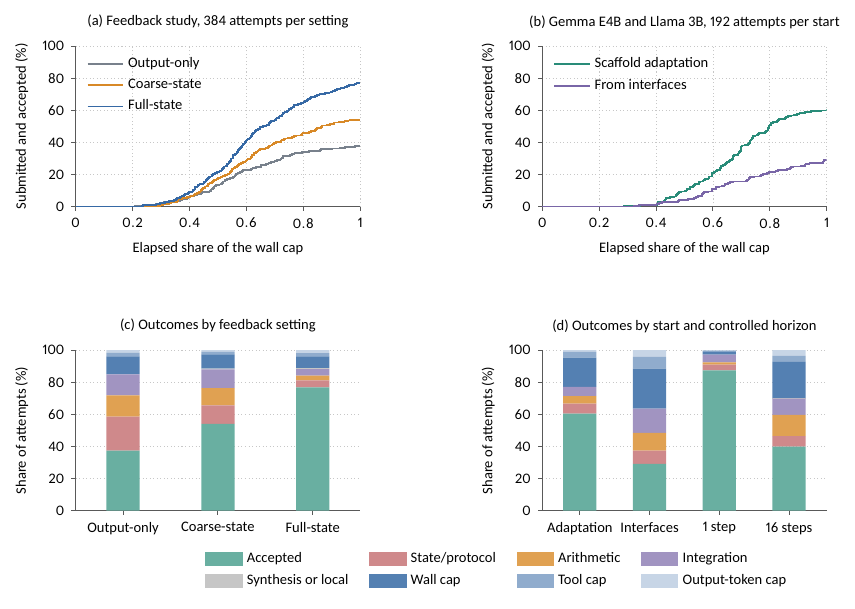}
\caption{Submission timing and outcomes of construction attempts. (a, b) Share of attempts submitted by each elapsed fraction of the wall cap and later accepted: (a) by feedback setting, pooled over four agents and three tasks; (b) by start on Gemma 4 E4B and Llama 3.2 3B, pooled over four agents. The controlled matrix varies the caps directly (Figure~\ref{fig:F04_state_budget}). (c, d) Outcome of every attempt: accepted, submitted and rejected with the grader\textquotesingle s primary diagnosis, or ended at the wall, tool, or output-token cap; (d) adds the controlled matrix at one and 16 steps with the 1$\times$ budget, pooled over its three added state conditions.}
\label{fig:F15_time_outcomes}
\end{figure}

\begin{table}[htbp]
\centering
\caption{\textbf{Outcome of every construction attempt.}}
\label{tab:T15-outcomes}
{\small\tablefont\setlength{\tabcolsep}{2pt}\resultrows
\begin{tabular*}{\linewidth}{@{\extracolsep{\fill}}llrrrrrrrrr@{}}
\toprule
Study & Setting & Accepted (\%) & State & Arith. & Integr. & Synth./loc. & Wall & Tools & Output & Caps/failures \\
\midrule
Feedback & Output-only & 144/384 (37.5) & 81 & 51 & 50 & 0 & 43 & 9 & 6 & 58/240 \\
 & Coarse-state & 208/384 (54.2) & 44 & 41 & 45 & 2 & 34 & 6 & 4 & 44/176 \\
 & Full-state & 295/384 (76.8) & 17 & 11 & 16 & 2 & 28 & 8 & 7 & 43/89 \\
\addlinespace[2pt]
Gemma, Llama & Adaptation & 116/192 (60.4) & 12 & 9 & 11 & 0 & 35 & 7 & 2 & 44/76 \\
 & Interfaces & 56/192 (29.2) & 16 & 21 & 29 & 0 & 48 & 14 & 8 & 70/136 \\
\addlinespace[2pt]
Controlled & 1 step & 252/288 (87.5) & 10 & 4 & 14 & 0 & 5 & 2 & 1 & 8/36 \\
 & 4 steps & 186/288 (64.6) & 20 & 22 & 29 & 0 & 27 & 3 & 1 & 31/102 \\
 & 8 steps & 162/288 (56.3) & 26 & 26 & 21 & 0 & 41 & 8 & 4 & 53/126 \\
 & 16 steps & 115/288 (39.9) & 19 & 38 & 29 & 1 & 65 & 11 & 10 & 86/173 \\
\bottomrule
\end{tabular*}
}
\par\smallskip{\footnotesize\raggedright Rejected attempts are split into those submitted and rejected, by the grader's primary diagnosis (state/protocol, arithmetic, integration, synthesis or local checks), and those that ended at the wall, tool, or output-token cap; no attempt that ended at a cap was accepted. The feedback rows pool four agents and three tasks; the Gemma, Llama rows pool four agents on Gemma 4 E4B and Llama 3.2 3B; the controlled rows pool four agents and three added state conditions at the 1$\times$ budget.\par}
\end{table}
\begin{table}[htbp]
\centering
\caption{\textbf{PPA-track attempts that end without a correct design.}}
\label{tab:T15-optimization}
{\small\tablefont\setlength{\tabcolsep}{2.5pt}\resultrows
\begin{tabular*}{\linewidth}{@{\extracolsep{\fill}}lrrrrrrr@{}}
\toprule
Agent & $n$ & Kept correct (\%) & Submitted incorrect & Wall cap & Tool cap & Token cap & No correct design (\%) \\
\midrule
Sol 5.6 & 384 & 260/384 (67.7) & 38 & 51 & 17 & 18 & 124/384 (32.3) \\
Astra 6 & 384 & 199/384 (51.8) & 45 & 82 & 35 & 23 & 185/384 (48.2) \\
Opus 5.5 & 384 & 212/384 (55.2) & 59 & 61 & 23 & 29 & 172/384 (44.8) \\
Fable 5.1 & 384 & 294/384 (76.6) & 52 & 13 & 13 & 12 & 90/384 (23.4) \\
\midrule
\textbf{All} & 1,536 & 965/1536 (62.8) & 194 & 207 & 88 & 82 & 571/1536 (37.2) \\
\bottomrule
\end{tabular*}
}
\par\smallskip{\footnotesize\raggedright Every attempt starts from the correct reference design of its target (384 per agent, 96 on each of the four targets). The RTL present when an attempt reaches a cap is graded; every attempt that reached a cap failed the correctness checks.\par}
\end{table}
\begin{table}[htbp]
\centering
\caption{\textbf{Public passing as the agent saw it and as the grader rechecked it.}}
\label{tab:T15-visible}
{\small\tablefont\setlength{\tabcolsep}{7pt}\resultrows
\begin{tabular}{@{}lrrrrr}
\toprule
Feedback & $n$ & Visible checks pass & Full public pass & Accepted & Visible pass, rejected \\
\midrule
Output-only & 384 & 219 & 199 & 144 & 75 \\
Coarse-state & 384 & 252 & 252 & 208 & 44 \\
Full-state & 384 & 314 & 314 & 295 & 19 \\
\bottomrule
\end{tabular}
}
\par\smallskip{\footnotesize\raggedright Visible checks are those the feedback setting reports to the agent; after submission the grader reruns the full public suite with state checks. Each column is a subset of the one before it. Rows pool four agents and three tasks.\par}
\end{table}

\subsection{Agent comparisons on matched instances}
\label{app:add-agents}
In every experiment the four agents attempt the same instances, so they are compared within an instance. Table~\ref{tab:T17-separation} tests, for each experiment, whether the agents are interchangeable within an instance, and Table~\ref{tab:T17-pairs} gives the pairwise differences on the correctness track.

Three experiments rank the agents, and their rankings reproduce across random halves of the instances (Table~\ref{tab:T19-split-half}). On the correctness track, Sol 5.6 has 289 of 472 attempts accepted, Opus 5.5 264, Astra 6 235, and Fable 5.1 228 ($p<0.001$); the Kendall $\tau$ between the orders of two random halves averages 0.78 ($p=0.001$), and the halves give the same order of all four agents in 64 of 200 splits. On the controlled matrix, Fable 5.1 has 445 of 864 attempts accepted, Astra 6 422, Opus 5.5 405, and Sol 5.6 385 ($p=0.001$); the split-half $\tau$ is 0.80 ($p=0.001$), and the halves give the same order in 102 of 200 splits. On the PPA track, Fable 5.1 improves 184 of 384 designs and Sol 5.6 151, against 123 and 111 for Astra 6 and Opus 5.5 ($p<0.001$); the split-half $\tau$ is 0.87 ($p=0.001$), and the halves give the same order in 116 of 200 splits. Each ranking describes its own experiment: Sol 5.6 is first on the correctness track, second on the PPA track, and last on the controlled matrix. The overall score of Appendix~\ref{app:add-score} combines the two tracks of the benchmark and leaves the diagnostic studies out.

On the controlled matrix, Sol 5.6 falls behind through designs it submits and the grader rejects. At 16 steps, each agent ends 93 to 96 of its 216 attempts at a cap. Sol 5.6 has 63 of its 121 submitted designs rejected, against 54 of 121 for Opus 5.5, 48 of 120 for Astra 6, and 35 of 123 for Fable 5.1. Of Sol 5.6's 63, 50 were submitted while failing public tests the agent had seen (24 of 35 for Fable 5.1), and the grader's primary diagnosis is arithmetic for 28, integration for 22, and state or protocol for 13. Its rejected submissions grow from 24 at one step to 63 at 16, against 18 and 35 for Fable 5.1, and at 16 steps it has 21 and 37 attempts accepted at the 1$\times$ and 1.5$\times$ budgets, against 37 and 50 for Fable 5.1. On the six scaffold-adaptation tasks of the correctness track, Sol 5.6 submits 196 of 232 attempts, more than any other agent, and 18 of those submissions fail the public tests it had seen, the fewest of the four.

On the correctness track, three pairwise differences are distinguishable from zero after a Holm adjustment over the six pairs (Table~\ref{tab:T17-pairs}): Sol 5.6 over Astra 6, 11.4 points with a 95\% interval of 6.4 to 16.5; Sol 5.6 over Fable 5.1, 12.9 points (8.1 to 18.0); and Opus 5.5 over Fable 5.1, 7.6 points (2.5 to 12.7). All four agents are accepted on 99 of the 472 instances and none on 74; on the other 299, the agents disagree. Any two agents disagree on about 35\% of the instances, so at this size the track detects a paired difference of about 8 points with 80\% power. Among the twelve tasks, three separate the agents after a Holm adjustment: Gemma 3 4B from interfaces ($p<0.001$), where Astra 6 and Fable 5.1 are accepted on 9 and 14 of 64 instances against 28 and 31 for Opus 5.5 and Sol 5.6; scaffold adaptation of Llama 3.2-1B ($p=0.007$), where Sol 5.6 is accepted on 54 of 64 instances and the others on 35 to 41; and scaffold adaptation of Qwen3.5-4B ($p=0.015$), where Opus 5.5 is accepted on 12 of 24 instances and the others on 20 or 21.

\begin{table}[htbp]
\centering
\caption{\textbf{Agents compared on matched instances.} Successes per agent out of the given number of instances, and the permutation $p$-value for the hypothesis that the four agents are interchangeable within an instance.}
\label{tab:T17-separation}
{\small\tablefont\setlength{\tabcolsep}{4.5pt}\resultrows
\begin{tabular}{@{}llrrrrrr@{}}
\toprule
Experiment & Instances & $n$ & Sol & Astra & Opus & Fable & $p$ \\
\midrule
Correctness track & All twelve tasks & 472 & 289 & 235 & 264 & 228 & $<$0.001 \\
 & Nano T2, from interfaces & 32 & 24 & 28 & 26 & 25 & 0.562 (1.000) \\
 & Nano T3, from interfaces & 32 & 22 & 22 & 24 & 23 & 0.965 (1.000) \\
 & Gemma 4 E4B, from interfaces & 24 & 9 & 3 & 7 & 6 & 0.162 (1.000) \\
 & Llama 3.2 3B, from interfaces & 24 & 7 & 5 & 9 & 10 & 0.457 (1.000) \\
 & Gemma 3 4B, from interfaces & 64 & 31 & 9 & 28 & 14 & $<$0.001 ($<$0.001) \\
 & Llama 3.2-1B, from interfaces & 64 & 25 & 29 & 27 & 20 & 0.312 (1.000) \\
 & Qwen3.5-0.8B, scaffold & 32 & 29 & 24 & 23 & 25 & 0.176 (1.000) \\
 & Qwen3.5-4B, scaffold & 24 & 20 & 21 & 12 & 21 & 0.002 (0.015) \\
 & Gemma 4 E4B, scaffold & 24 & 16 & 12 & 17 & 9 & 0.066 (0.598) \\
 & Llama 3.2 3B, scaffold & 24 & 17 & 15 & 16 & 14 & 0.842 (1.000) \\
 & Gemma 3 4B, scaffold & 64 & 35 & 26 & 35 & 26 & 0.104 (0.829) \\
 & Llama 3.2-1B, scaffold & 64 & 54 & 41 & 40 & 35 & $<$0.001 (0.007) \\
\addlinespace[2pt]
Feedback study & Output-only and coarse-state & 192 & 85 & 88 & 86 & 93 & 0.814 \\
\addlinespace[2pt]
Controlled matrix & All cells & 864 & 385 & 422 & 405 & 445 & 0.001 \\
 & 0.5$\times$ budget & 288 & 25 & 24 & 28 & 37 & 0.131 \\
 & 1$\times$ budget & 288 & 169 & 178 & 175 & 193 & 0.105 \\
 & 1.5$\times$ budget & 288 & 191 & 220 & 202 & 215 & 0.016 \\
\addlinespace[2pt]
PPA track & Design stays correct & 384 & 260 & 199 & 212 & 294 & $<$0.001 \\
 & Design improved & 384 & 151 & 123 & 111 & 184 & $<$0.001 \\
\addlinespace[2pt]
Regrading & Accepted designs exact at 32 steps & 69 & 6/15 & 4/18 & 7/16 & 7/20 & 0.586 \\
\bottomrule
\end{tabular}
}
\par\smallskip{\footnotesize\raggedright $p$: share of 20,000 random within-instance permutations of the agent labels whose Cochran $Q$ is at least the observed one; for the twelve tasks, the Holm-adjusted value follows in parentheses. The feedback-study row excludes the full-state setting, which is part of the correctness track. Controlled-matrix instances are block, horizon, added condition, and budget. The regrading row is not matched: it gives, for each agent, how many of its accepted submissions among the regraded designs are exact on all 32 new sequences at 32 steps, and its $p$-value permutes the agent labels over the 69 submissions.\par}
\end{table}

\begin{table}[htbp]
\centering
\caption{\textbf{Pairwise differences on the correctness track, with paired bootstrap 95\% intervals.} Each pair is compared on the same 472 instances.}
\label{tab:T17-pairs}
{\small\tablefont\setlength{\tabcolsep}{5pt}\resultrows
\begin{tabular}{@{}lrrrr@{}}
\toprule
Pair & Accepted & Discordant & Difference & $p$ \\
\midrule
Sol 5.6 and Astra 6 & 289 / 235 & 111 / 57 & $+11.4$\,\textsubscript{\tiny(6.4--16.5)} & $<$0.001 ($<$0.001) \\
Sol 5.6 and Opus 5.5 & 289 / 264 & 95 / 70 & $+5.3$\,\textsubscript{\tiny(0.0--10.6)} & 0.061 (0.123) \\
Sol 5.6 and Fable 5.1 & 289 / 228 & 118 / 57 & $+12.9$\,\textsubscript{\tiny(8.1--18.0)} & $<$0.001 ($<$0.001) \\
Astra 6 and Opus 5.5 & 235 / 264 & 63 / 92 & $-6.1$\,\textsubscript{\tiny($-$10.8 to $-$0.8)} & 0.024 (0.073) \\
Astra 6 and Fable 5.1 & 235 / 228 & 81 / 74 & $+1.5$\,\textsubscript{\tiny($-$3.6 to 6.8)} & 0.630 (0.630) \\
Opus 5.5 and Fable 5.1 & 264 / 228 & 100 / 64 & $+7.6$\,\textsubscript{\tiny(2.5--12.7)} & 0.006 (0.024) \\
\bottomrule
\end{tabular}
}
\par\smallskip{\footnotesize\raggedright Discordant: instances on which only the first or only the second agent of the pair is accepted. Difference: first minus second, in percentage points. Intervals are percentile bootstraps (2,000 resamples of instances within each task, fixed seed). $p$: exact McNemar test, with the Holm-adjusted value over the six pairs in parentheses.\par}
\end{table}

\subsection{Overall score}
\label{app:add-score}
The overall score of Section~\ref{sec:score} multiplies a correctness-track score by a PPA-track score. The correctness score is the acceptance rate on each of the twelve correctness-track tasks, averaged over the tasks, so the tasks with 64 instances weigh no more than those with 24. The PPA score is the rate of improved designs on each of Qwen3.5-0.8B and 4B, Gemma 4 E4B, and Llama 3.2 3B, averaged over the four. Table~\ref{tab:leaderboard} gives the score, its two factors, and the other rates of both tracks. Intervals are paired percentile bootstraps: each resample draws instances with replacement within each task and keeps the four agents' outcomes on an instance together.

Fable 5.1 and Sol 5.6 score 25.0 and 24.5, and Astra 6 and Opus 5.5 score 16.8 and 16.5. The four differences between an agent of the first pair and one of the second are 7.7 to 8.5 points (Table~\ref{tab:T18-pairs}). The bootstrap intervals of all four exclude zero, and a paired permutation test gives a Holm-adjusted $p$ of 0.002 for each. Within each pair the difference is 0.5 and 0.3 points, and Fable 5.1 has the highest score in 58\% of the resamples against 42\% for Sol 5.6. We also test whether the ranking reproduces (Table~\ref{tab:T19-split-half}). Each split divides the matched instances of every correctness task and PPA target into two random halves and computes the score on each half. Both halves rank Sol 5.6 and Fable 5.1 above Astra 6 and Opus 5.5 in all 200 splits, against 17\% of splits when the four agents' outcomes are permuted within instances ($p=0.001$).

Which of the first two agents leads depends on how the tracks are combined (Table~\ref{tab:T18-definitions}). Sol 5.6 has the higher correctness score and Fable 5.1 the higher PPA score. Fable 5.1 is first under the product, and Sol 5.6 under the mean of the two track scores and when the sixteen tasks weigh equally, which gives the PPA track one quarter of the weight. Under every definition that includes the PPA track, both score above Astra 6 and Opus 5.5. We multiply the track scores because the benchmark asks for a design that is both correct and efficient: the product is high only when both rates are, and it is the probability that one attempt at a randomly drawn correctness task and one at a randomly drawn PPA target both succeed, if the two are independent.

Applying Equation~\ref{eq:ppa} at the clock each design achieves after place and route changes 107 of the 1,536 labels and gives scores of 23.7 for Sol 5.6, 22.8 for Fable 5.1, 14.9 for Astra 6, and 15.6 for Opus 5.5 (Table~\ref{tab:T06_scores}); each of the four differences between an agent of the first pair and one of the second has a Holm-adjusted $p$ of at most 0.002.

\begin{table}[htbp]
\centering
\caption{\textbf{Leaderboard values for both tracks.} Score: $100\,\bar c\,\bar q$ with a paired bootstrap 95\% interval (Section~\ref{sec:score}). Attempts are per agent, and all other values except Hours are percentages. Correctness track: $\bar c$, acceptance rate averaged over the twelve tasks; Accepted, every gate of Equation~\ref{eq:acceptance} passes, with a Wilson 95\% interval; Subm., the attempt ended with a submission before any cap; Public and Held-out, the frozen design is exact on that suite; Hours, agent-hours per accepted design. PPA track: $\bar q$, rate of improved designs (Equation~\ref{eq:ppa}) averaged over the four targets; Correct, the design stays exact; Routed, improved at the clock achieved after place and route; Hours, agent-hours per improved design. Hours include failed attempts.}
\label{tab:leaderboard}
{\footnotesize\tablefont\setlength{\tabcolsep}{1.5pt}\resultrows
\begin{tabular*}{\linewidth}{@{\extracolsep{\fill}}l*{11}{c}@{}}
\toprule
 & Both tracks & \multicolumn{6}{c}{Correctness track (472 attempts)} & \multicolumn{4}{c}{PPA track (384 attempts)} \\
\cmidrule(lr){2-2}\cmidrule(lr){3-8}\cmidrule(lr){9-12}
Agent & Score & $\bar c$ & Accepted & Subm. & Public & Held-out & Hours & $\bar q$ & Correct & Routed & Hours \\
\midrule
\textbf{Fable 5.1} & \cellcolor{blueC!16}\textbf{25.0}\,\textsubscript{\tiny(21.8--28.4)} & 52.2 & 48.3\,\textsubscript{\tiny(43.8--52.8)} & 66.9 & 55.3 & 57.2 & 31.4 & 47.9 & 76.6 & 43.8 & 24.9 \\
\textbf{Sol 5.6} & \cellcolor{blueC!16}\textbf{24.5}\,\textsubscript{\tiny(21.2--28.1)} & 62.4 & 61.2\,\textsubscript{\tiny(56.8--65.5)} & 77.1 & 66.5 & 70.3 & 23.9 & 39.3 & 67.7 & 38.0 & 31.6 \\
\textbf{Astra 6} & \cellcolor{blueC!12}\textbf{16.8}\,\textsubscript{\tiny(14.1--19.7)} & 52.4 & 49.8\,\textsubscript{\tiny(45.3--54.3)} & 74.8 & 60.6 & 63.3 & 29.3 & 32.0 & 51.8 & 28.4 & 39.8 \\
\textbf{Opus 5.5} & \cellcolor{blueC!12}\textbf{16.5}\,\textsubscript{\tiny(13.7--19.4)} & 57.1 & 55.9\,\textsubscript{\tiny(51.4--60.3)} & 77.5 & 64.2 & 66.3 & 26.3 & 28.9 & 55.2 & 27.3 & 43.0 \\
\bottomrule
\end{tabular*}
}
\par\smallskip{\footnotesize\raggedright Nano T2, nano T3, and Qwen3.5-0.8B come from the full-state setting of the feedback study (32 attempts per agent and task). Qwen3.5-4B, Gemma 4 E4B, and Llama 3.2 3B have 24 attempts per agent and task, and Gemma 3 4B and Llama 3.2-1B have 64. Rows are ordered by score. The first two agents are not separated, nor are the last two. Per-task results: Tables~\ref{tab:leaderboard_tasks} and~\ref{tab:ppa}.\par}
\end{table}

\begin{table}[htbp]
\centering
\caption{\textbf{Pairwise differences in the overall score, with paired bootstrap 95\% intervals.}}
\label{tab:T18-pairs}
{\small\tablefont\setlength{\tabcolsep}{7pt}\resultrows
\begin{tabular}{@{}lrr@{}}
\toprule
Pair & Difference & $p$ \\
\midrule
Sol 5.6 and Astra 6 & $+7.7$\,\textsubscript{\tiny(3.7--12.0)} & $<$0.001 (0.002) \\
Sol 5.6 and Opus 5.5 & $+8.0$\,\textsubscript{\tiny(3.8--12.3)} & $<$0.001 (0.002) \\
Sol 5.6 and Fable 5.1 & $-0.5$\,\textsubscript{\tiny($-$5.1 to 4.0)} & 0.836 (1.000) \\
Astra 6 and Opus 5.5 & $+0.3$\,\textsubscript{\tiny($-$3.6 to 4.2)} & 0.896 (1.000) \\
Astra 6 and Fable 5.1 & $-8.2$\,\textsubscript{\tiny($-$12.6 to $-$4.0)} & $<$0.001 (0.002) \\
Opus 5.5 and Fable 5.1 & $-8.5$\,\textsubscript{\tiny($-$12.8 to $-$4.3)} & $<$0.001 (0.002) \\
\bottomrule
\end{tabular}
}
\par\smallskip{\footnotesize\raggedright Difference: first minus second, in score points. Intervals use the paired bootstrap of the score. $p$: paired permutation test that swaps the outcomes of the two agents within an instance (20,000 draws), with the Holm-adjusted value over the six pairs in parentheses.\par}
\end{table}

\begin{table}[htbp]
\centering
\caption{\textbf{The score under other definitions.} Each row applies one definition to the same attempts.}
\label{tab:T18-definitions}
{\small\tablefont\setlength{\tabcolsep}{7pt}\resultrows
\begin{tabular}{@{}lrrrr@{}}
\toprule
Definition & Fable 5.1 & Sol 5.6 & Astra 6 & Opus 5.5 \\
\midrule
Product of the two track scores (the score) & \textbf{25.0} & 24.5 & 16.8 & 16.5 \\
Mean of the two track scores & 50.1 & \textbf{50.8} & 42.2 & 43.0 \\
Sixteen tasks, equal weight & 51.1 & \textbf{56.6} & 47.3 & 50.1 \\
Correctness track only & 52.2 & \textbf{62.4} & 52.4 & 57.1 \\
PPA track only & \textbf{47.9} & 39.3 & 32.0 & 28.9 \\
\bottomrule
\end{tabular}
}
\par\smallskip{\footnotesize\raggedright Sixteen tasks: the twelve correctness tasks and the four PPA targets, each with equal weight. The highest value in each row is bold.\par}
\end{table}

\begin{table}[htbp]
\centering
\caption{\textbf{The agent ranking reproduces across random halves of the instances.} Each of 200 splits divides the matched instances of every task, cell, or PPA target into two halves, and the order on one half is compared with the order on the other.}
\label{tab:T19-split-half}
{\small\tablefont\setlength{\tabcolsep}{4pt}\resultrows
\begin{tabular}{@{}lrlrrr@{}}
\toprule
Study & $n$ & Order on all instances & Agreement & Null & $p$ \\
\midrule
Overall score & 856 & Fable $>$ Sol $>$ Astra $>$ Opus & 100.0\% & 16.7\% & 0.001 \\
 & & \quad Kendall $\tau$ between halves & 0.43 & -0.01 & 0.062 \\
Controlled matrix & 864 & Fable $>$ Astra $>$ Opus $>$ Sol & 0.80 & 0.01 & 0.001 \\
PPA track & 384 & Fable $>$ Sol $>$ Astra $>$ Opus & 0.87 & -0.01 & 0.001 \\
Correctness track & 472 & Sol $>$ Opus $>$ Astra $>$ Fable & 0.78 & 0.00 & 0.001 \\
\bottomrule
\end{tabular}
}
\par\smallskip{\footnotesize\raggedright Overall score: the share of splits in which both halves place the same two agents on top, then the Kendall $\tau$ between the two halves' scores. Other rows: Kendall $\tau$ between the halves' acceptance rates (PPA track: improved designs). Null: the same statistic after permuting the four agents' outcomes within each instance. $p$: share of 1,000 permuted data sets, each averaged over 40 splits, whose statistic is at least the observed one. Controlled-matrix instances are matched blocks within each horizon, condition, and budget cell.\par}
\end{table}

\section{PPA track and backend}
\label{app:ppa}
\ifcompact\begin{figure}[tbp]
\centering
\includegraphics[width=\linewidth]{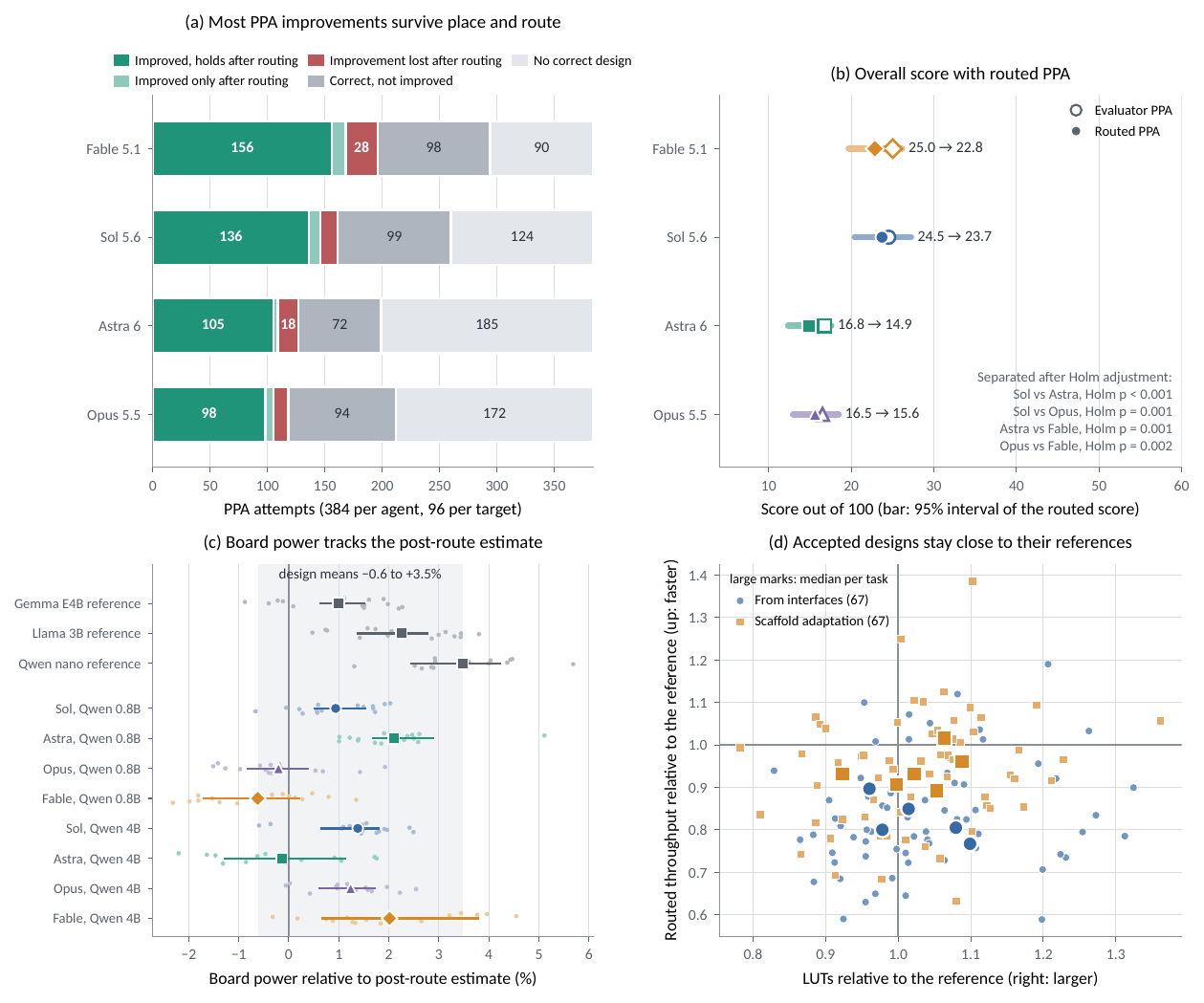}
\caption{\textbf{PPA designs after place and route, and agent designs on the board.} (a) Outcome of the 384 PPA attempts per agent, 96 on each target, when each correct design is placed and routed with three seeds and judged by Equation~\ref{eq:ppa} at its achieved clock. (b) Overall score with the evaluator's PPA values (hollow) and with routed PPA values (filled); bars are 95\% bootstrap intervals of the routed score. (c) Board power relative to the post-route activity estimate for three references and eight agent designs: mean of 15 measurement windows (marker), range of the three seed means (line), and single windows (dots); the shaded band spans the design means. (d) Routed throughput and LUTs of the 134 accepted correctness-track designs whose three routes complete, relative to the task's reference; large marks are per-task medians.}
\label{fig:key_hardware}
\end{figure}
\fi

\subsection{Power methodology and reference schedules}
\label{app:power}
Table~\ref{tab:reference_hardware} compares the baseline and optimized schedules of the Qwen3.5-0.8B and 4B reference designs before place and route.
\begin{table}[H]
\centering
\caption{\textbf{The optimized schedules raise throughput by 12.0\% and estimated energy efficiency by 7.8\%.} Baseline and optimized schedules of the Qwen3.5-0.8B and 4B reference designs.}
\label{tab:reference_hardware}
{\small\tablefont\setlength{\tabcolsep}{3pt}\resultrows
\begin{tabular*}{\linewidth}{@{\extracolsep{\fill}}l*{5}{c}}
\toprule
Workload & {\resultstack\begin{tabular}[b]{@{}c@{}}Throughput\\(tokens/s)\end{tabular}} & {\resultstack\begin{tabular}[b]{@{}c@{}}Throughput\\gain\end{tabular}} & {\resultstack\begin{tabular}[b]{@{}c@{}}Est. efficiency\\gain\end{tabular}} & {\resultstack\begin{tabular}[b]{@{}c@{}}Power\\(W)\end{tabular}} & {\resultstack\begin{tabular}[b]{@{}c@{}}LUTs\\(approx.)\end{tabular}} \\
\midrule
Qwen3.5-0.8B & {\resultstack\begin{tabular}[b]{@{}c@{}}3.999\\\textbf{4.479}\end{tabular}} & \cellcolor{tealC!36}\textbf{+12.0\%} & \cellcolor{tealC!25}\textbf{+7.8\%} & {\resultstack\begin{tabular}[b]{@{}c@{}}8.98\\\textbf{9.33}\end{tabular}} & {\resultstack\begin{tabular}[b]{@{}c@{}}61.8k\\\textbf{62.1k}\end{tabular}} \\
Qwen3.5-4B & {\resultstack\begin{tabular}[b]{@{}c@{}}0.800\\\textbf{0.896}\end{tabular}} & \cellcolor{tealC!36}\textbf{+12.0\%} & \cellcolor{tealC!25}\textbf{+7.8\%} & {\resultstack\begin{tabular}[b]{@{}c@{}}10.09\\\textbf{10.48}\end{tabular}} & {\resultstack\begin{tabular}[b]{@{}c@{}}69.0k\\\textbf{69.3k}\end{tabular}} \\
\bottomrule
\end{tabular*}
}
\par\smallskip{\footnotesize\raggedright Top: baseline; bold: optimized. Within each pair: same arithmetic and memory model, 126\,MHz, batch size one, eight-token prompt. Power is FPGA plus DRAM; LUTs are approximate. Gains are relative to baseline; efficiency is throughput/power.\par}
\end{table}

\paragraph{Paired comparison.}
Both optimized schedules raise throughput by 12.0\% and add about 300 LUTs (under 1\%) in the pre-route estimate. After place and route (Appendix~\ref{app:add-physical}), the optimized designs use 61.5k (0.8B) and 69.2k (4B) LUTs, against the estimated 62.1k and 69.3k. Both designs time-multiplex a shared compute engine and stream weights from external storage, so the larger model adds computation and memory traffic without a proportionally larger datapath. Within each workload, the two schedules use the same arithmetic and therefore have the same fidelity.

The optimized schedules raise FPGA-plus-DRAM power by about 3.9\%, so the 12.0\% throughput gain gives an estimated efficiency gain of 7.8\%. Energy per token falls from 2.2456 to 2.0831\,J (Qwen3.5-0.8B) and from 12.6125 to 11.6964\,J (Qwen3.5-4B).

\paragraph{Evidence levels.}
We report power at three levels: estimates, post-route reports, and board measurements. Estimates come from a power budget, which links power and work counts to energy, or from a post-synthesis power report, which assumes a clock, voltage, temperature, process, and switching activity. Post-route reports add physical implementation information, and board measurements observe the running design (Appendix~\ref{app:add-physical}). Vendor guidance~\citep{amd_power} emphasizes both switching annotation and implementation information.

\paragraph{Sensitivity.}
Let $g_R=R_1/R_0$ and $g_P=P_1/P_0$. The energy-efficiency gain is $g_R/g_P-1$, so it vanishes when power rises as much as throughput: the 10.5\% throughput gain for Qwen3.5-nano gives no energy benefit if power also rises by 10.5\%, and the 12.0\% gains for 0.8B and 4B break even at a 12.0\% power increase.

\subsection{Physical implementation and board power}
\label{app:add-physical}
The baseline and optimized schedules of the 0.8B and 4B reference designs were each placed and routed with three implementation seeds and measured on the board with five repeats per seed (Figure~\ref{fig:F09_physical}, Tables~\ref{tab:T10-results} and~\ref{tab:T10-components}). Every baseline implementation and every optimized 4B implementation meets 126\,MHz. The optimized 0.8B design meets it on one of three seeds; the other two close at a lower clock, so its mean operating clock is 125.3\,MHz and its throughput gain is 11.4--11.8\%. Mean post-route LUT counts change by $-0.2$\% for both workloads, where the estimates of Table~\ref{tab:reference_hardware} showed an increase of about 300 LUTs, and BRAM rises by 2.2\% and 1.6\%.

Measured board power rises by 3.1\% (0.8B) and 3.7\% (4B), and energy per token falls by 7.8\% and 7.1\%. The resulting efficiency gains, 8.5\% and 7.7\%, lie within a percentage point of the estimated 7.8\%. Paired by seed and repeat, all 15 board pairs of each workload show an efficiency gain, from 3.5\% to 13.6\% for 0.8B and from 4.7\% to 12.4\% for 4B, with seed means of 7.5--9.3\% and 6.9--8.2\%. All implementations target the AMD Alveo U250 device xcu250-figd2104-2L-e and were built with Vivado 2024.1, the version recorded in the Vivado logs, timing reports, and bitstream headers. Board power is read from the U250's on-board sensors through XRT 2.16.204 and sums the core (0.85\,V), auxiliary (1.8\,V), and DRAM (1.2\,V) rails, sampled at 1\,kHz. Each seed is measured in five non-overlapping 10\,s windows. The reported board power is the total of the three rails during these windows, the same quantity as the post-route estimates it is compared with; each rail's median idle power over the 30\,s before the workload is recorded and is not subtracted. The memory-service model is cycle-relative: a request waits 48 cycles, each 64-byte beat then takes one cycle, bursts have 8 to 16 beats, and the interface delivers at most one beat per cycle at 126\,MHz; under the bounded-memory-stall condition of the regrading and the controlled matrix, response delays are drawn uniformly from 0 to 256 cycles. The held-out backpressure test and the local checks use shorter stalls of one to three cycles (Appendix~\ref{app:tests}).

\paragraph{Further references and agent designs.}
The nano, Gemma 4 E4B, and Llama 3.2 3B references were placed and routed with three seeds each and measured on the board with five windows per seed (Table~\ref{tab:T08_references}). All operate at 126\,MHz. Gemma 4 E4B decodes 0.940 tokens/s with 73.6k LUTs and an estimated 10.61\,W, Llama 3.2 3B 1.120 tokens/s with 67.1k LUTs and 10.07\,W, and nano 365.12 tokens/s with 46.8k LUTs and 5.92\,W; the nano rate comes from a single prepared transaction. The Gemma 3 4B and Llama 3.2-1B references were placed and routed with three seeds each and close timing at 126\,MHz on every seed: Gemma 3 4B decodes 0.883 tokens/s with 73.5k LUTs and an estimated 10.55\,W, and Llama 3.2-1B 2.961 tokens/s with 60.4k LUTs and 8.99\,W. One improved PPA design per agent on each Qwen3.5 target, the one with the largest nominal gain, was also measured on the board, with three seeds and five windows each. Across these eight designs and the three references, mean board power differs from the post-route activity estimate by $-0.62$ to $+3.47$\%, and seed means by $-1.70$ to $+4.21$\% (Table~\ref{tab:T09_power}, Figure~\ref{fig:F06_power_and_references}).

\paragraph{FPGA and ASIC implementation flows.}
\label{app:fpga_asic}
Every physical result in this paper comes from the FPGA flow described above. Vivado implementation places a synthesized netlist onto the U250's existing resources (look-up tables, flip-flops, block RAM, DSP slices, and programmable interconnect), routes it, checks timing, and writes a bitstream~\citep{vivado_impl}. The quantities it yields are resource counts, the clock the routed design achieves, and, on the board, rail power. OpenROAD~\citep{openroad} and the flows built on it target ASIC design: starting from RTL and a process design kit, they synthesize to a standard-cell library and carry the design through floorplanning, macro and standard-cell placement, clock-tree synthesis, global and detailed routing, and layout generation in GDSII. Their outputs are a cell area, routed timing, congestion, and power for a chip that would be fabricated, with design-rule checks before signoff. The two flows therefore answer different questions about the same RTL. The contract, the oracle, and the simulation-based tests are independent of the implementation flow; the PPA criterion of Equation~\ref{eq:ppa} is stated in FPGA LUTs at a 126\,MHz clock. Extending the PPA track to ASICs would require choosing a PDK and standard-cell library, generating or selecting memory macros for the caches and weight buffers, writing timing constraints for that process, adding physical verification, and restating the area term in cell area. Such results would not be comparable with the LUT counts and board power reported here.

\begin{figure}[tbp]
\centering
\includegraphics[width=0.94\linewidth]{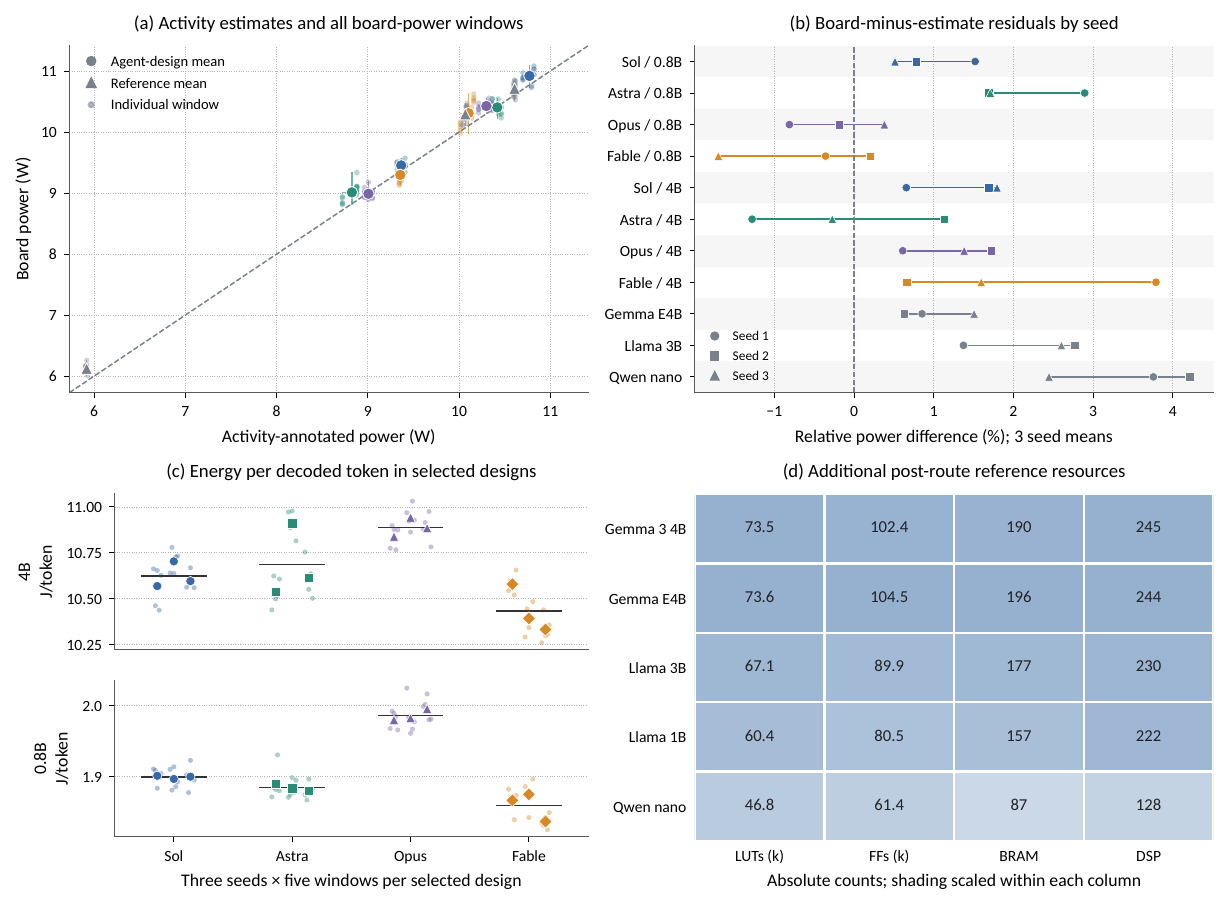}
\caption{Power and further backend references. Small points are individual windows; large power points are design means, and ranges span the observed windows. Residual symbols mark the three seed means, with five windows per seed. Energy panels use separate vertical scales for the two targets; large points are seed means and black bars design means. Resource-cell shading is scaled within columns, and numbers give absolute resources. Agent power uses one preselected design per agent and target.}
\label{fig:F06_power_and_references}
\end{figure}
\begin{table}[H]
\centering
\caption{\textbf{Backend reference values for additional normalization targets.}}
\label{tab:T08_references}
{\footnotesize\tablefont\setlength{\tabcolsep}{3.5pt}\renewcommand{\arraystretch}{1.04}
\begin{tabular*}{\linewidth}{@{\extracolsep{\fill}}lrrrrrrr@{}}
\toprule
\multicolumn{1}{c}{} & \multicolumn{2}{c}{Operating point} & \multicolumn{4}{c}{Implementation resources} & \multicolumn{1}{c}{Power} \\
\cmidrule(lr){2-3}\cmidrule(lr){4-7}\cmidrule(lr){8-8}
Target & MHz & Rate & LUT & FF & BRAM & DSP & Power W \\
\midrule
Gemma 3 4B & 126.0 & 0.883 & 73,536 & 102,370 & 190 & 245 & 10.55 \\
Gemma E4B & 126.0 & 0.940 & 73,594 & 104,546 & 196 & 244 & 10.61 \\
Llama 3B & 126.0 & 1.120 & 67,149 & 89,898 & 177 & 230 & 10.07 \\
Llama 1B & 126.0 & 2.961 & 60,436 & 80,544 & 157 & 222 & 8.99 \\
Qwen nano & 126.0 & 365.120 & 46,836 & 61,428 & 87 & 128 & 5.92 \\
\bottomrule
\end{tabular*}}
\par\smallskip{\footnotesize\raggedright Values summarize three route seeds per reference. Rate is tokens/s under the declared workload; nano uses a single-transaction probe. Power covers FPGA plus DRAM and excludes host. Numerical fidelity and RTL exactness are separate qualification requirements.\par}
\end{table}

\begin{table}[H]
\centering
\caption{\textbf{Activity-annotated and board-power comparison.}}
\label{tab:T09_power}
{\footnotesize\tablefont\setlength{\tabcolsep}{3.5pt}\renewcommand{\arraystretch}{1.04}
\begin{tabular*}{\linewidth}{@{\extracolsep{\fill}}lrrrrr@{}}
\toprule
\multicolumn{1}{c}{} & \multicolumn{1}{c}{Nested repeats} & \multicolumn{2}{c}{Boundary power} & \multicolumn{2}{c}{Relative residual} \\
\cmidrule(lr){2-2}\cmidrule(lr){3-4}\cmidrule(lr){5-6}
Design & Windows & Estimate W & Board W & Difference \% & \shortstack{Seed-mean range\\\%} \\
\midrule
Sol / 0.8B & 15/15 & 9.366 & 9.453 & 0.94 & 0.51--1.52 \\
Astra / 0.8B & 15/15 & 8.826 & 9.011 & 2.10 & 1.69--2.89 \\
Opus / 0.8B & 15/15 & 9.007 & 8.988 & -0.21 & -0.81--0.38 \\
Fable / 0.8B & 15/15 & 9.356 & 9.298 & -0.62 & -1.70--0.20 \\
\addlinespace[3pt]
Sol / 4B & 15/15 & 10.772 & 10.921 & 1.38 & 0.66--1.79 \\
Astra / 4B & 15/15 & 10.419 & 10.404 & -0.14 & -1.28--1.13 \\
Opus / 4B & 15/15 & 10.300 & 10.427 & 1.24 & 0.61--1.72 \\
Fable / 4B & 15/15 & 10.104 & 10.309 & 2.02 & 0.66--3.79 \\
\addlinespace[3pt]
Gemma E4B & 15/15 & 10.609 & 10.715 & 1.00 & 0.63--1.51 \\
Llama 3B & 15/15 & 10.068 & 10.295 & 2.25 & 1.37--2.77 \\
Qwen nano & 15/15 & 5.917 & 6.123 & 3.47 & 2.45--4.21 \\
\bottomrule
\end{tabular*}}
\par\smallskip{\footnotesize\raggedright Five windows are nested within each of three implementation seeds. Difference is board/estimate minus one. The eight agent designs are selected by original nominal gain.\par}
\end{table}

\begin{figure}[htbp]
\centering
\includegraphics[width=\linewidth,height=.62\textheight,keepaspectratio]{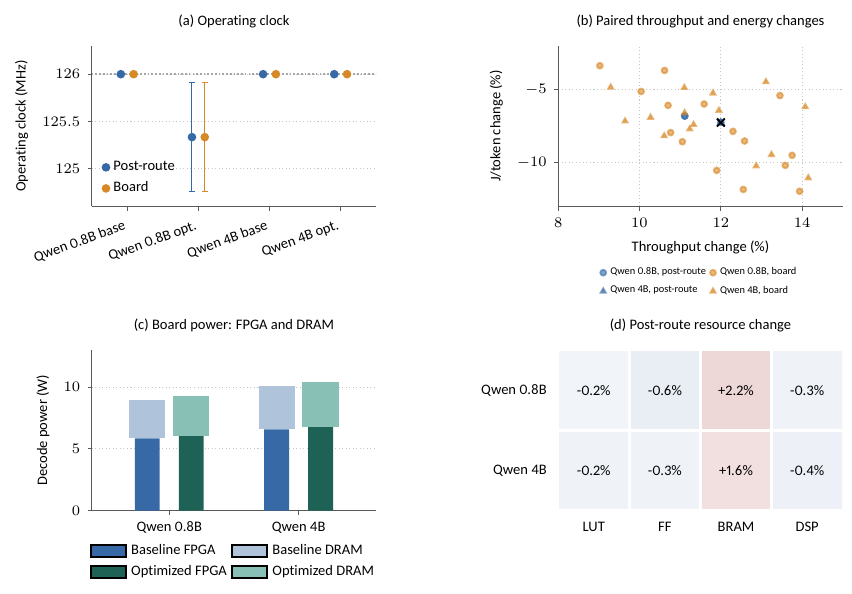}
\caption{Physical implementation of paired baseline and optimized Qwen3.5-0.8B and 4B designs. (a) Operating clock, mean and standard deviation over three implementation seeds. (b) Optimized-to-baseline changes in throughput and J/token, paired by target, stage, seed, and repeat; crosses mark the earlier 126\,MHz pairs. (c) Board power split into FPGA and attached-DRAM components. (d) Change in mean post-route resources. Throughput uses the operating clock, and all energy values share the FPGA-plus-DRAM decode boundary.}
\label{fig:F09_physical}
\end{figure}

\begin{table}[htbp]
\centering
\caption{\textbf{Post-route and board results: timing, throughput, and energy.}}
\label{tab:T10-results}
{\small\tablefont\setlength{\tabcolsep}{3pt}\resultrows
\begin{tabular*}{\linewidth}{@{\extracolsep{\fill}}lllrrrrrrrr@{}}
\toprule
Target & Stage & Schedule & MHz & Timing & LUT k & Token/s & Gain \% & W & J/token & Eff. \% \\
\midrule
Qwen 0.8B & Post-route & base & 126.0 & 3/3 & 61.6 & 3.999 & \textemdash & 8.95 & 2.237 & \textemdash \\
Qwen 0.8B & Post-route & opt. & 125.3 & \cellcolor{orange!18}1/3 & 61.5 & 4.455 & \cellcolor{tealC!34}+11.4 & 9.27 & 2.081 & \cellcolor{tealC!25}+7.5 \\
Qwen 0.8B & Board & base & 126.0 & 3/3 & 61.6 & 3.985 & \textemdash & 8.98 & 2.254 & \textemdash \\
Qwen 0.8B & Board & opt. & 125.3 & \cellcolor{orange!18}1/3 & 61.5 & 4.457 & \cellcolor{tealC!35}+11.8 & 9.27 & 2.079 & \cellcolor{tealC!27}+8.5 \\
Qwen 4B & Post-route & base & 126.0 & 3/3 & 69.3 & 0.800 & \textemdash & 10.05 & 12.564 & \textemdash \\
Qwen 4B & Post-route & opt. & 126.0 & 3/3 & 69.2 & 0.896 & \cellcolor{tealC!35}+12.0 & 10.44 & 11.650 & \cellcolor{tealC!26}+7.8 \\
Qwen 4B & Board & base & 126.0 & 3/3 & 69.3 & 0.797 & \textemdash & 10.05 & 12.602 & \textemdash \\
Qwen 4B & Board & opt. & 126.0 & 3/3 & 69.2 & 0.891 & \cellcolor{tealC!35}+11.7 & 10.42 & 11.703 & \cellcolor{tealC!25}+7.7 \\
\bottomrule
\end{tabular*}
}
\par\smallskip{\footnotesize\raggedright Clock and LUT values are means over three implementation seeds. Timing counts implementations that meet 126\,MHz (orange when some miss it). Throughput and FPGA-plus-DRAM power are pooled over equal token counts; J/token is power divided by throughput; gains compare each optimized row with its baseline (teal). Board rows pool five repeats per seed.\par}
\end{table}
\begin{table}[htbp]
\centering
\caption{\textbf{Post-route and board results: slack, resources, and power components.}}
\label{tab:T10-components}
{\small\tablefont\setlength{\tabcolsep}{1.5pt}\resultrows
\begin{tabular*}{\linewidth}{@{\extracolsep{\fill}}lllrrrrrrrr@{}}
\toprule
Target & Stage & Schedule & WNS ns & TNS ns & FF & BRAM & DSP & Mcycles/token & FPGA W & DRAM W \\
\midrule
Qwen 0.8B & Post-route & base & 0.17 & 0.00 & 79,319 & 151.3 & 224 & 31.51 & 5.82 & 3.13 \\
Qwen 0.8B & Post-route & opt. & -0.06 & -0.06 & 78,843 & 154.7 & 223 & 28.13 & 6.02 & 3.25 \\
Qwen 0.8B & Board & base & 0.17 & 0.00 & 79,319 & 151.3 & 224 & 31.62 & 5.84 & 3.14 \\
Qwen 0.8B & Board & opt. & -0.06 & -0.06 & 78,843 & 154.7 & 223 & 28.12 & 6.03 & 3.23 \\
Qwen 4B & Post-route & base & 0.16 & 0.00 & 91,389 & 168.7 & 232 & 157.50 & 6.54 & 3.52 \\
Qwen 4B & Post-route & opt. & 0.05 & 0.00 & 91,122 & 171.3 & 231 & 140.62 & 6.79 & 3.65 \\
Qwen 4B & Board & base & 0.16 & 0.00 & 91,389 & 168.7 & 232 & 158.01 & 6.56 & 3.49 \\
Qwen 4B & Board & opt. & 0.05 & 0.00 & 91,122 & 171.3 & 231 & 141.45 & 6.77 & 3.66 \\
\bottomrule
\end{tabular*}
}
\par\smallskip{\footnotesize\raggedright WNS is the worst negative slack over seeds and TNS the mean total negative slack. Cycles per token and resources are means. Power components are time-weighted over equal token counts and sum to boundary power.\par}
\end{table}

\subsection{Schedule mechanisms and operating conditions}
\label{app:add-schedule}
A cycle model shows where the optimized schedule saves time. For both workloads, compute, memory waits, and control take 65, 28, and 7\% of the cycles per token under the baseline schedule and 73, 22, and 5\% under the optimized one. Doubling memory bandwidth raises the Qwen3.5-0.8B baseline throughput by 14\%, from 3.999 to 4.563 tokens/s. In the cycle and power model for Qwen3.5-4B, enabling buffering, overlap, and the revised schedule together lowers cycles per token by 10.7\%, and overlap gives the largest single reduction. In sweeps that vary bandwidth, latency, prompt length, or resource budget one at a time around the nominal operating point, the optimized schedule keeps its advantage, and its relative gain is largest when memory latency is high or bandwidth is low.

\subsection{PPA track: agent optimization}
\label{app:add-optimization}
Starting from the correct reference design of Qwen3.5-0.8B, Qwen3.5-4B, Gemma 4 E4B, or Llama 3.2 3B, agents try to improve throughput, resources, or energy efficiency while keeping the design correct (Figure~\ref{fig:F11_optimization}, Table~\ref{tab:T12-outcomes}). Each agent makes 96 attempts per target. Of the 1,536 attempts, 965 keep the design correct and 569 also meet Equation~\ref{eq:ppa}, where $R_0$ and $L_0$ are the throughput and LUT count of the starting design: for 0.8B and 4B the baseline of Table~\ref{tab:reference_hardware} (3.999 and 0.800 tokens/s, 61.8k and 69.0k LUTs), and for Gemma 4 E4B and Llama 3.2 3B the evaluator's values for their reference designs (0.940 and 1.120 tokens/s, 74.4k and 66.6k LUTs). Recomputing the three gains from the measured throughput, LUT, and energy values and applying Equation~\ref{eq:ppa} to each attempt reproduces all 1,536 labels.

Per agent and target, between 18 of 96 (Opus 5.5, 0.8B) and 55 of 96 (Fable 5.1, 4B) attempts are improved, at 20.7 to 51.8 agent-hours per improved design. Among the 569 improved designs, the median throughput gain per target is 14.2 to 14.5\%, the median efficiency gain 10.9 to 11.2\%, and the median LUT change $-1.1$ to $-0.7$\%. Because the criterion caps LUT growth, this subset is selected partly on LUT change; over all 965 correct submissions, median gains are smaller and LUT use rises slightly (Table~\ref{tab:T12-changes}). The other 571 attempts end without a correct design, from 90 of 384 for Fable 5.1 to 185 of 384 for Astra 6 (Table~\ref{tab:T15-optimization}). Of these, 194 submit an incorrect design, and 377 reach a cap (207 on wall time, 88 on tool calls, and 82 on output tokens) with RTL that fails the correctness checks. Because every attempt here starts from the same correct design, these gains can be compared across agents, unlike the throughput and LUT counts of accepted correctness-track designs in Appendix~\ref{sec:agent_hardware}.

\begin{figure}[htbp]
\centering
\includegraphics[width=\linewidth,height=.62\textheight,keepaspectratio]{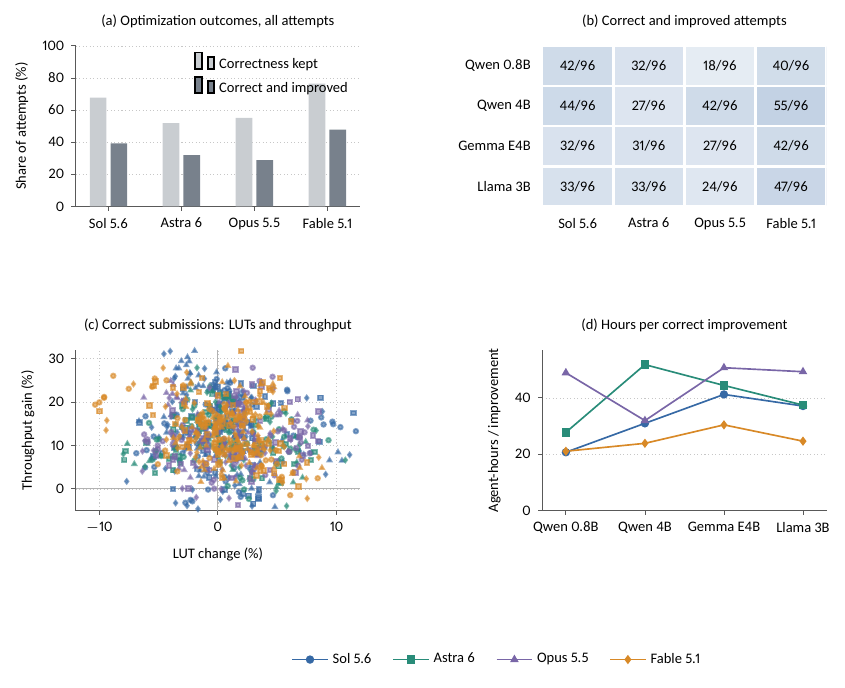}
\caption{PPA track: 1,536 attempts, 96 per target and agent, each starting from the correct reference design. (a) Share of all attempts that keep correctness, and that also meet the PPA criterion. (b) Correct and improved attempts by target and agent. (c) Throughput gain against LUT change for correct submissions, including unchanged designs and regressions. (d) Agent-hours per improved design, counting every attempt. An attempt meets the PPA criterion (Equation~\ref{eq:ppa}) when its throughput is at least 5\% above the starting design and its LUT count at most 2\% above it.}
\label{fig:F11_optimization}
\end{figure}

\begin{table}[htbp]
\centering
\caption{\textbf{PPA-track outcomes by target and agent, with Wilson 95\% intervals.}}
\label{tab:T12-outcomes}
{\small\tablefont\setlength{\tabcolsep}{7pt}\resultrows
\begin{tabular}{@{}llrrr}
\toprule
Target & Agent & Correct (\%) & Improved (\%) & h/improved \\
\midrule
Qwen 0.8B & Sol 5.6 & 72/96 (75.0) & \cellcolor{blueC!24}42/96 (43.8)\,\textsubscript{\tiny(34.3--53.7)} & 20.70 \\
Qwen 0.8B & Astra 6 & 61/96 (63.5) & \cellcolor{blueC!19}32/96 (33.3)\,\textsubscript{\tiny(24.7--43.2)} & 27.76 \\
Qwen 0.8B & Opus 5.5 & 46/96 (47.9) & \cellcolor{blueC!13}18/96 (18.8)\,\textsubscript{\tiny(12.2--27.7)} & 48.90 \\
Qwen 0.8B & Fable 5.1 & 63/96 (65.6) & \cellcolor{blueC!23}40/96 (41.7)\,\textsubscript{\tiny(32.3--51.7)} & 21.05 \\
Qwen 4B & Sol 5.6 & 63/96 (65.6) & \cellcolor{blueC!25}44/96 (45.8)\,\textsubscript{\tiny(36.2--55.8)} & 30.96 \\
Qwen 4B & Astra 6 & 39/96 (40.6) & \cellcolor{blueC!17}27/96 (28.1)\,\textsubscript{\tiny(20.1--37.8)} & 51.80 \\
Qwen 4B & Opus 5.5 & 62/96 (64.6) & \cellcolor{blueC!24}42/96 (43.8)\,\textsubscript{\tiny(34.3--53.7)} & 31.94 \\
Qwen 4B & Fable 5.1 & 83/96 (86.5) & \cellcolor{blueC!30}55/96 (57.3)\,\textsubscript{\tiny(47.3--66.7)} & 23.87 \\
Gemma 4 E4B & Sol 5.6 & 59/96 (61.5) & \cellcolor{blueC!19}32/96 (33.3)\,\textsubscript{\tiny(24.7--43.2)} & 41.22 \\
Gemma 4 E4B & Astra 6 & 48/96 (50.0) & \cellcolor{blueC!19}31/96 (32.3)\,\textsubscript{\tiny(23.8--42.2)} & 44.41 \\
Gemma 4 E4B & Opus 5.5 & 56/96 (58.3) & \cellcolor{blueC!17}27/96 (28.1)\,\textsubscript{\tiny(20.1--37.8)} & 50.70 \\
Gemma 4 E4B & Fable 5.1 & 75/96 (78.1) & \cellcolor{blueC!24}42/96 (43.8)\,\textsubscript{\tiny(34.3--53.7)} & 30.42 \\
Llama 3B & Sol 5.6 & 66/96 (68.8) & \cellcolor{blueC!20}33/96 (34.4)\,\textsubscript{\tiny(25.6--44.3)} & 37.10 \\
Llama 3B & Astra 6 & 51/96 (53.1) & \cellcolor{blueC!20}33/96 (34.4)\,\textsubscript{\tiny(25.6--44.3)} & 37.44 \\
Llama 3B & Opus 5.5 & 48/96 (50.0) & \cellcolor{blueC!16}24/96 (25.0)\,\textsubscript{\tiny(17.4--34.5)} & 49.28 \\
Llama 3B & Fable 5.1 & 73/96 (76.0) & \cellcolor{blueC!27}47/96 (49.0)\,\textsubscript{\tiny(39.2--58.8)} & 24.64 \\
\bottomrule
\end{tabular}
}
\par\smallskip{\footnotesize\raggedright The PPA criterion (Equation~\ref{eq:ppa}) requires the submission to remain correct, to raise throughput by at least 5\% over the starting design, and to use at most 2\% more LUTs. Every attempt counts in each denominator and in the hours per improved design.\par}
\end{table}
\begin{table}[htbp]
\centering
\caption{\textbf{Hardware changes among correct PPA-track submissions.}}
\label{tab:T12-changes}
{\small\tablefont\setlength{\tabcolsep}{4pt}\resultrows
\begin{tabular*}{\linewidth}{@{\extracolsep{\fill}}llrrrrr@{}}
\toprule
Target & Agent & Throughput \% & LUT saved \% & Eff. gain \% & Tools med. & Tokens k \\
\midrule
Qwen 0.8B & Sol 5.6 & \cellcolor{tealC!17}+12.9 & \cellcolor{orange!8}-0.3 & \cellcolor{tealC!15}+9.8 & 313.0 & 196.6 \\
Qwen 0.8B & Astra 6 & \cellcolor{tealC!16}+11.3 & \cellcolor{orange!9}-1.3 & \cellcolor{tealC!14}+9.2 & 305.0 & 194.3 \\
Qwen 0.8B & Opus 5.5 & \cellcolor{tealC!16}+11.3 & \cellcolor{orange!10}-2.6 & \cellcolor{tealC!14}+9.3 & 321.5 & 193.7 \\
Qwen 0.8B & Fable 5.1 & \cellcolor{tealC!16}+11.6 & \cellcolor{orange!8}-0.1 & \cellcolor{tealC!14}+9.5 & 312.0 & 191.7 \\
Qwen 4B & Sol 5.6 & \cellcolor{tealC!18}+14.4 & \cellcolor{tealC!8}+0.2 & \cellcolor{tealC!16}+12.2 & 400.0 & 234.6 \\
Qwen 4B & Astra 6 & \cellcolor{tealC!15}+10.7 & \cellcolor{orange!8}-0.4 & \cellcolor{tealC!14}+8.1 & 397.0 & 232.9 \\
Qwen 4B & Opus 5.5 & \cellcolor{tealC!16}+12.5 & \cellcolor{orange!8}-0.4 & \cellcolor{tealC!15}+9.9 & 390.0 & 230.4 \\
Qwen 4B & Fable 5.1 & \cellcolor{tealC!16}+11.7 & \cellcolor{orange!9}-1.2 & \cellcolor{tealC!15}+9.8 & 402.0 & 222.5 \\
Gemma 4 E4B & Sol 5.6 & \cellcolor{tealC!18}+14.8 & \cellcolor{orange!8}-0.6 & \cellcolor{tealC!16}+11.5 & 475.5 & 265.9 \\
Gemma 4 E4B & Astra 6 & \cellcolor{tealC!15}+9.6 & \cellcolor{tealC!8}+0.2 & \cellcolor{tealC!13}+7.8 & 463.0 & 274.6 \\
Gemma 4 E4B & Opus 5.5 & \cellcolor{tealC!16}+11.6 & \cellcolor{orange!9}-1.7 & \cellcolor{tealC!13}+7.5 & 457.0 & 270.4 \\
Gemma 4 E4B & Fable 5.1 & \cellcolor{tealC!18}+14.6 & \cellcolor{orange!9}-1.4 & \cellcolor{tealC!15}+10.0 & 456.0 & 269.6 \\
Llama 3B & Sol 5.6 & \cellcolor{tealC!18}+14.2 & \cellcolor{orange!8}-0.3 & \cellcolor{tealC!15}+9.9 & 442.5 & 254.9 \\
Llama 3B & Astra 6 & \cellcolor{tealC!16}+12.2 & \cellcolor{orange!8}-0.7 & \cellcolor{tealC!13}+7.9 & 446.0 & 264.3 \\
Llama 3B & Opus 5.5 & \cellcolor{tealC!16}+11.5 & \cellcolor{orange!9}-1.1 & \cellcolor{tealC!14}+8.4 & 440.0 & 250.7 \\
Llama 3B & Fable 5.1 & \cellcolor{tealC!17}+12.7 & \cellcolor{orange!9}-0.8 & \cellcolor{tealC!15}+9.7 & 422.0 & 260.9 \\
\bottomrule
\end{tabular*}
}
\par\smallskip{\footnotesize\raggedright Hardware columns give median paired changes among correct submissions. Efficiency gain is $(R/P)/(R_0/P_0)-1$, and LUT saved is the negated LUT change. Tool and token medians cover all attempts.\par}
\end{table}

\paragraph{Place and route of agent designs.}
Each of the 965 correct designs was placed and routed with three implementation seeds, 2,895 runs in all: 2,855 complete, 27 fail to route, and 13 time out, so 925 designs complete all three (Figure~\ref{fig:F04_postroute_ppa}, Table~\ref{tab:T05_ppa}). After routing, a design counts as improved when it is correct, all three routes complete, its mean throughput at the achieved clock is at least 1.05 times the starting design's, and its mean LUT count is at most 1.02 times the baseline at the same stage. Of the 1,536 attempts, 528 meet this rule, against 569 at the nominal clock: 495 keep the label, 74 lose it, and 33 gain it. Fable 5.1 loses the most labels, 28, which leaves it 168 routed improvements against 184 nominal; Sol 5.6 has 146, Astra 6 109, and Opus 5.5 105. Among routed improvements, the median throughput gain per agent and target is 9.7 to 20.0\% and the median LUT change $-3.8$ to $+0.3$\%. Timing closure at 126\,MHz is recorded separately; per agent and target, 65 to 87\% of the route runs close it. Table~\ref{tab:T06_scores} recomputes the overall score with the routed labels; the correctness factor is unchanged.

\begin{figure}[tbp]
\centering
\includegraphics[width=\linewidth]{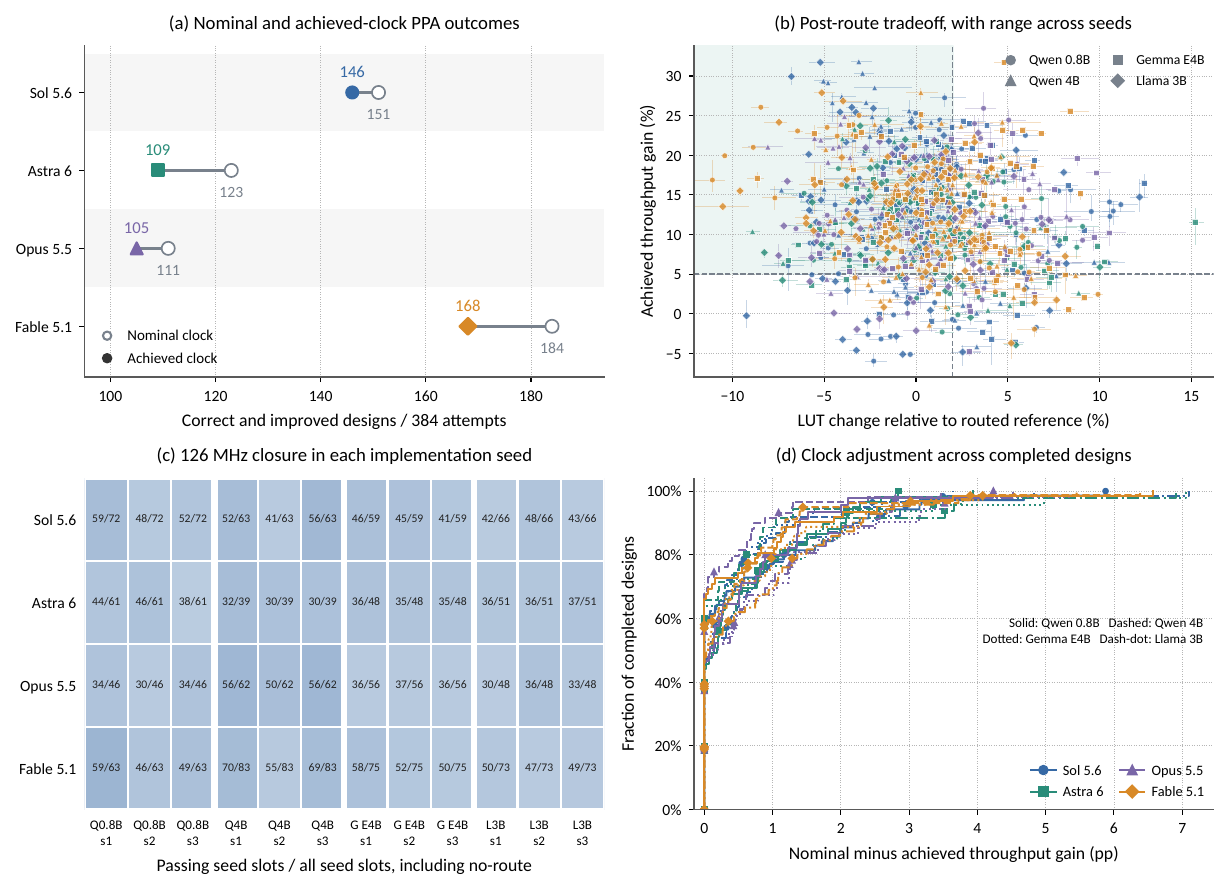}
\caption{Nominal-clock and achieved-clock PPA outcomes. Counts keep all 1,536 attempts in the denominator. Scatter points and distributions condition on three completed routes; thin ranges span the three seeds of a design. The pale region marks the improvement rule. The closure matrix keeps failed routes in its denominators. Clock-adjustment curves show the distribution of nominal minus achieved throughput gain.}
\label{fig:F04_postroute_ppa}
\end{figure}
\begin{table}[H]
\centering
\caption{\textbf{Nominal-clock and post-route PPA comparison.}}
\label{tab:T05_ppa}
{\small\tablefont\setlength{\tabcolsep}{3.5pt}\renewcommand{\arraystretch}{1.04}
\begin{tabular*}{\linewidth}{@{\extracolsep{\fill}}llrrrrrrr@{}}
\toprule
\multicolumn{2}{c}{} & \multicolumn{3}{c}{Improved designs} & \multicolumn{2}{c}{Implementation completeness} & \multicolumn{2}{c}{Median change} \\
\cmidrule(lr){3-5}\cmidrule(lr){6-7}\cmidrule(lr){8-9}
Agent & Target & Nominal & Routed & R/L/N & \shortstack{Complete\\designs} & \shortstack{126 MHz\\seed slots} & \shortstack{Rate\\\%} & \shortstack{LUT\\\%} \\
\midrule
Fable 5.1 & Qwen 0.8B & 40 & 36 & 35/5/1 & 62/63 & 154/189 & 16.2 & -1.0 \\
Fable 5.1 & Qwen 4B & 55 & 47 & 45/10/2 & 79/83 & 194/249 & 14.1 & 0.3 \\
\addlinespace[3pt]
Fable 5.1 & Gemma E4B & 42 & 34 & 33/9/1 & 71/75 & 160/225 & 14.5 & -0.9 \\
Fable 5.1 & Llama 3B & 47 & 51 & 43/4/8 & 71/73 & 146/219 & 14.1 & -1.4 \\
\addlinespace[3pt]
Sol 5.6 & Qwen 0.8B & 42 & 40 & 39/3/1 & 70/72 & 159/216 & 14.1 & -0.4 \\
Sol 5.6 & Qwen 4B & 44 & 43 & 41/3/2 & 61/63 & 149/189 & 20.0 & -1.9 \\
\addlinespace[3pt]
Sol 5.6 & Gemma E4B & 32 & 25 & 25/7/0 & 57/59 & 132/177 & 15.4 & 0.2 \\
Sol 5.6 & Llama 3B & 33 & 38 & 31/2/7 & 63/66 & 133/198 & 15.9 & -1.9 \\
Astra 6 & Qwen 0.8B & 32 & 27 & 27/5/0 & 59/61 & 128/183 & 14.0 & -0.9 \\
Astra 6 & Qwen 4B & 27 & 25 & 24/3/1 & 35/39 & 92/117 & 11.2 & -1.2 \\
Astra 6 & Gemma E4B & 31 & 27 & 27/4/0 & 47/48 & 106/144 & 12.7 & -1.1 \\
Astra 6 & Llama 3B & 33 & 30 & 27/6/3 & 48/51 & 109/153 & 14.1 & -1.5 \\
Opus 5.5 & Qwen 0.8B & 18 & 17 & 17/1/0 & 45/46 & 98/138 & 11.4 & -3.8 \\
Opus 5.5 & Qwen 4B & 42 & 43 & 39/3/4 & 59/62 & 162/186 & 12.1 & -0.8 \\
Opus 5.5 & Gemma E4B & 27 & 19 & 19/8/0 & 52/56 & 109/168 & 9.7 & -0.3 \\
Opus 5.5 & Llama 3B & 24 & 26 & 23/1/3 & 46/48 & 99/144 & 14.0 & -1.2 \\
\bottomrule
\end{tabular*}}
\par\smallskip{\footnotesize\raggedright Each row contains 96 attempts. R/L/N: retained/lost/newly improved. Improved requires correctness, three completed routes, at least 5\% achieved-throughput gain, and at most 2\% LUT growth. Rate/LUT columns are conditional medians among improved designs.\par}
\end{table}

\begin{table}[H]
\centering
\caption{\textbf{Overall score with the PPA component evaluated at achieved clock, with paired bootstrap 95\% intervals.}}
\label{tab:T06_scores}
{\small\tablefont\setlength{\tabcolsep}{3.5pt}\renewcommand{\arraystretch}{1.04}
\begin{tabular*}{\linewidth}{@{\extracolsep{\fill}}lrrrr@{}}
\toprule
\multicolumn{1}{c}{} & \multicolumn{2}{c}{Overall score} & \multicolumn{2}{c}{PPA reclassification} \\
\cmidrule(lr){2-3}\cmidrule(lr){4-5}
Agent & Original score & Updated score & Change & PPA score \\
\midrule
Fable 5.1 & 25.02 & 22.84\,\textsubscript{\tiny(19.69--26.15)} & \cellcolor{orangeC!12}-2.18 & 43.75 \\
Sol 5.6 & 24.53 & 23.71\,\textsubscript{\tiny(20.41--27.22)} & \cellcolor{orangeC!12}-0.81 & 38.02 \\
Astra 6 & 16.78 & 14.87\,\textsubscript{\tiny(12.35--17.56)} & \cellcolor{orangeC!12}-1.91 & 28.39 \\
Opus 5.5 & 16.51 & 15.62\,\textsubscript{\tiny(12.95--18.47)} & \cellcolor{orangeC!12}-0.89 & 27.34 \\
\bottomrule
\end{tabular*}}
\par\smallskip{\footnotesize\raggedright The correctness component is unchanged. The score is the product of the two track scores, as a percentage; tasks are averaged equally within a track.\par}
\end{table}

\section{Reference designs across decoder families}
\label{app:references}

\subsection{Configurations and detailed checks for additional decoders}
\label{app:extended_models}
\label{app:gemma}
\ifcompact\begin{table}[tb]
\centering
\caption{\textbf{RTL exactness and contract fidelity can each fail while the other holds.} Six reference workloads under the two numerical criteria. Five references pass every RTL check; Qwen3.6-27B has integration failures, and Gemma 2's contract misses the 99\% fidelity threshold.}
\label{tab:extended_reference}\label{tab:bench_overview}
{\small\tablefont\setlength{\tabcolsep}{3pt}\resultrows
\begin{tabular*}{\linewidth}{@{\extracolsep{\fill}}lrrr>{\bfseries}r}
\toprule
 & \multicolumn{2}{c}{\textbf{RTL vs. integer oracle}} & \multicolumn{2}{c}{\textbf{Integer oracle vs. FP32}} \\
\cmidrule(lr){2-3}\cmidrule(lr){4-5}
Target workload & Public exact & Held-out exact & Top-1 matches & Agreement (\%) \\
\midrule
Qwen3.5-0.8B & 3120/3120 & 232/232 & 7931/8000 & 99.14 \\
Qwen3.5-4B & 3680/3680 & 288/288 & 7937/8000 & 99.21 \\
\addlinespace[3pt]
Gemma 2 2B & 1320/1320 & 80/80 & 1574/1600 & \cellcolor{orange!20}98.38 \\
Gemma 3 4B (text) & 1600/1600 & 240/240 & 1584/1600 & 99.00 \\
Qwen3.6-27B & \cellcolor{orange!20}1312/1320 & \cellcolor{orange!20}70/80 & 1584/1600 & 99.00 \\
Llama 3.2-1B & 1600/1600 & 240/240 & 1589/1600 & 99.31 \\
\bottomrule
\end{tabular*}}
\par\smallskip{\footnotesize\raggedright Orange marks incomplete numerical exactness or fidelity below 99\%. Exactness counts test cases; fidelity counts top-1 decisions on common prefixes, which are correlated within a sequence. Gemma 3 4B and Llama 3.2-1B use the 32-step suites of the correctness track; Gemma 2 2B and Qwen3.6-27B use the earlier eight-step suites. Qwen3.5 tests: Appendix~\ref{app:tests}; additional decoders: Appendix~\ref{app:extended_models}.\par}
\end{table}
\fi
The four additional reference designs of Sections~\ref{sec:tasks} and~\ref{sec:correctness} (Table~\ref{tab:extended_reference}) are for Gemma 2 2B~\citep{gemma2card} (\texttt{google/gemma-2-2b}), the text decoder of Gemma 3 4B~\citep{gemma3card} (\texttt{google/gemma-3-4b-it}), Qwen3.6-27B, and Llama 3.2-1B. Image processing is excluded from the Gemma 3 task. For the three attention-only targets, the GDN and convolution components of $s_t$ in Equation~\ref{eq:transition} are empty.

The Gemma 2 2B public suite, like the suites of the earlier eight-step Gemma 3 4B and Llama 3.2-1B runs (Appendix~\ref{app:gemma3_llama1b}), has 1,000 cases on primitive operators, 128 on an attention layer, 128 on a complete decoder block, 32 on one-step model execution, 16 on eight-step decoding, and 16 on reset/isolation. Its held-out suite has 16 cases each on new token prefixes, supported nonzero cache positions, reset after activity, bounded in-order stalls, and shape-compatible parameter reload. The 32-step suites of the Gemma 3 4B and Llama 3.2-1B tasks have 1,024 operator cases, 192 on individual layers, 192 on a complete decoder block, 64 on one-step execution, 64 on 32-step decoding, and 64 on reset isolation, 1,600 in all, and 48 held-out cases in each of the same five families, 240 in all.

On common prefixes, the Gemma 2 2B contract matches FP32 on 1,574 of 1,600 top-1 decisions and the Gemma 3 4B contract on 1,584 (Table~\ref{tab:fidelity_positions} gives the percentages by position). Gemma 2 2B therefore misses the 99\% fidelity threshold although its RTL matches its integer oracle on every case. Building these reference designs involved no agent attempts.

\paragraph{Qwen3.6-27B and Llama 3.2-1B.}
The Qwen3.6-27B configuration~\citep{qwen36config} is a dense hybrid decoder with 64 layers, hidden width 5,120, and a full-attention layer every fourth layer. Like the Qwen3.5 configurations~\citep{qwen08,qwen4}, it has a gated attention output, but reusing the Qwen3.5 scaffold still required checking gate arithmetic, state layouts, and address ranges. Its public suite has 1,000 operator cases, 128 GDN-layer cases, 128 full-attention-layer cases, 32 one-step cases, 16 eight-step cases, and 16 reset cases. All operator and layer cases pass, but four one-step and four eight-step cases fail. Table~\ref{tab:heldout_reference_detail} gives its ten held-out failures by condition.

Llama 3.2-1B~\citep{llama32card} is a compact text-only decoder from a third family. Its suites have the same allocation as the Gemma 3 4B suites, under its own attention, normalization, and positional contract, and all 1,600 public and 240 held-out cases of its 32-step suite pass. 

Table~\ref{tab:fidelity_positions} gives fidelity by position for the four additional targets, with 200 decisions at each of eight positions. Pooled top-1 agreement is 1,574 of 1,600 for Gemma 2 2B, 1,584 for Gemma 3 4B, 1,584 for Qwen3.6-27B, and 1,589 for Llama 3.2-1B. Only Gemma 2 2B falls below the 99\% fidelity threshold. Qwen3.6-27B's RTL matches its integer oracle on 1,312 of 1,320 public and 70 of 80 held-out cases, Gemma 2 2B's on all 1,400 cases of its suites, and Gemma 3 4B's and Llama 3.2-1B's on all 1,840 of theirs (Table~\ref{tab:extended_reference}).

\subsubsection{Validation levels}
\label{sec:reference_stages}
The pooled case counts in Table~\ref{tab:extended_reference} weight every test case equally, but most public cases exercise operators and layers. By level, Gemma 2 2B, Gemma 3 4B, and Llama 3.2-1B pass every level. Qwen3.6-27B passes 1,312 of 1,320 public cases overall: all 1,256 operator and single-layer cases, but only 28/32 (87.5\%) of the one-step cases and 12/16 (75.0\%) of the eight-step cases.

\begin{table}[H]
\centering
\caption{\textbf{Held-out reference checks by condition.} Qwen3.6-27B fails most often under bounded memory stalls.}
\label{tab:heldout_reference_detail}
{\small\tablefont\setlength{\tabcolsep}{3pt}\resultrows
\begin{tabular*}{\linewidth}{@{\extracolsep{\fill}}l*{5}{c}}
\toprule
Target & New prefixes & Cache position & Reset & Stalls & Reload \\
\midrule
Gemma 2 2B & 16/16 & 16/16 & 16/16 & 16/16 & 16/16 \\
Gemma 3 4B (text) & 16/16 & 16/16 & 16/16 & 16/16 & 16/16 \\
Qwen3.6-27B & \cellcolor{orange!20}14/16 & \cellcolor{orange!20}13/16 & 16/16 & \cellcolor{orange!20}12/16 & \cellcolor{orange!20}15/16 \\
Llama 3.2-1B & 16/16 & 16/16 & 16/16 & 16/16 & 16/16 \\
\bottomrule
\end{tabular*}}
\par\smallskip{\footnotesize\raggedright Exact cases / total. Reset is reset after activity; stalls are bounded and in order; parameters are reloaded while idle. Five conditions of 16 cases give 80 per target. Configuration details: Appendix~\ref{app:extended_models}.\par}
\end{table}

Table~\ref{tab:heldout_reference_detail} breaks the held-out cases down by condition. Qwen3.6-27B's failures span several conditions: four of 16 cases fail under bounded stalls, three at supported cache positions, two on new prefixes, and one on idle parameter reload, while all 16 reset-after-activity cases pass. The other three targets pass all five conditions under their own contracts.

\subsubsection{Fidelity by position}
\label{sec:position_fidelity}
\begin{table}[H]
\centering
\caption{\textbf{Fidelity by decode position.} Top-1 agreement with FP32 (\%) on common prefixes.}
\label{tab:fidelity_positions}
{\small\tablefont\setlength{\tabcolsep}{2pt}\resultrows
\begin{tabular*}{\linewidth}{@{\extracolsep{\fill}}l*{9}{r}}
\toprule
Target & 1 & 2 & 3 & 4 & 5 & 6 & 7 & 8 & Pooled \\
\midrule
Qwen3.5-0.8B & 99.4 & 99.5 & \cellcolor{orange!20}98.7 & 99.0 & 99.5 & \cellcolor{orange!20}98.3 & 99.4 & 99.3 & \textbf{99.14} \\
Qwen3.5-4B & 99.5 & 99.1 & 99.1 & 99.0 & 99.0 & 99.5 & 99.3 & 99.2 & \textbf{99.21} \\
Gemma 2 2B & 99.0 & \cellcolor{orange!20}97.5 & \cellcolor{orange!20}98.5 & 99.0 & \cellcolor{orange!20}97.5 & \cellcolor{orange!20}97.5 & 99.0 & 99.0 & \cellcolor{orange!20}\textbf{98.38} \\
Gemma 3 4B (text) & 99.5 & 99.0 & 99.5 & 99.0 & \cellcolor{orange!20}98.5 & \cellcolor{orange!20}98.5 & 99.5 & \cellcolor{orange!20}98.5 & \textbf{99.00} \\
Qwen3.6-27B & 99.5 & 99.0 & \cellcolor{orange!20}98.5 & 99.0 & 100.0 & 99.0 & 99.5 & \cellcolor{orange!20}97.5 & \textbf{99.00} \\
Llama 3.2-1B & 100.0 & 99.0 & \cellcolor{orange!20}98.0 & 99.5 & 99.5 & 100.0 & 99.5 & 99.0 & \textbf{99.31} \\
\bottomrule
\end{tabular*}}
\par\smallskip{\footnotesize\raggedright Orange marks values below 99\%; the fidelity threshold applies to the pooled score. Each position has 1,000 decisions for Qwen3.5-0.8B/4B and 200 for the other targets. Positions share sequences, so their decisions are correlated.\par}
\end{table}

Table~\ref{tab:fidelity_positions} breaks the pooled fidelity of all six targets into eight decode positions. Qwen3.5-0.8B meets the threshold on the pooled score but falls below 99\% at positions 3 and 6, whereas Qwen3.5-4B stays at or above 99\% at every position. Gemma 3 4B and Qwen3.6-27B both pool to exactly 99\%, with three and two positions below it, and Llama 3.2-1B's pooled 99.31\% includes 98.0\% at position 3. Gemma 2 2B has the lowest pooled fidelity, with four positions below 99\% and a minimum of 97.5\%, although its RTL is exact on every case.

\subsection{Qwen3.6-27B repairs and Gemma 2 2B precision}
\label{app:add-reference}
Qwen3.6-27B and Gemma 2 2B do not qualify, for different reasons (Section~\ref{sec:correctness}). We revised the Qwen3.6-27B RTL twice and evaluated eight alternative integer contracts for Gemma 2 2B.

The two revisions of the Qwen3.6-27B reference reduce its failures from eighteen cases, all at the integration levels (Appendix~\ref{sec:reference_stages}), to one or two (Figure~\ref{fig:F08_reference}a, Table~\ref{tab:T09-integration}). Revision A passes all 80 held-out cases but misses one eight-step public case and one regression case under stalls. Revision B passes every public and regression case but misses one held-out parameter-reload case. Neither revision passes every check, so Qwen3.6-27B does not qualify.

Gemma 2 2B's fidelity shortfall is a property of its contract: seven of eight alternative integer contracts reach the 99\% fidelity threshold, from 99.12\% with 8-bit calibrated per-channel scaling to 99.90\% at 16 bits (Table~\ref{tab:T09-precision}). Higher precision costs hardware: LUT counts rise from 58.4k to 84.1k, and throughput falls from 1.48 to 0.90 tokens/s. All four additional reference designs meet timing, and Gemma 3 4B and Llama 3.2-1B pass every gate (Table~\ref{tab:T09-gates}).

\begin{figure}[htbp]
\centering
\includegraphics[width=\linewidth,height=.62\textheight,keepaspectratio]{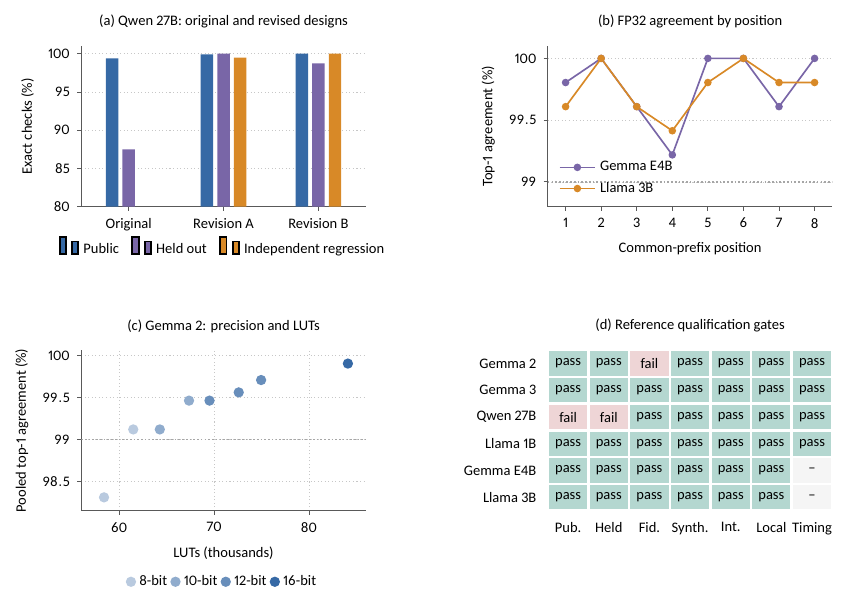}
\caption{Reference studies. (a) Qwen3.6-27B: the public and held-out suites for the first design and for two revisions, with the revisions\textquotesingle{} regression suites. (b) FP32 agreement by common-prefix position for Gemma 4 E4B and Llama 3.2 3B; positions share sequences, and the 99\% line is the threshold for the pooled score. (c) Gemma 2 2B contracts of different precision against LUT count. (d) Qualification gates by target (numerical, fidelity, synthesis, integrity, local, and timing); a dash marks timing that is not available.}
\label{fig:F08_reference}
\end{figure}

\begin{table}[htbp]
\centering
\caption{\textbf{Qwen3.6-27B integration and regression suites by design version.}}
\label{tab:T09-integration}
{\small\tablefont\setlength{\tabcolsep}{7pt}\resultrows
\begin{tabular}{@{}lllr}
\toprule
Cond. & Design & Suite & Passed (\%) \\
\midrule
Main & Original & Public & 1312/1320 (99.4) \\
Main & Original & Held out & 70/80 (87.5) \\
Main & Revision A & Public & 1319/1320 (99.9) \\
Main & Revision A & Held out & 80/80 (100.0) \\
Main & Revision A & Independent regression & 199/200 (99.5) \\
Main & Revision B & Public & 1320/1320 (100.0) \\
Main & Revision B & Held out & 79/80 (98.8) \\
Main & Revision B & Independent regression & 200/200 (100.0) \\
\bottomrule
\end{tabular}
}
\par\smallskip{\footnotesize\raggedright Each row is one design version and suite. A failed test family shows where a version fails; identifying the defect requires further diagnosis.\par}
\end{table}
\begin{table}[htbp]
\centering
\caption{\textbf{Gemma 2 2B precision and resource tradeoffs.}}
\label{tab:T09-precision}
{\small\tablefont\setlength{\tabcolsep}{4pt}\resultrows
\begin{tabular*}{\linewidth}{@{\extracolsep{\fill}}llrrrrrrr@{}}
\toprule
Contract & Bits & Rounding & LUT entries & Matches & \% & $\geq$99\% & LUT k & Token/s \\
\midrule
Original & \textemdash & Unspecified & \textemdash & 1574/1600 & 98.375 & \cellcolor{orange!18}fail & \textemdash & \textemdash \\
G2-C01 & 8 & Toward zero & 256 & 4027/4096 & 98.315 & \cellcolor{orange!18}fail & 58.4 & 1.482 \\
G2-C02 & 8 & Nearest even & 512 & 4060/4096 & 99.121 & pass & 61.5 & 1.432 \\
G2-C03 & 10 & Nearest even & 256 & 4060/4096 & 99.121 & pass & 64.3 & 1.285 \\
G2-C04 & 10 & Nearest even & 512 & 4074/4096 & 99.463 & pass & 67.4 & 1.242 \\
G2-C05 & 12 & Nearest even & 256 & 4074/4096 & 99.463 & pass & 69.5 & 1.144 \\
G2-C06 & 12 & Toward zero & 512 & 4078/4096 & 99.561 & pass & 72.6 & 1.105 \\
G2-C07 & 12 & Nearest even & 1,024 & 4084/4096 & 99.707 & pass & 75.0 & 1.085 \\
G2-C08 & 16 & Nearest even & 1,024 & 4092/4096 & 99.902 & pass & 84.1 & 0.903 \\
\bottomrule
\end{tabular*}
}
\par\smallskip{\footnotesize\raggedright Each row is a separate integer contract with its own oracle. The RTL is exact under every contract; the $\geq$99\% column tests pooled FP32 agreement against the fidelity threshold.\par}
\end{table}
\begin{table}[htbp]
\centering
\caption{\textbf{Numerical and hardware gates for the four additional reference designs.}}
\label{tab:T09-gates}
{\small\tablefont\setlength{\tabcolsep}{4pt}\resultrows
\begin{tabular*}{\linewidth}{@{\extracolsep{\fill}}lrrrrrrrr@{}}
\toprule
Target & Public & Held out & Fid. & Synth. & Integr. & Local & Timing & WNS (ns) \\
\midrule
Gemma 2 2B & pass & pass & \cellcolor{orange!18}fail & pass & pass & pass & pass & 0.26 \\
Gemma 3 4B & pass & pass & pass & pass & pass & pass & pass & 0.20 \\
Qwen 27B & \cellcolor{orange!18}fail & \cellcolor{orange!18}fail & pass & pass & pass & pass & pass & 0.07 \\
Llama 1B & pass & pass & pass & pass & pass & pass & pass & 0.05 \\
\bottomrule
\end{tabular*}
}
\par\smallskip{\footnotesize\raggedright Qualification uses the Public through Local columns; Timing and the worst negative slack (WNS) are reported for reference. Shaded cells mark unmet criteria.\par}
\end{table}

\section{Earlier task series}
\label{app:cohorts}
\label{sec:agents}
\label{app:add-primary}
\label{app:sa-comparison}
\label{sec:agent_hardware}
Before the runs of the correctness track, the four agents ran an earlier series of four tasks under full-state feedback: Qwen3.5-nano T1 (one decoder layer), T2 (the four-layer decoder for one step), and T3 (eight recurrent steps), each built from interfaces with 16 attempts per agent, and scaffold adaptation of Qwen3.5-0.8B with 12 attempts per agent (Table~\ref{tab:sa-budgets}). The nano tasks follow the order in which a decoder accelerator is assembled, so each task requires more integration than the one before. Wall-time caps and test suites also grow from T1 to T3, so a difference between T1 and T3 combines integration, horizon, budget, and suite size. Sol 5.6 and Astra 6 ran these tasks first, on shared task instances, for 96 nano attempts and 24 adaptation attempts. A feedback pilot added 32 output-only Sol 5.6 attempts on T2 and T3, which gives the 152 attempts of this cohort. Opus 5.5 and Fable 5.1 ran the four tasks later with the same harness, caps, test counts, and feedback. Because the two cohorts ran at different times, we compare agents only within a cohort. 

\begin{table}[t]
\centering\small
\caption{\textbf{Task allocation and caps for Sol 5.6 and Astra 6.} Pairs is the number of task instances, each attempted by both agents. Output k is the output-token cap in thousands; Public and Held out are test counts. The feedback comparison adds 16 output-only Sol 5.6 attempts each on T2 and T3 and pairs them with the full-state attempts counted here. The 0.8B task supplies a shared scaffold; the nano tasks supply their interface and starting files.}
\label{tab:sa-budgets}
\tablefont
\begin{tabular}{lrrrrrr}
\toprule
Task & Pairs & Wall h & Tool cap & Output k & Public & Held out \\
\midrule
Nano T1 & 16 & 3 & 180 & 90 & 1128 & 128 \\
Nano T2 & 16 & 5 & 260 & 140 & 1320 & 64 \\
Nano T3 & 16 & 6 & 320 & 180 & 1384 & 96 \\
Qwen 0.8B & 12 & 12 & 420 & 260 & 3120 & 232 \\
\bottomrule
\end{tabular}
\end{table}

\begin{table}[tb]
\centering
\caption{\textbf{Earlier task series: every agent has fewer attempts accepted on the recurrent decoder (T3) than on the single layer (T1).} Bold rows give final acceptance as rates and counts, with public-pass counts below. Each agent has 16 attempts on each nano task and 12 on 0.8B, all under full-state feedback.}
\label{tab:key_results}
{\small\tablefont\setlength{\tabcolsep}{4pt}\resultrows
\begin{tabularx}{\linewidth}{@{}>{\raggedright\arraybackslash}p{.17\linewidth}*{4}{>{\centering\arraybackslash}X}>{\centering\arraybackslash}p{.11\linewidth}@{}}
\toprule
Agent / measure & {\resultstack\begin{tabular}[b]{@{}c@{}}\textbf{T1}\\Single layer\end{tabular}} & {\resultstack\begin{tabular}[b]{@{}c@{}}\textbf{T2}\\Full decoder\\(one step)\end{tabular}} & {\resultstack\begin{tabular}[b]{@{}c@{}}\textbf{T3}\\Recurrent\\(eight steps)\end{tabular}} & {\resultstack\begin{tabular}[b]{@{}c@{}}\textbf{0.8B}\\Scaffold\\adaptation\end{tabular}} & {\resultstack\begin{tabular}[b]{@{}c@{}}T1 to T3\\drop (pp)\end{tabular}} \\
\midrule
\textbf{Sol 5.6} & \cellcolor{blueC!41}{\resultstack\begin{tabular}[b]{@{}c@{}}\textbf{81.3\%}\\{\scriptsize 13/16}\end{tabular}} & \cellcolor{blueC!33}{\resultstack\begin{tabular}[b]{@{}c@{}}\textbf{62.5\%}\\{\scriptsize 10/16}\end{tabular}} & \cellcolor{blueC!18}{\resultstack\begin{tabular}[b]{@{}c@{}}\textbf{31.3\%}\\{\scriptsize 5/16}\end{tabular}} & \cellcolor{blueC!31}{\resultstack\begin{tabular}[b]{@{}c@{}}\textbf{58.3\%}\\{\scriptsize 7/12}\end{tabular}} & \textbf{50.0} \\
{\footnotesize Public pass} & {\footnotesize 14/16} & {\footnotesize 12/16} & {\footnotesize 9/16} & {\footnotesize 9/12} & --- \\
\addlinespace[3pt]
\textbf{Astra 6} & \cellcolor{blueC!38}{\resultstack\begin{tabular}[b]{@{}c@{}}\textbf{75.0\%}\\{\scriptsize 12/16}\end{tabular}} & \cellcolor{blueC!36}{\resultstack\begin{tabular}[b]{@{}c@{}}\textbf{68.8\%}\\{\scriptsize 11/16}\end{tabular}} & \cellcolor{blueC!24}{\resultstack\begin{tabular}[b]{@{}c@{}}\textbf{43.8\%}\\{\scriptsize 7/16}\end{tabular}} & \cellcolor{blueC!35}{\resultstack\begin{tabular}[b]{@{}c@{}}\textbf{66.7\%}\\{\scriptsize 8/12}\end{tabular}} & \textbf{31.3} \\
{\footnotesize Public pass} & {\footnotesize 14/16} & {\footnotesize 13/16} & {\footnotesize 10/16} & {\footnotesize 10/12} & --- \\
\midrule
\textbf{Opus 5.5} & \cellcolor{blueC!30}{\resultstack\begin{tabular}[b]{@{}c@{}}\textbf{56.3\%}\\{\scriptsize 9/16}\end{tabular}} & \cellcolor{blueC!33}{\resultstack\begin{tabular}[b]{@{}c@{}}\textbf{62.5\%}\\{\scriptsize 10/16}\end{tabular}} & \cellcolor{blueC!27}{\resultstack\begin{tabular}[b]{@{}c@{}}\textbf{50.0\%}\\{\scriptsize 8/16}\end{tabular}} & \cellcolor{blueC!38}{\resultstack\begin{tabular}[b]{@{}c@{}}\textbf{75.0\%}\\{\scriptsize 9/12}\end{tabular}} & \textbf{6.3} \\
{\footnotesize Public pass} & {\footnotesize 11/16} & {\footnotesize 10/16} & {\footnotesize 11/16} & {\footnotesize 10/12} & --- \\
\addlinespace[3pt]
\textbf{Fable 5.1} & \cellcolor{blueC!38}{\resultstack\begin{tabular}[b]{@{}c@{}}\textbf{75.0\%}\\{\scriptsize 12/16}\end{tabular}} & \cellcolor{blueC!38}{\resultstack\begin{tabular}[b]{@{}c@{}}\textbf{75.0\%}\\{\scriptsize 12/16}\end{tabular}} & \cellcolor{blueC!33}{\resultstack\begin{tabular}[b]{@{}c@{}}\textbf{62.5\%}\\{\scriptsize 10/16}\end{tabular}} & \cellcolor{blueC!46}{\resultstack\begin{tabular}[b]{@{}c@{}}\textbf{91.7\%}\\{\scriptsize 11/12}\end{tabular}} & \textbf{12.5} \\
{\footnotesize Public pass} & {\footnotesize 12/16} & {\footnotesize 13/16} & {\footnotesize 10/16} & {\footnotesize 11/12} & --- \\
\bottomrule
\end{tabularx}
}
\par\smallskip{\footnotesize\raggedright Final acceptance requires every gate; attempts that end at a cap count as failures. Blue shading uses one rate scale. Within a task, agents share starting files, tools, and caps. Caps and test suites also grow from T1 to T3, so the drop describes the tasks as a whole (Appendix~\ref{sec:agents}). Sol 5.6 and Astra 6 ran first; Opus 5.5 and Fable 5.1 ran later with the same harness, caps, and test counts. pp: percentage points, computed before rounding.\par}
\end{table}

\paragraph{Acceptance.}
Every agent has fewer attempts accepted on the recurrent decoder (T3) than on the single layer (T1) (Table~\ref{tab:key_results}): by 50.0 and 31.3 percentage points for Sol 5.6 and Astra 6, and by 12.5 and 6.3 points for Fable 5.1 and Opus 5.5. Pooled over the four agents, acceptance is similar on T1 and T2 (46/64 and 43/64) and falls to 30/64 on T3. The controlled matrix (Appendix~\ref{app:add-state}) varies the horizon from one starting implementation. On Qwen3.5-0.8B, 7 to 11 of each agent's 12 attempts are accepted. This task measures adaptation of a common parameterized scaffold. 

Passing the public suite overstates acceptance in both cohorts. Nine Sol 5.6 and nine Astra 6 submissions pass it and fail final acceptance, 16 of the 18 on the held-out suite, as do six Opus 5.5 submissions (three diagnosed as integration, two as state/protocol, and one as arithmetic) and one Fable 5.1 submission (arithmetic). Lower acceptance and longer attempts also compound in cost: from T1 to T3, agent-hours per accepted design grow from 2.32 to 15.14 for Sol 5.6 and from 2.48 to 10.09 for Astra 6, and Opus 5.5 and Fable 5.1 need 7.39 and 5.96 on T3.

\paragraph{Feedback pilot.}
Sol 5.6 attempted 16 instances each of T2 and T3 under output-only feedback, paired with its full-state attempts on the same instances. Full-state feedback adds two accepted attempts per task, from 8 to 10 of 16 on T2 and from 3 to 5 of 16 on T3, with four discordant pairs each (exact McNemar $p=0.625$). The feedback study of Appendix~\ref{app:add-feedback} repeats the comparison with four agents and 32 instances per task. Listing~\ref{lst:sa-output-only} gives the record of one output-only T3 attempt that passes the public checks visible to the agent and fails the recheck after submission. In the record, \texttt{visible\_public} counts the public checks the agent saw, \texttt{canonical\_public} the state-aware public recheck, and \texttt{hidden\_numerical} the held-out suite.

\paragraph{Hardware of accepted designs.}
Throughput is the 126\,MHz target clock divided by elapsed cycles per token, at batch size one and an eight-token prompt under the common memory-service model. For the seven accepted Sol 5.6 designs and the eight accepted Astra 6 designs on Qwen3.5-0.8B, median decode throughput is 4.398 and 4.337 tokens/s, with ranges of 4.240--4.494 and 4.225--4.527 tokens/s; median mapped LUT counts are 61,995 and 61,910 (Table~\ref{tab:sa-qor}). Sol 5.6 has the higher median throughput, but the two nondominated designs in the throughput--LUT plane are both Astra 6 submissions, at (61,325 LUTs, 4.392 tokens/s) and (61,680 LUTs, 4.527 tokens/s). The PPA track compares the agents' optimization from a common correct design (Appendix~\ref{app:add-optimization}).

\begin{table}[H]
\centering\small
\caption{\textbf{Hardware of accepted Qwen3.5-0.8B designs.} Medians and throughput ranges are over each agent's accepted designs.}
\label{tab:sa-qor}
\tablefont
\begin{tabular*}{\linewidth}{@{\extracolsep{\fill}}lrrlrrr}
\toprule
Agent & Accepted & Median tok/s & Range & Median LUT & BRAM & DSP \\
\midrule
Sol 5.6 & 7 & 4.398 & 4.240--4.494 & 61,995 & 226 & 128 \\
Astra 6 & 8 & 4.337 & 4.225--4.527 & 61,910 & 224 & 128 \\
\bottomrule
\end{tabular*}
\end{table}

\begin{lstlisting}[float=tp,basicstyle=\ttfamily\scriptsize,breaklines=true,label={lst:sa-output-only},caption={Record of an output-only T3 attempt by Sol 5.6. The submission passes the public checks visible to the agent and fails the state-aware recheck after submission. Times are elapsed seconds.}]
run_id=sol_T3_oo_04
agent=Sol 5.6; harness=MathCode
task=T3; pair=T3-04; feedback=output_only
agent_exit_s=18417; termination=submitted
tool_calls=118; failed_calls=7
input_tokens=309874; output_tokens=75993
visible_public=1384/1384
freeze; agent_access=closed; grader_feedback_to_agent=false
canonical_public=1378/1384
hidden_numerical=95/96
synthesis=true; integrity=true
local_required=false; local_exact=0/0
grader_exit_s=18687
accepted=false; outcome=public_integration
\end{lstlisting}

\section{Benchmark sizes and scoring units}
\label{app:related_sizes}
\begin{table}[H]
\centering
\caption{Size and scoring of representative RTL benchmarks: VerilogEval~\citep{liu2023verilogeval}, RTLLM~\citep{rtllm}, RTLLM 2.0~\citep{openllm}, RTL-Repo~\citep{rtlrepo}, TuRTLe~\citep{turtle} and CVDP~\citep{cvdp}. Counts keep each source's unit. TuRTLe reuses several of the other datasets, so its counts overlap theirs. For AIChipBench, the count is agent tasks; its 9,136 agent attempts are 152 by Sol 5.6 and Astra 6 on the earlier task series (Appendix~\ref{app:sa-comparison}) and 8,984 in all other experiments (Table~\ref{tab:additional_attempts}).}
\label{tab:related_scale}
{\tiny
\setlength{\tabcolsep}{4pt}
\renewcommand{\arraystretch}{1.1}
\tablefont
\begin{tabularx}{\linewidth}{@{}l r l >{\raggedright\arraybackslash}X >{\raggedright\arraybackslash}X@{}}
\toprule
Benchmark & Count & Unit & Content & Scoring \\
\midrule
VerilogEval & 156 & problems & HDLBits modules, from combinational logic to state machines & Simulation against a reference \\
RTLLM & 30 & designs & Median 52 HDL lines per design; median 121 netlist cells over 29 designs (one RAM design excluded) & Syntax, function, synthesized design quality \\
RTLLM 2.0 & 50 & designs & Descriptions, reference RTL and test cases & Syntax, function, design quality \\
RTL-Repo & 4,098 & examples & Multi-file Verilog projects; 1,174-example test split & Exact match and edit similarity of one completed line \\
TuRTLe & 1,174 + 223 & samples & 1,174 RTL-Repo line completions and 223 designs from VeriGen (17), VerilogEval (156) and RTLLM (50); 40 open LLMs evaluated & Syntax, function, synthesis, PPA; exact match for line completion \\
CVDP & 783 & problems & 13 categories of design, verification, debugging and comprehension; agentic and non-agentic formats & Hidden test harnesses; BLEU or LLM judging for comprehension answers \\
\midrule
\bfseries AIChipBench & \bfseries 13 & \bfseries tasks & \bfseries Twelve correctness-track tasks (Qwen3.5-nano T2 and T3, Qwen3.5-0.8B and 4B, and Gemma 4 E4B, Llama 3.2 3B, Gemma 3 4B, and Llama 3.2-1B from two starts) and T1 of the earlier series; 9,136 agent attempts & \bfseries Exact outputs and state on public and held-out checks; synthesis; integrity; local checks where required \\
\bottomrule
\end{tabularx}}
\end{table}

\begin{table}[H]
\centering
\caption{\textbf{Neighboring benchmarks and what they grade,} as their sources describe them.}
\label{tab:related}
{\footnotesize\tablefont\setlength{\tabcolsep}{4pt}
\begin{tabularx}{\linewidth}{@{}>{\raggedright\arraybackslash}p{0.19\linewidth}>{\raggedright\arraybackslash}X>{\raggedright\arraybackslash}X@{}}
\toprule
Benchmark & What the system produces & How it is graded \\
\midrule
VerilogEval, RTLLM, TuRTLe, RTL-Repo & A module or design from a description, or a completed line & Simulation against references; synthesis or PPA in RTLLM and TuRTLe \\
ChipBench & Hierarchical modules, CPU components, bug fixes, reference models & Golden references on random stimuli \\
CVDP, LBC-bench & Design, verification, debugging, and comprehension outputs; agentic repository tasks & Hidden test harnesses; BLEU or LLM judging for comprehension \\
Build-A-Chip-Bench & An ASIC layout from a workload specification & Hidden functional tests, routability, timing, design rules; PPA \\
AIChipBench & Decoder RTL in a fixed module structure, and faster correct designs, over repeated budgeted attempts & Exact outputs and committed state against an integer oracle under held-out sequences, resets, and stalls; synthesis; PPA after FPGA routing; board power for reference designs \\
\bottomrule
\end{tabularx}}
\end{table}

\ifcompact\paragraph{Verification across representations.}
EDASYTH~\citep{edasyth} gates transitions among symbolic calculations, simulation, and circuit artifacts, EEBench~\citep{eebench} evaluates electrical engineering designs through simulation and cost, and the large-circuit-model perspective~\citep{lcm} studies learning across specifications, RTL, netlists, and layouts. AIChipBench checks one transition, from a quantized model to RTL, against an executable state-transition contract.

\paragraph{Other work on LLMs for hardware design.}
Architect Labs~\citep{architect} studies RTL generation with verifiable rewards, ChatEDA~\citep{chateda} orchestrates the tools that carry RTL to a manufacturable layout (RTL-to-GDSII), and ChipNeMo~\citep{chipnemo} studies engineering assistance and EDA script generation.\fi

\end{document}